\pdfoutput=1
\documentclass{article}

\IfFileExists{iclr2027_conference.sty}{\usepackage{iclr2027_conference}}{\usepackage{local_review}}
\usepackage{times}

\usepackage[utf8]{inputenc}
\usepackage[T1]{fontenc}
\usepackage{hyperref}
\usepackage{url,xurl,microtype,placeins}
\usepackage{booktabs,multirow}
\usepackage[table]{xcolor}
\definecolor{oraclegray}{gray}{0.90}
\definecolor{lamcolor}{RGB}{180,0,110}
\definecolor{refblue}{HTML}{24558A}
\hypersetup{colorlinks=true,linkcolor=refblue,citecolor=refblue,urlcolor=refblue}
\newif\ifshowrevision
\showrevisionfalse
\definecolor{revcolor}{RGB}{0,0,255}
\ifshowrevision
  \DeclareRobustCommand{\rev}[1]{{\color{revcolor}#1}}
  \newenvironment{revblock}{\color{revcolor}}{}
\else
  \DeclareRobustCommand{\rev}[1]{#1}
  \newenvironment{revblock}{}{}
\fi
\usepackage{graphicx}
\usepackage{float}
\usepackage{amsfonts,amsmath,amssymb,amsthm,bm}
\usepackage{centernot}
\usepackage{algorithm}
\usepackage{algpseudocode}
\usepackage{tikz}
\usetikzlibrary{positioning, arrows.meta}
\usepackage{subcaption}
\usepackage{wrapfig}
\usepackage{needspace}

\newtheorem{theorem}{Theorem}
\newtheorem{proposition}{Proposition}
\newtheorem{corollary}{Corollary}
\newtheorem{lemma}{Lemma}
\newtheorem{definition}{Definition}
\newtheorem{assumption}{Assumption}
\newtheorem{theoremprime}{Theorem}

\theoremstyle{remark}
\newtheorem{remark}{Remark}

\newcommand{\E}{\mathbb{E}}
\newcommand{\M}{\mathcal{M}}
\newcommand{\MB}{\mathcal{M}_{\mathcal{B}^{a}}}
\newcommand{\MBz}{\mathcal{M}_{\mathcal{B}^{z}}}
\newcommand{\B}{\mathcal{B}}
\newcommand{\Ba}{\mathcal{B}^{a}}
\newcommand{\Bz}{\mathcal{B}^{z}}
\newcommand{\MBzH}{\mathcal{M}_{\mathcal{B}^{z}_{H}}}
\newcommand{\pH}{p^{(1)}_{\mathcal{B},H}}
\newcommand{\rH}{r^{(1)}_{\mathcal{B},H}}
\newcommand{\Bdec}{\mathcal{B}^{\mathrm{dec}}}
\newcommand{\Dlab}{\mathcal{D}_{L}}
\newcommand{\TV}{D_{\mathrm{TV}}}
\DeclareMathOperator*{\supp}{supp}

\newcommand{\restated}[3]{\medskip\noindent\textbf{#1~\ref{#2}} (#3).\ }

\title{Latent Action Policies from Suboptimal Data}

\author{%
\parbox[t]{\dimexpr\textwidth-2\tabcolsep\relax}{\centering\normalfont
Donghyeon Ki$^{1,*}$, Xiaoyu Chen$^{2}$, Jun Ho Seo$^{1}$, Hayeong Lee$^{1}$, Kyungyoon Kim$^{1}$, \\[2pt]
Byung-Jun Lee$^{1,\dagger}$, Li Zhao$^{2,\dagger}$ \\[5pt]
{\small $^{1}$Korea University \qquad $^{2}$Microsoft Research Asia}
}%
}
\date{}
\iclrfinalcopy
\hypersetup{
  pdftitle={Latent Action Policies from Suboptimal Data},
  pdfauthor={Donghyeon Ki, Xiaoyu Chen, Jun Ho Seo, Hayeong Lee, Kyungyoon Kim, Byung-Jun Lee, Li Zhao}
}

\begin{document}

\maketitle
\lhead{Preprint}
\begingroup
\renewcommand{\thefootnote}{\ensuremath{*}}
\footnotetext[3]{Interns at Microsoft Research Asia.}
\renewcommand{\thefootnote}{\ensuremath{\dagger}}
\footnotetext[0]{Corresponding authors.}
\endgroup

\begin{abstract}
Latent Action Models (LAMs) learn control representations from action-free video, but existing pipelines typically train latent policies with behavioral cloning, without an explicit policy improvement objective. Although offline RL offers a natural approach to policy improvement, standard methods do not transfer directly to learned latent action spaces. We find that the structure of learned latent actions makes naive offline RL unreliable: directly maximizing value over latent actions or fitting a unimodal policy can produce latent actions that cannot be realized by physical execution, while the latent representation can merge physical controls that receive different rewards. To understand when latent-space optimization remains valid, we derive conditions under which policy improvement transfers to decoded execution and under which latent and physical optima align. Building on these insights, we propose LAPIS for goal-conditioned latent policy improvement, grounding value-based policy improvement in observed state--latent transitions and using a conditional flow policy to capture their multimodal structure. \rev{LAPIS outperforms baselines on most tasks, with the clearest gains in visual manipulation, and remains effective even with only 1\% action labels, where competing methods largely collapse.}
\end{abstract}

\section{Introduction}\label{sec:intro}

Latent Action Models (LAMs) have emerged as a promising paradigm for learning control representations from action-free video by inferring latent actions that explain observed state transitions~\citep{ye2025lapa,chen2025villax,chen2024igor}. Because latent-action learning captures transition and outcome structure rather than task optimality, it can naturally exploit large collections of diverse and suboptimal video, including failed trajectories that would be poor demonstrations but remain informative for learning how the environment changes. Yet existing LAM pipelines typically train the resulting latent policies with behavioral cloning (BC)~\citep{schmidt2024lapo,ye2025lapa,chen2024igor,bruce2024genie}, without explicitly optimizing latent policies for improved task return. This creates a gap between the broad data that LAMs can learn from and the policies that current pipelines can extract from it.

\begin{figure}[t]
\centering
\includegraphics[width=\textwidth]{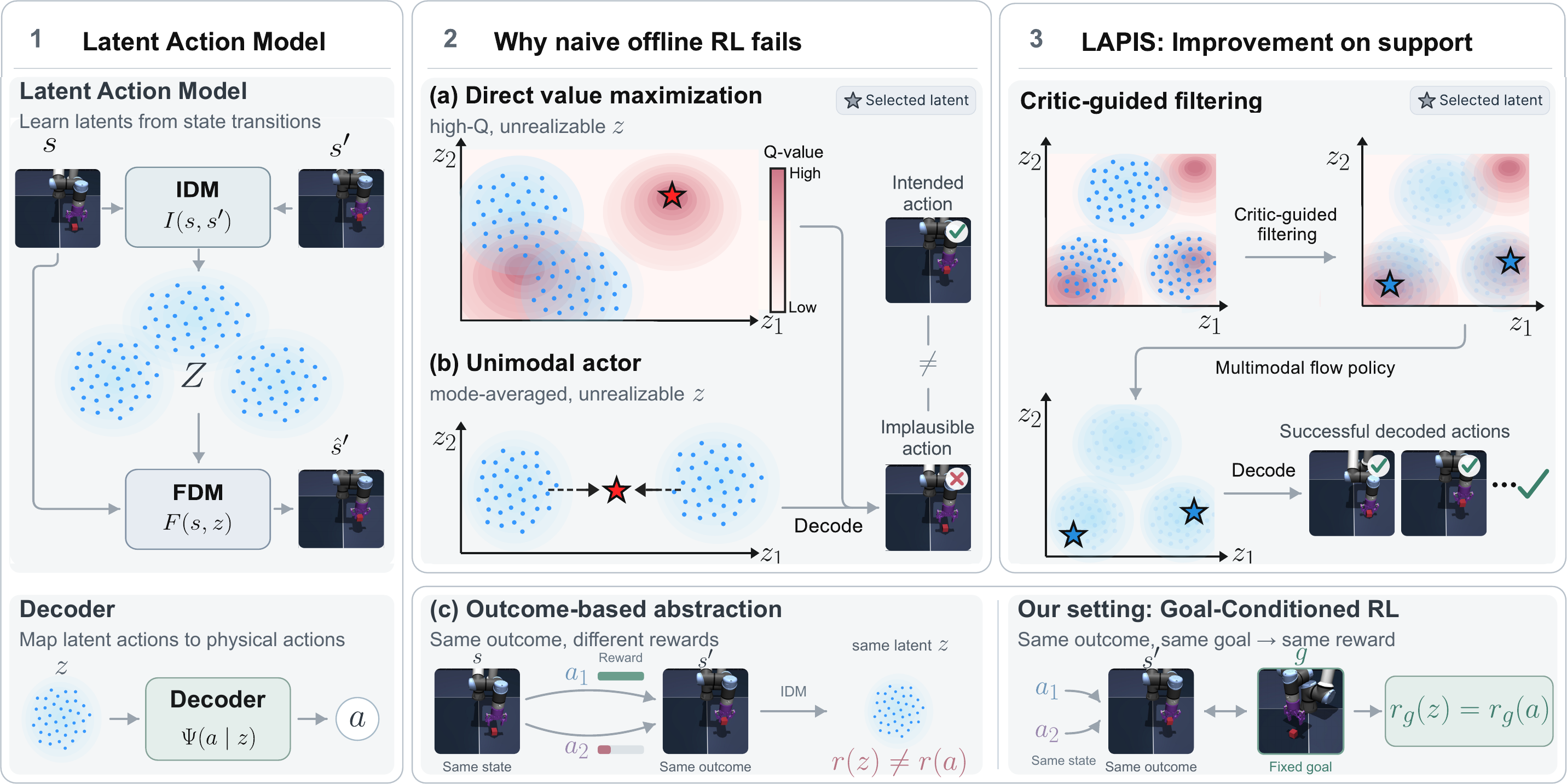}
\caption{\textbf{Overview of LAPIS.} Left: a Latent Action Model learns latent actions from action-free trajectories and a decoder from limited action labels. Middle: naive offline RL can produce unrealizable latent actions through value maximization or unimodal policy fitting, while outcome-based abstraction can merge reward-distinct controls. Right: LAPIS performs support-aware policy improvement with critic-guided filtering and a multimodal flow policy, instantiated with an outcome-compatible goal-conditioned objective.}
\label{fig:overview}
\end{figure}

Offline reinforcement learning (RL) offers a natural way to improve beyond the observed behavior~\citep{levine2020offline,kostrikov2022iql,fujimoto2019bcq,ki2025prior,lu2025gooddiffusion}, but applying standard offline RL directly in a learned latent action space is not straightforward. We identify three properties that distinguish learned latent actions from physical actions. The first two concern \emph{realizability}---whether the transition outcome represented by a latent action can be physically produced from the current state: (i) \emph{state-conditioned realizability}, whereby a latent action realizable at one state can be unrealizable at another; and (ii) \emph{non-convex realizable set}, whereby interpolating between realizable latent actions can yield unrealizable ones. The third, (iii) \emph{outcome-based abstraction}, arises because distinct physical controls that reach the same outcome can map to the same latent action.

These properties expose two distinct failure modes of naive offline RL in latent action spaces. The first is an optimization--execution mismatch. Because the realizable latent set is state-dependent and potentially non-convex, standard offline RL can move outside this set by directly maximizing latent-action value or through unimodal policy fitting, producing latent actions that cannot be faithfully realized by physical execution. This raises the question: \emph{when does latent-space policy improvement transfer to decoded physical execution?} The second is an objective mismatch. Because outcome-based abstraction can merge physical controls with different rewards, the optimal latent policy need not correspond to the optimal physical policy. This raises a separate question: \emph{when do the latent and physical policy optima align?}

To answer these questions, we analyze paired physical- and latent-action batch MDPs constructed from the same observed transitions. For the first question, we show that batch-constrained latent policy improvement transfers to physical execution under deterministic dynamics and supported outcome identifiability. For the second, we show that latent and physical optima align under an outcome-determined objective within the decoder-covered, batch-constrained physical policy class. These results characterize when latent-space policy improvement remains valid and provide the principles behind our method.

Building on these insights, we introduce LAPIS (Latent Action Policy Improvement on Support), a latent policy improvement framework motivated by these failure modes. Rather than directly optimizing a latent actor toward arbitrary high-value actions, LAPIS grounds value-based policy improvement in observed state--latent transitions and uses a conditional flow policy to capture their potentially multimodal structure. LAPIS is instantiated with a goal-conditioned objective based on future states, which is naturally compatible with the outcome semantics captured by latent actions. At inference, classifier-free guidance provides a simple way to adjust the emphasis between behavior and high-value policy branches. \rev{With the same LAM and $5\%$ action labels, LAPIS achieves the highest success among latent-space methods on all five OGBench visual manipulation tasks and outperforms prior semi-supervised offline RL baselines on nearly all tasks with only $5\%$ or $1\%$ action labels.} Further experiments examine the contributions of these design choices.

\section{Related Work}
\label{sec:related_work}

\paragraph{Latent action models from action-free video}
A growing body of work learns controllable representations from action-free video by inferring latent actions that explain observed transitions~\citep{edwards2019ilpo, rybkin2019learning, baker2022vpt, bruce2024genie, genie2}. Recent approaches learn such representations using discrete or continuous latent codes and extend latent action modeling to richer video and vision-language-action settings~\citep{schmidt2024lapo, ye2025lapa, chen2024igor, gao2025adaworld, zhang2025whatlams, chen2025villax}. Their control pipelines predominantly train latent policies by behavioral cloning and use a small labeled set to decode latent actions into physical controls. LAPO~\citep{schmidt2024lapo} additionally incorporates reinforcement learning by applying online PPO to fine-tune the physical action policy after latent-action pretraining. In contrast, we study policy improvement directly in the learned latent action space using offline data, with scarce action labels used for physical action decoding.

\paragraph{Action-free and semi-supervised reinforcement learning.}
A related line of work studies learning control from trajectories with missing or limited action annotations, using inferred actions, latent representations, or other forms of auxiliary supervision to exploit action-free experience~\citep{torabi2018bco,baker2022vpt,ghosh2023icvf,park2023hiql,zheng2023semiaf}. Most closely related is LAQ~\citep{chang2022laq}, which performs Q-learning over latent actions from state-only data but relies on a hand-specified controller for execution. In contrast, we learn a latent action space from action-free data together with a physical-action decoder from scarce action labels, and study offline policy improvement directly in this learned latent action space. This setting makes the relationship between latent-space optimization and decoded physical execution central, motivating our analysis of when latent policy improvement transfers to physical control. Further related work is discussed in Appendix~\ref{app:related_work}.

\section{Preliminaries}\label{sec:prelim}

\paragraph{Goal-conditioned reinforcement learning}
We model the environment as an MDP $\M=(S,A,p,r,\gamma)$, with dynamics $p(s'\mid s,a)$, bounded reward $r$, and initial-state distribution $\mu_0$. We write $(s,a,s')=(s_t,a_t,s_{t+1})$ for a transition. A policy $\pi(a\mid s)$ induces the usual state-action and state value functions $Q^\pi_\M$ and $V^\pi_\M$, with expected return $J_\M(\pi)=\E_{s_0\sim\mu_0}[V^\pi_\M(s_0)]$. We refer to $\M$ as the ground MDP to distinguish it from the batch MDPs introduced in Section~\ref{sec:setup}. In the goal-conditioned setting, the policy additionally conditions on a goal $g$, $\pi(a\mid s,g)$. Contrastive RL~\citep{eysenbach2022crl} casts goal-conditioned control as density-ratio estimation, bypassing explicit modeling of transition dynamics or reward functions. The goal-reaching reward is $(1-\gamma)\,p(s'{=}g\mid s,a)$, and the corresponding goal-conditioned value equals the discounted future-state occupancy:
\begin{align}
Q^\pi_g(s,a) &= p^\pi(s_{t+}{=}g\mid s,a)
= (1-\gamma)\sum_{k=1}^{\infty}\gamma^{k-1}p^\pi(s_{t+k}{=}g\mid s_t=s,a_t=a),
\end{align}
where the right-hand side denotes the discounted occupancy of future states visited from $(s,a)$ under $\pi$. To illustrate the contrastive principle, consider a critic parameterized as a bilinear score $f_\phi(s,a,g)=\phi(s,a)^\top\psi(g)/\sqrt{d}$ and trained by noise-contrastive binary classification, with a positive goal $g^+$ sampled from the future-state occupancy and a negative goal $g^-\sim p_B$ sampled from the dataset state distribution,
\begin{equation}\label{eq:nce}
\max_{\phi,\psi}\ \E_{(s,a),\,g^+,\,g^-}\bigl[\log\sigma(f_\phi(s,a,g^+))+\log(1-\sigma(f_\phi(s,a,g^-)))\bigr].
\end{equation}
For the ideal binary classification problem, the optimal score is the log density ratio, $f^*=\log Q^\pi_g-\log p_B(g)$ wherever the ratio is defined. LAPIS uses a TD-InfoNCE estimator, detailed in Appendix~\ref{app:tdinfonce}. In the offline setting, directly maximizing this critic can drive the actor outside the support of the fixed behavior data, so offline CRL regularizes policy improvement with behavioral cloning,
\[
\mathcal{L}_{\mathrm{CRL}}(\pi)=\E_{(s,a,g)\sim\mathcal{D}}\bigl[-f_\phi(s,\pi(s,g),g)+\alpha\,\|\pi(s,g)-a\|_2^2\bigr],
\]
improving the actor toward high-value actions while keeping it close to the observed behavior.

\paragraph{Latent action models}
A Latent Action Model (LAM) learns a latent action space $Z$ from an action-free dataset $\B=\{(s_i,s_i')\}_{i=1}^N$ by training an inverse dynamics model $I$ and a forward dynamics model $F$, with $z=I(s,s')$ and $\hat s'=F(s,z)$ with a bottlenecked latent representation,
\begin{equation}\label{eq:lam}
\min_{I,F}\ \E_{(s,s')\sim\B}\bigl[\|s'-F(s,I(s,s'))\|_2^2\bigr].
\end{equation}
Relabeling each transition by its latent action $z_i=I(s_i,s_i')$ yields the latent dataset $\Bz=\{(s_i,z_i,s_i')\}_{i=1}^N$. A separate small labeled dataset $\Bdec=\{(s_i,a_i,s_i')\}_{i=1}^{N_{\mathrm{dec}}}$ with $N_{\mathrm{dec}}\ll N$ supplies the action supervision for the decoder.

The deployed system composes a \emph{latent policy} with a \emph{decoder}. The latent policy $\pi_z(z\mid s)$ selects a latent transition at the current state, and the state-marginal decoder $\Psi(a\mid z)$ maps the chosen latent to a physical action, trained on $\Bdec$ by $\max_\Psi \E_{\Bdec}[\log\Psi(a\mid z)]$ and frozen thereafter. Deployment samples $z\sim\pi_z(\cdot\mid s)$ then $a\sim\Psi(\cdot\mid z)$, giving the deployed policy $\bar\pi=\Psi\circ\pi_z$. Section~\ref{sec:setup} introduces an ideal state-conditioned batch decoder $\Psi^*(a\mid s,z)$ only for the transfer analysis.

\section{Latent Action Structure and Failures}\label{sec:analysis}

A latent policy selects a latent action representing a desired transition outcome, which must ultimately be realized through physical control at deployment. Section~\ref{sec:properties} characterizes the realizability and abstraction structure of learned latent actions. Section~\ref{sec:consequences} then shows how realizability can break the connection between latent-space optimization and physical execution, while outcome abstraction can misalign latent and physical objectives.

\subsection{Structural properties of learned latent action spaces}\label{sec:properties}

\begin{figure}[t]
\centering
\includegraphics[width=\textwidth]{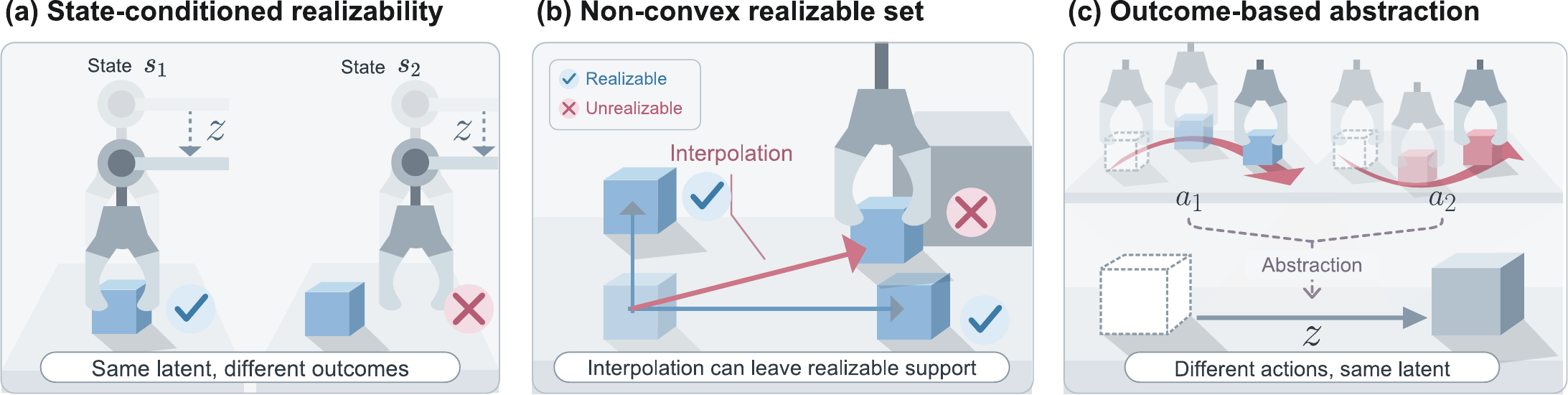}\par
\vspace{3pt}
\includegraphics[width=\textwidth]{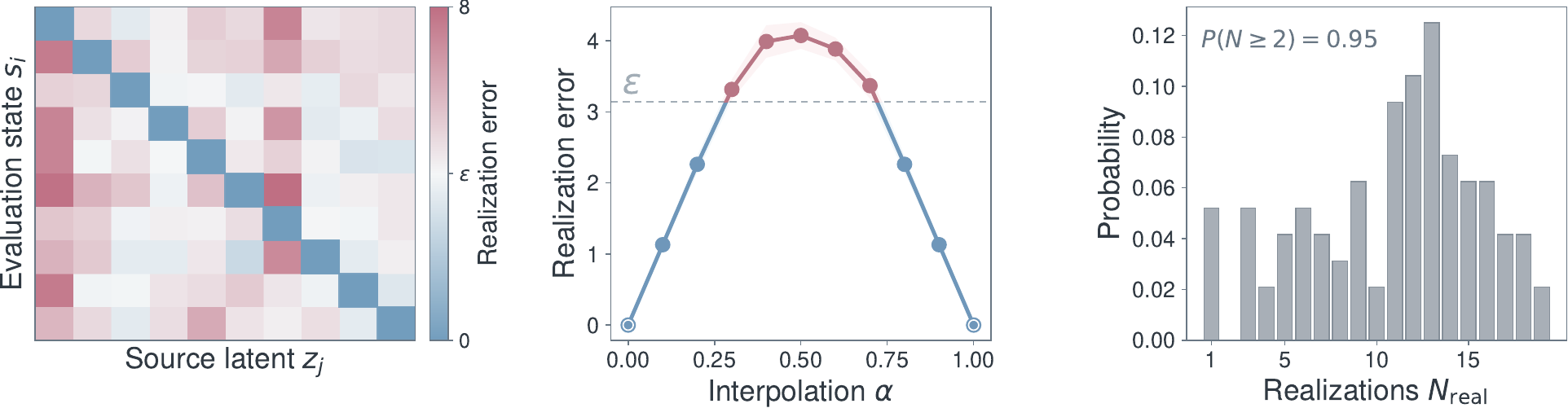}
\caption{\textbf{Structural properties of learned latent action spaces.} \emph{Top row:} concept schematics for \textbf{(a)}~\emph{state-conditioned realizability} (the same latent can be realizable at $s_1$ but not $s_2$), \textbf{(b)}~the \emph{non-convex realizable set} (interpolating two realizable latents can pass through unrealizable ones), and \textbf{(c)}~\emph{outcome-based abstraction} (distinct physical controls reaching the same successor share one latent). \emph{Bottom row:} empirical diagnostics for the experimental LAM on \nolinkurl{visual-cube-single-play-v0}; the diagnostic protocol is in Appendix~\ref{app:fig1-protocol}.}
\label{fig:properties_concept}
\vspace{-0.5\baselineskip}
\end{figure}

To \rev{quantify} whether a target latent can be physically realized from the current state, we define its realization error as the infimum of its distance to the latents of reachable outcomes. \rev{We use it as a diagnostic (Figure~\ref{fig:properties_concept}, Appendix~\ref{app:fig1-protocol}) and to bound the transfer gap outside the data (Appendix~\ref{app:realization-return}).}

\begin{definition}[Latent realizability]\label{def:realizable}
Let $\mathcal R(s):=\bigcup_{a\in A}\supp p(\cdot\mid s,a)$ be the set of successor states reachable from $s$. The \emph{latent realization error} of a target latent $z$ at state $s$ is
\[
\delta(s,z)=\inf_{s'\in\mathcal R(s)}\bigl\|I(s,s')-z\bigr\|_2,
\]
and $z$ is \emph{$\epsilon$-realizable} at $s$ if $\delta(s,z)\le\epsilon$, defining the realizable set $\mathcal Z^\epsilon_{\mathrm{real}}(s)=\{z:\delta(s,z)\le\epsilon\}$.
\end{definition}

The observed set $\Bz$ contains only a subset of realizable state--latent pairs; absence from $\Bz$ does not imply unrealizability. Figure~\ref{fig:properties_concept} illustrates the properties discussed below.

\paragraph{State-conditioned realizability} Whether a latent is realizable depends on the current state: the same $z$ can be realizable from $s_1$ but not $s_2$, so realizability is a property of the pair $(s,z)$, not of $z$ alone ($\mathcal Z^\epsilon_{\mathrm{real}}(s_1)\neq\mathcal Z^\epsilon_{\mathrm{real}}(s_2)$). In Figure~\ref{fig:properties_concept}(a), we take latents with known realizations at their source states and search for physical realizations of the same latents from other states. Latents have near-zero realization error at their source states, while $73\%$ of transfers to other states exceed the diagnostic threshold under finite search. \rev{LAPO~\citep[Sec.~4.1]{schmidt2024lapo} notes that the action behind a latent can depend on the state; this is harmless when imitating the latents observed at $s$, but offline RL moves away from them and can select latents that are realizable only from other states.}

\paragraph{Non-convex realizable set} Even at a fixed state, the space between two realizable latents need not be realizable. For $z_1,z_2\in\mathcal Z^\epsilon_{\mathrm{real}}(s)$, an interpolation $z_\alpha=(1-\alpha)z_1+\alpha z_2$ need not be realizable, so $\mathcal Z^\epsilon_{\mathrm{real}}(s)$ need not be convex. In Figure~\ref{fig:properties_concept}(b), we select pairs of realizable endpoint latents at the same state and search for physical realizations along the interpolation between them. The endpoints admit known physical realizations, while interior errors exceed the diagnostic threshold.

\paragraph{Outcome-based abstraction} The encoder $z=I(s,s')$ labels the transition outcome rather than the physical command, so distinct actions reaching the same successor from $s$ share a latent. In Figure~\ref{fig:properties_concept}(c), we search for substantially different action chunks that reach each target outcome within a fixed tolerance; $95\%$ of sampled targets admit at least two such realizations.

\subsection{Why the properties break naive offline RL}\label{sec:consequences}

These properties become consequential when offline RL is applied in the learned latent space, giving rise to two distinct failure modes: an optimization--execution mismatch and an objective mismatch.

\paragraph{Optimization--execution mismatch} \emph{State-conditioned realizability} and a \emph{non-convex realizable set} imply that physically realizable latent actions depend on the current state, while interpolations between realizable latents can be unrealizable. As a result, directly optimizing the actor to maximize learned value can move it toward unrealizable latents despite behavior regularization, while a unimodal actor can place probability between separated realizable modes and generate unrealizable latents. Thus, even when the critic favors a high-value latent action $z$, if $z$ is unrealizable, decoding and physical execution can reach a different successor than the outcome represented by $z$. Consequently, latent-space policy improvement need not translate to improvement after decoding,
\[
\mathcal J_{\mathrm{latent}}(\pi_z')\ge\mathcal J_{\mathrm{latent}}(\pi_z)
\;\;\centernot{\Longrightarrow}\;\;
J_\M(\Psi\circ\pi_z')\ge J_\M(\Psi\circ\pi_z).
\]
Here $\mathcal J_{\mathrm{latent}}$ denotes the learned latent-space optimization objective.

\paragraph{Objective mismatch} \emph{Outcome-based abstraction} can erase reward-relevant distinctions between physical controls. When distinct controls $a_1\neq a_2$ reach the same successor $s'$, they share the latent action $z=I(s,s')$, even if an action-dependent reward assigns $r(s,a_1,s')\neq r(s,a_2,s')$. Because the latent policy can select only $z$, it cannot distinguish between these reward-distinct physical controls. Consequently, a policy that is optimal in the latent-action MDP need not decode to an optimal physical-action policy,
\[
\pi_z^*\in\arg\max_{\pi_z\in\Pi_{\Bz}}J_{\MBz}(\pi_z)
\;\;\centernot{\Longrightarrow}\;\;
(\Psi^*\circ\pi_z^*)\in\arg\max_{\pi\in\Pi_{\Ba}}J_\M(\pi).
\]
Here, $\Pi_{\Ba}$ and $\Pi_{\Bz}$ denote the physical and latent policy classes restricted to observed state--action and state--latent pairs, respectively, and $\Psi^\star$ is the batch-exact decoder defined in Section~\ref{sec:setup}.

These failure modes raise two questions: when does latent-space policy improvement transfer to physical execution, and when do latent and physical policy optima align? Section~\ref{sec:cfg-crl} establishes conditions for both, which motivate the design of LAPIS in Section~\ref{sec:lapis}.

\section{When Does Latent Policy Improvement Work?}\label{sec:cfg-crl}\label{sec:setup}

To answer these questions, we construct paired batch MDPs from the same observed transitions under physical and latent action representations,
\[
\MB=(S\cup\{s_{\mathrm{abs}}\},A,p_{\Ba},r_{\Ba},\gamma),\qquad
\MBz=(S\cup\{s_{\mathrm{abs}}\},Z,p_{\Bz},r_{\Bz},\gamma).
\]
The two MDPs are constructed from $\Ba=\{(s_i,a_i,s_i')\}_{i=1}^N$ and its latent relabeling $\Bz=\{(s_i,z_i,s_i')\}_{i=1}^N$, with $z_i=I(s_i,s_i')$, while unobserved pairs transition to the zero-reward absorbing state $s_{\mathrm{abs}}$. Because realizability is state-conditioned, support in $\MBz$ is over observed pairs $(s,z)$, not $z$ globally. We further assume \emph{supported outcome identifiability}: each supported $(s,z)$ corresponds to a unique successor $F_{\B}(s,z)$, so that a supported latent action has a well-defined transition outcome at each state. This supported successor map is distinct from the learned forward predictor $F$. For the analysis, we deploy a latent policy through the \emph{batch-exact state-conditioned decoder} $\Psi^*(a\mid s,z)$, which selects physical actions observed at $s$ to produce the successor represented by $z$ in the batch. The support domain, initial-state and terminal conventions, batch rewards, and decoder construction are specified in Appendix~\ref{app:proofs-notation}. All exact results below use this common construction. Appendix~\ref{app:prior-theory} discusses how our analysis extends classical batch-constrained RL and MDP abstraction to learned latent actions.

\subsection{When does latent policy improvement transfer?}

We first ask when an improvement measured in the latent batch MDP is preserved by decoded physical execution. To avoid selecting unrealizable latent actions, we restrict the latent policy to state--latent transitions observed in the batch.

\begin{definition}[Batch-constrained policy]\label{def:bc}
A policy is batch-constrained if its trajectory uses only observed state--action pairs in $\Ba$. A latent batch-constrained policy is defined analogously over $\Bz$.
\end{definition}

\rev{Every observed state--latent pair $(s,z)\in\Bz$ has a physical realization by construction ($\delta(s,z)=0$ in Definition~\ref{def:realizable}), and the batch-exact decoder $\Psi^*$ selects physical actions observed at $s$ that realize this outcome.} A batch-constrained latent policy selects only latent actions that are physically realizable at the current state. 

Support alone, however, does not ensure exact transfer. Under stochastic dynamics, the same decoded action can reach a different successor from the one used to evaluate $z$ in the latent batch MDP, breaking the correspondence between latent evaluation and physical execution. We therefore assume deterministic ground dynamics, which leads to Theorem~\ref{thm:eval}.

\begin{assumption}[Deterministic ground dynamics]\label{ass:detT}
The successor is a deterministic function of the current state and physical action, $s'=T(s,a)$.
\end{assumption}

\begin{theorem}[On-support policy improvement transfer]\label{thm:eval}
Under Assumption~\ref{ass:detT}, let $\pi_z$ be a batch-constrained latent policy. Then the latent return equals the decoded physical return, $J_{\MBz}(\pi_z)=J_\M(\Psi^*\circ\pi_z)$, and consequently, for any two such latent policies $\pi_z$ and $\pi_z'$,
\[
J_{\MBz}(\pi_z')\ge J_{\MBz}(\pi_z)\quad\Longrightarrow\quad J_\M(\Psi^*\circ\pi_z')\ge J_\M(\Psi^*\circ\pi_z).
\]
\end{theorem}

\vspace{-0.6\baselineskip}
\begin{proof}
See Appendix~\ref{app:proof-eval-exact}.
\end{proof}

Thus, return improvement between batch-constrained latent policies transfers to their batch-exact decoded policies. Appendix~\ref{app:stochastic-transfer} extends this exact result to stochastic dynamics, bounding the transfer gap by the mismatch between the latent batch transition and the transition induced by decoded execution. \rev{Appendix~\ref{app:realization-return} bounds the gap outside the data via the realization error (Definition~\ref{def:realizable}).}

\subsection{When do latent and physical optima align?}

Even when latent policy improvement transfers to decoded execution as in Theorem~\ref{thm:eval}, this alone does not guarantee alignment between the optimal latent and physical-action policies. Under a generic action-dependent reward, \emph{outcome-based abstraction} can collapse reward-distinct physical controls into the same latent action, removing potentially necessary distinctions for physical optimality.

\begin{assumption}[Outcome-determined reward]\label{ass:reward}
The reward depends on a transition only through the current and next states, $r(s,a,s')=r(s,s')$.
\end{assumption}

Under Assumption~\ref{ass:reward}, physical actions reaching the same successor from the same state receive the same immediate reward, removing the reward ambiguity introduced by outcome-based abstraction. We therefore obtain the following alignment result.

\begin{theorem}[Optimal-policy alignment]\label{thm:align}
Under Assumptions~\ref{ass:detT}--\ref{ass:reward}, for the same initial distribution $\mu_0$,
\[
\pi_z^*\in\arg\max_{\pi_z\in\Pi_{\Bz}}J_{\MBz}(\pi_z)
\quad\Longrightarrow\quad
(\Psi^*\circ\pi_z^*)\in\arg\max_{\pi\in\Pi_{\Ba}}J_\M(\pi).
\]
\end{theorem}

\vspace{-0.6\baselineskip}
\begin{proof}
See Appendix~\ref{app:proof-anypolicy}.
\end{proof}

\section{LAPIS: Policy Improvement on Support}\label{sec:lapis}

Section~\ref{sec:cfg-crl} motivates batch-constrained latent policy improvement over supported state--latent transitions. LAPIS follows this principle by using critic-guided filtering to select high-value transitions from the observed state--latent pairs and a conditional flow policy to capture their potentially multimodal distribution. We instantiate LAPIS in goal-conditioned RL, using future states as goals to obtain a reward-label-free objective compatible with outcome-based latent actions. Appendix~\ref{app:gaps} relates the practical algorithm to the supported-improvement analysis and explains how the learned components instantiate its principles.

\paragraph{Critic-guided transition filtering}\label{sec:gate}
LAPIS fits a goal-conditioned contrastive critic $f_\phi(s,z,g)$ for the behavior latent policy $\pi_\beta^z$ and uses it to score only the state--latent transitions observed in $\Bz$. We retain high-value transitions using a learned state--goal threshold,
\begin{equation}\label{eq:gate}
G(s,z,g):=\mathbf1\bigl[f_\phi(s,z,g)>\hat V(s,g)\bigr],
\end{equation}
where $\hat V(s,g)$ is learned by expectile regression in the critic's score space (Appendix~\ref{app:tdinfonce}). We show in Appendix~\ref{app:proofs-gate} that, for the optimal density-ratio score $f^\star$, critic-guided filtering corresponds to a supported reweighting that preserves support and improves return. \rev{For the TD-InfoNCE critic used in practice, this guarantee does not apply directly: its score is determined only up to an offset $c(s,z)$, which can change the ordering of latents (Appendix~\ref{app:critic-ordering}). We therefore compare the raw score with an offset-free score in Section~\ref{sec:exp-offset}.}

\paragraph{Multimodal policy extraction}\label{sec:cfgcrl-algo}
To capture the potentially multimodal distribution of gated high-value transitions, LAPIS uses conditional flow matching\rev{~\citep{lipman2023flow}}. For $x_0\sim\mathcal N(0,I)$ and target latent $z$, we use the linear path $x_t=(1-t)x_0+tz$ with target velocity $z-x_0$. The velocity model $v_\theta(x_t,t;s,g,o)$ is trained with two branches: $o{=}0$ models all observed transitions, while $o{=}1$ models only the high-value transitions selected by $G(s,z,g)$,
\begin{equation}\label{eq:actor}
\mathcal{L}_{\mathrm{actor}}=\E\bigl[G(s,z,g)\,\|v_\theta(x_t,t;s,g,1)-(z-x_0)\|_2^2+\lambda\,\|v_\theta(x_t,t;s,g,0)-(z-x_0)\|_2^2\bigr].
\end{equation}
At inference, classifier-free guidance\rev{~\citep{ho2022cfg}} combines the behavior and high-value branches,
\[
v^{\mathrm{CFG}}_w=(1-w)\,v_\theta(\cdot;0)+w\,v_\theta(\cdot;1),
\]
where $w{=}0$ recovers the behavior branch, $w{=}1$ the high-value branch, and $w{>}1$ further emphasizes the high-value branch.

\begin{algorithm}[t]
\caption{LAPIS training and deployment}\label{alg:cfgcrl}
\begin{algorithmic}[1]
\Require transition batch $\mathcal B$, labeled set $\mathcal B_{\mathrm{dec}}$, guidance scale $w$
\State Train IDM $I(s,s')$, FDM $F(s,z)$, and decoder $\Psi(a\mid z)$; form $\mathcal B^z$
\Repeat
    \State Score with $f_\phi(s,z,g)$ and gate with $G(s,z,g)$ in Eq.~\eqref{eq:gate}
    \State Fit $v_\theta(x_t,t;s,g,o)$ by minimizing $\mathcal L_{\mathrm{actor}}$ in Eq.~\eqref{eq:actor}
\Until{convergence}
\State \textbf{Deploy:} generate $z$ with $v_w^{\mathrm{CFG}}$ and execute $a\sim\Psi(\cdot\mid z)$
\end{algorithmic}
\end{algorithm}

\section{Experiments}\label{sec:exp}

Our experiments address \rev{six} questions: \textbf{(Q1)} How does LAPIS perform in learned latent action spaces, and does standard goal-conditioned offline RL remain effective? \textbf{(Q2)} How does LAPIS compare with goal-conditioned baselines adapted to limited action supervision? \textbf{(Q3)} How important is it to keep latent policy improvement on supported state--latent transitions? \textbf{(Q4)} Which components of LAPIS drive its gains? \textbf{(Q5)} How does decoder conditioning affect performance across action-label budgets? \rev{\textbf{(Q6)} Does the offset in the TD-InfoNCE score change the gate's decisions?}

\paragraph{Benchmarks} We evaluate on the OGBench visual suite~\citep{park2025ogbench}, covering \texttt{cube} and \texttt{scene} manipulation as well as \texttt{antmaze} navigation, and additionally on visual D4RL Franka Kitchen~\citep{fu2020d4rl,gupta2020relay} for multi-subtask manipulation. All tasks are evaluated in a goal-conditioned offline RL setting with $64\times64$ RGB observations; example observations from OGBench and Kitchen are shown in Figures~\ref{fig:envs} and~\ref{fig:kitchen-envs}, respectively. We use \emph{$x\%$ labels} to denote a setting where $x\%$ of the trajectories have action annotations while the remaining $100{-}x\%$ are available as action-free data, and evaluate at $100\%$, $5\%$, and $1\%$ labels.

\paragraph{Baselines and evaluation} In Tables~\ref{tab:main}--\ref{tab:semisup}, \textcolor{lamcolor}{magenta method names} indicate the use of a LAM. \rev{We compare LAPIS against GCBC, flow-based GCBC (F-GCBC), GCIVL~\citep{park2025ogbench}, goal-conditioned IQL \citep[GCIQL;][]{kostrikov2022iql}, QRL~\citep{wang2023qrl}, HIQL~\citep{park2023hiql}, and contrastive RL \citep[CRL;][]{eysenbach2022crl} in Table~\ref{tab:main}. Table~\ref{tab:semisup} extends the comparison to goal-conditioned baselines adapted to limited action supervision, adding SS-ORL~\citep{zheng2023semiaf} and VPT-RL~\citep{baker2022vpt}.} For OGBench, we report success rates as mean $\pm$ std over four seeds, evaluating $20$ episodes on each of five tasks per environment. All latent action models use a temporal gap of $H=3$, with $z_t=I(s_t,s_{t+H})$. \rev{The controller executes only the first decoded action before replanning, and Theorem~\ref{thm:eval-h} in Appendix~\ref{app:proof-hstep} extends Theorem~\ref{thm:eval} to this setting.} Detailed implementation settings for each algorithm are provided in Appendix~\ref{app:expdetails}.

\subsection{Goal-conditioned offline RL in learned latent action spaces (Q1)}\label{sec:exp-main}
\vspace{-0.3\baselineskip}
\begin{table}[t]
\centering

\setlength{\tabcolsep}{3pt}
\caption{\rev{Success rate (\%) on the OGBench visual suite. Gray columns use real actions with $100\%$ labels; the others use the LAM with $5\%$ labels. LAPIS$^\dagger$ uses the same critic with IDQL-style selection (Appendix~\ref{app:extraction-grid}).}}
\label{tab:main}
\vspace{-0.65\baselineskip}
\resizebox{\textwidth}{!}{%
\begin{tabular}{l >{\columncolor{oraclegray}}c >{\columncolor{oraclegray}}c >{\columncolor{oraclegray}}c >{\columncolor{oraclegray}}c >{\columncolor{oraclegray}}c >{\columncolor{oraclegray}}c >{\columncolor{oraclegray}}c >{\columncolor{oraclegray}}c cccccccc}
\toprule
& \multicolumn{8}{>{\columncolor{oraclegray}}c}{$100\%$ labels (oracle)} & \multicolumn{8}{c}{$5\%$ labels (w/ LAM)} \\
\cmidrule(lr){2-9}\cmidrule(lr){10-17}
\texttt{Task} & \texttt{GCBC} & \texttt{F-GCBC} & \rev{\texttt{GCIVL}} & \texttt{GCIQL} & \rev{\texttt{QRL}} & \rev{\texttt{HIQL}} & \texttt{CRL} & \texttt{LAPIS} & \textcolor{lamcolor}{\texttt{F-GCBC}} & \textcolor{lamcolor}{\texttt{GCIVL}} & \textcolor{lamcolor}{\texttt{GCIQL}} & \textcolor{lamcolor}{\texttt{QRL}} & \textcolor{lamcolor}{\texttt{HIQL}} & \textcolor{lamcolor}{\texttt{CRL}} & \textcolor{lamcolor}{\texttt{LAPIS}} & \textcolor{lamcolor}{\texttt{LAPIS}$^\dagger$} \\
\midrule
\nolinkurl{antmaze-medium} & \rev{12{\scriptsize$\,\pm0$}} & 8{\scriptsize$\,\pm1$} & \rev{23{\scriptsize$\,\pm3$}} & \rev{12{\scriptsize$\,\pm2$}} & \rev{0{\scriptsize$\,\pm0$}} & \rev{94{\scriptsize$\,\pm2$}} & \textbf{95}{\scriptsize$\,\pm2$} & 48{\scriptsize$\,\pm3$} & 8{\scriptsize$\,\pm1$} & \rev{26{\scriptsize$\,\pm5$}} & 7{\scriptsize$\,\pm2$} & \rev{34{\scriptsize$\,\pm4$}} & \rev{\textbf{89}{\scriptsize$\,\pm2$}} & 11{\scriptsize$\,\pm2$} & 49{\scriptsize$\,\pm3$} & \rev{\textbf{89}{\scriptsize$\,\pm4$}} \\
\nolinkurl{antmaze-large} & \rev{4{\scriptsize$\,\pm2$}} & 1{\scriptsize$\,\pm0$} & \rev{6{\scriptsize$\,\pm1$}} & \rev{3{\scriptsize$\,\pm1$}} & \rev{1{\scriptsize$\,\pm1$}} & \rev{66{\scriptsize$\,\pm3$}} & \textbf{85}{\scriptsize$\,\pm3$} & 14{\scriptsize$\,\pm3$} & 1{\scriptsize$\,\pm0$} & \rev{6{\scriptsize$\,\pm2$}} & 2{\scriptsize$\,\pm1$} & \rev{9{\scriptsize$\,\pm3$}} & \rev{49{\scriptsize$\,\pm4$}} & 3{\scriptsize$\,\pm1$} & 10{\scriptsize$\,\pm2$} & \rev{\textbf{58}{\scriptsize$\,\pm6$}} \\
\nolinkurl{antmaze-giant} & \rev{0{\scriptsize$\,\pm0$}} & 0{\scriptsize$\,\pm0$} & \rev{1{\scriptsize$\,\pm1$}} & \rev{0{\scriptsize$\,\pm0$}} & \rev{1{\scriptsize$\,\pm1$}} & \rev{8{\scriptsize$\,\pm5$}} & \textbf{47}{\scriptsize$\,\pm1$} & 0{\scriptsize$\,\pm0$} & 0{\scriptsize$\,\pm0$} & \rev{1{\scriptsize$\,\pm1$}} & 0{\scriptsize$\,\pm0$} & \rev{0{\scriptsize$\,\pm0$}} & \rev{3{\scriptsize$\,\pm2$}} & 0{\scriptsize$\,\pm0$} & 0{\scriptsize$\,\pm0$} & \rev{\textbf{4}{\scriptsize$\,\pm2$}} \\
\nolinkurl{antmaze-teleport} & \rev{5{\scriptsize$\,\pm0$}} & 7{\scriptsize$\,\pm1$} & \rev{8{\scriptsize$\,\pm1$}} & \rev{7{\scriptsize$\,\pm1$}} & \rev{10{\scriptsize$\,\pm4$}} & \rev{42{\scriptsize$\,\pm4$}} & \textbf{51}{\scriptsize$\,\pm1$} & 23{\scriptsize$\,\pm1$} & 6{\scriptsize$\,\pm1$} & \rev{6{\scriptsize$\,\pm2$}} & 4{\scriptsize$\,\pm1$} & \rev{14{\scriptsize$\,\pm7$}} & \rev{34{\scriptsize$\,\pm3$}} & 6{\scriptsize$\,\pm1$} & 21{\scriptsize$\,\pm3$} & \rev{\textbf{39}{\scriptsize$\,\pm6$}} \\
\midrule
\nolinkurl{cube-single} & \rev{4{\scriptsize$\,\pm2$}} & 11{\scriptsize$\,\pm2$} & \rev{58{\scriptsize$\,\pm8$}} & \rev{30{\scriptsize$\,\pm7$}} & \rev{28{\scriptsize$\,\pm14$}} & \rev{\textbf{86}{\scriptsize$\,\pm6$}} & 18{\scriptsize$\,\pm10$} & 45{\scriptsize$\,\pm4$} & 4{\scriptsize$\,\pm1$} & \rev{45{\scriptsize$\,\pm3$}} & 5{\scriptsize$\,\pm2$} & \rev{5{\scriptsize$\,\pm4$}} & \rev{56{\scriptsize$\,\pm6$}} & 4{\scriptsize$\,\pm2$} & \textbf{62}{\scriptsize$\,\pm5$} & \rev{33{\scriptsize$\,\pm8$}} \\
\nolinkurl{cube-double} & \rev{1{\scriptsize$\,\pm1$}} & 5{\scriptsize$\,\pm3$} & \rev{10{\scriptsize$\,\pm3$}} & \rev{2{\scriptsize$\,\pm1$}} & \rev{5{\scriptsize$\,\pm1$}} & \rev{\textbf{38}{\scriptsize$\,\pm5$}} & 1{\scriptsize$\,\pm0$} & 30{\scriptsize$\,\pm6$} & 4{\scriptsize$\,\pm1$} & \rev{9{\scriptsize$\,\pm4$}} & 1{\scriptsize$\,\pm1$} & \rev{1{\scriptsize$\,\pm1$}} & \rev{18{\scriptsize$\,\pm9$}} & 1{\scriptsize$\,\pm0$} & \textbf{31}{\scriptsize$\,\pm6$} & \rev{10{\scriptsize$\,\pm2$}} \\
\nolinkurl{cube-triple} & \rev{14{\scriptsize$\,\pm2$}} & 19{\scriptsize$\,\pm0$} & \rev{16{\scriptsize$\,\pm2$}} & \rev{16{\scriptsize$\,\pm2$}} & \rev{16{\scriptsize$\,\pm1$}} & \rev{21{\scriptsize$\,\pm1$}} & 13{\scriptsize$\,\pm3$} & \textbf{22}{\scriptsize$\,\pm1$} & 16{\scriptsize$\,\pm1$} & \rev{16{\scriptsize$\,\pm1$}} & 12{\scriptsize$\,\pm6$} & \rev{12{\scriptsize$\,\pm5$}} & \rev{17{\scriptsize$\,\pm2$}} & 11{\scriptsize$\,\pm4$} & \textbf{20}{\scriptsize$\,\pm0$} & \rev{10{\scriptsize$\,\pm4$}} \\
\nolinkurl{cube-quadruple} & \rev{5{\scriptsize$\,\pm1$}} & 11{\scriptsize$\,\pm3$} & \rev{0{\scriptsize$\,\pm0$}} & \rev{7{\scriptsize$\,\pm1$}} & \rev{4{\scriptsize$\,\pm1$}} & \rev{11{\scriptsize$\,\pm3$}} & 4{\scriptsize$\,\pm1$} & \textbf{16}{\scriptsize$\,\pm1$} & 10{\scriptsize$\,\pm3$} & \rev{0{\scriptsize$\,\pm0$}} & 3{\scriptsize$\,\pm1$} & \rev{4{\scriptsize$\,\pm1$}} & \rev{4{\scriptsize$\,\pm2$}} & 4{\scriptsize$\,\pm2$} & \textbf{14}{\scriptsize$\,\pm3$} & \rev{1{\scriptsize$\,\pm1$}} \\
\midrule
\nolinkurl{scene} & \rev{13{\scriptsize$\,\pm2$}} & 20{\scriptsize$\,\pm3$} & \rev{25{\scriptsize$\,\pm4$}} & \rev{12{\scriptsize$\,\pm2$}} & \rev{14{\scriptsize$\,\pm2$}} & \rev{43{\scriptsize$\,\pm6$}} & 13{\scriptsize$\,\pm3$} & \textbf{45}{\scriptsize$\,\pm1$} & 9{\scriptsize$\,\pm2$} & \rev{17{\scriptsize$\,\pm2$}} & 6{\scriptsize$\,\pm1$} & \rev{7{\scriptsize$\,\pm1$}} & \rev{22{\scriptsize$\,\pm7$}} & 6{\scriptsize$\,\pm4$} & \textbf{29}{\scriptsize$\,\pm4$} & \rev{16{\scriptsize$\,\pm3$}} \\
\bottomrule
\end{tabular}%
}
\vspace{-0.8\baselineskip}
\end{table}

\rev{Table~\ref{tab:main} compares goal-conditioned offline RL methods in the learned latent action space with $5\%$ labels, with a $100\%$-label oracle as a reference; results without LAM pretraining are in Appendix~\ref{app:wolam}. LAPIS is best on all five manipulation tasks and, compared with LAPIS trained on real actions with $100\%$ labels, loses little performance on every task except \texttt{scene}. CRL maximizes the critic with a unimodal BC-regularized actor, so it can leave the observed latent modes for unrealizable high-value latents or average separated modes; LAPIS instead filters observed state--latent transitions and fits their multimodal distribution with a flow policy. On \texttt{antmaze}, LAM+HIQL exceeds LAPIS, as long-horizon navigation favors less conservative extraction than the LAPIS gate (Appendix~\ref{app:fully-supervised-antmaze}). With the same critic, IDQL-style selection (LAPIS$^\dagger$) closes this gap on all four \texttt{antmaze} tasks.}

\begin{wraptable}{r}{0.5\textwidth}
\centering
\vspace{-1.2\baselineskip}

\setlength{\tabcolsep}{3pt}
\caption{Normalized score (\%) on visual D4RL Kitchen with $100\%$ and $5\%$ action labels.}
\label{tab:kitchen}
\vspace{-0.75\baselineskip}
\resizebox{\linewidth}{!}{%
\begin{tabular}{l >{\columncolor{oraclegray}}c >{\columncolor{oraclegray}}c >{\columncolor{oraclegray}}c ccc}
\toprule
& \multicolumn{3}{>{\columncolor{oraclegray}}c}{$100\%$ labels (oracle)}
& \multicolumn{3}{c}{$5\%$ labels (w/ LAM)} \\
\cmidrule(lr){2-4}\cmidrule(lr){5-7}
\texttt{Task} & \texttt{GCIQL} & \texttt{CRL} & \texttt{LAPIS} & \textcolor{lamcolor}{\texttt{GCIQL}} & \textcolor{lamcolor}{\texttt{CRL}} & \textcolor{lamcolor}{\texttt{LAPIS}} \\
\midrule
\texttt{partial} & 21{\scriptsize$\,\pm9$} & 28{\scriptsize$\,\pm4$} & \textbf{60}{\scriptsize$\,\pm4$} & 19{\scriptsize$\,\pm9$} & 26{\scriptsize$\,\pm12$} & \textbf{44}{\scriptsize$\,\pm6$} \\
\texttt{mixed}   & 22{\scriptsize$\,\pm3$} & 31{\scriptsize$\,\pm9$} & \textbf{53}{\scriptsize$\,\pm7$} & 21{\scriptsize$\,\pm9$} & 25{\scriptsize$\,\pm12$} & \textbf{41}{\scriptsize$\,\pm5$} \\
\bottomrule
\end{tabular}%
}
\vspace{-1.5\baselineskip}
\end{wraptable}
\paragraph{Beyond OGBench: visual Franka Kitchen}
We further evaluate on visual D4RL Franka Kitchen, a multi-subtask manipulation benchmark. Table~\ref{tab:kitchen} reports the percentage of required subtasks completed on the \texttt{partial} and \texttt{mixed} target-task sets; details are provided in Appendix~\ref{app:data-evaluation}. LAPIS also consistently outperforms GCIQL and CRL on both target-task sets with $100\%$ and $5\%$ action labels. 

\subsection{Comparison under limited action supervision (Q2)}\label{sec:exp-semisup}
\vspace{-0.4\baselineskip}
\begin{table}[t]
\centering

\setlength{\tabcolsep}{5pt}
\caption{\rev{Success rate (\%) on the OGBench visual suite with $5\%$ and $1\%$ action labels.} We compare LAPIS with goal-conditioned methods adapted to learning under limited physical-action supervision. \rev{With $1\%$ labels, the LAM, decoder, and policy of LAPIS are all trained with $1\%$ labels, unlike the decoder study in Table~\ref{tab:decoder-cond}.}}
\label{tab:semisup}
\vspace{-0.65\baselineskip}
\resizebox{\textwidth}{!}{%
\begin{tabular}{l ccccc ccccc}
\toprule
& \multicolumn{5}{c}{$5\%$ labels} & \multicolumn{5}{c}{$1\%$ labels} \\
\cmidrule(lr){2-6}\cmidrule(lr){7-11}
\texttt{Task} & \texttt{SS-ORL} & \texttt{VPT-RL} & \rev{\texttt{GCIVL}} & \texttt{HIQL} & \textcolor{lamcolor}{\texttt{LAPIS}} & \texttt{SS-ORL} & \texttt{VPT-RL} & \rev{\texttt{GCIVL}} & \texttt{HIQL} & \textcolor{lamcolor}{\texttt{LAPIS}} \\
\midrule
\rev{\nolinkurl{antmaze-medium}} & \rev{23{\scriptsize$\,\pm3$}} & \rev{11{\scriptsize$\,\pm2$}} & \rev{2{\scriptsize$\,\pm1$}} & \rev{42{\scriptsize$\,\pm38$}} & \rev{\textbf{49}{\scriptsize$\,\pm3$}} & \rev{0{\scriptsize$\,\pm0$}} & \rev{0{\scriptsize$\,\pm1$}} & \rev{0{\scriptsize$\,\pm0$}} & \rev{0{\scriptsize$\,\pm0$}} & \rev{\textbf{29}{\scriptsize$\,\pm6$}} \\
\rev{\nolinkurl{antmaze-large}} & \rev{6{\scriptsize$\,\pm1$}} & \rev{4{\scriptsize$\,\pm2$}} & \rev{0{\scriptsize$\,\pm1$}} & \rev{7{\scriptsize$\,\pm13$}} & \rev{\textbf{10}{\scriptsize$\,\pm2$}} & \rev{0{\scriptsize$\,\pm0$}} & \rev{0{\scriptsize$\,\pm0$}} & \rev{0{\scriptsize$\,\pm0$}} & \rev{0{\scriptsize$\,\pm0$}} & \rev{\textbf{7}{\scriptsize$\,\pm3$}} \\
\rev{\nolinkurl{antmaze-giant}} & \rev{\textbf{1}{\scriptsize$\,\pm1$}} & \rev{0{\scriptsize$\,\pm0$}} & \rev{0{\scriptsize$\,\pm0$}} & \rev{0{\scriptsize$\,\pm0$}} & \rev{0{\scriptsize$\,\pm0$}} & \rev{0{\scriptsize$\,\pm0$}} & \rev{0{\scriptsize$\,\pm0$}} & \rev{0{\scriptsize$\,\pm0$}} & \rev{0{\scriptsize$\,\pm0$}} & \rev{0{\scriptsize$\,\pm0$}} \\
\rev{\nolinkurl{antmaze-teleport}} & \rev{13{\scriptsize$\,\pm1$}} & \rev{8{\scriptsize$\,\pm1$}} & \rev{3{\scriptsize$\,\pm3$}} & \rev{12{\scriptsize$\,\pm12$}} & \rev{\textbf{21}{\scriptsize$\,\pm3$}} & \rev{4{\scriptsize$\,\pm4$}} & \rev{3{\scriptsize$\,\pm2$}} & \rev{0{\scriptsize$\,\pm0$}} & \rev{2{\scriptsize$\,\pm2$}} & \rev{\textbf{11}{\scriptsize$\,\pm2$}} \\
\midrule
\nolinkurl{cube-single} & 55{\scriptsize$\,\pm3$} & 48{\scriptsize$\,\pm6$} & 0{\scriptsize$\,\pm0$} & 24{\scriptsize$\,\pm8$} & \textbf{62}{\scriptsize$\,\pm5$} & 0{\scriptsize$\,\pm0$} & 0{\scriptsize$\,\pm0$} & 0{\scriptsize$\,\pm0$} & 2{\scriptsize$\,\pm4$} & \textbf{50}{\scriptsize$\,\pm6$} \\
\nolinkurl{cube-double} & 11{\scriptsize$\,\pm6$} & 17{\scriptsize$\,\pm13$} & 0{\scriptsize$\,\pm0$} & 5{\scriptsize$\,\pm2$} & \textbf{31}{\scriptsize$\,\pm6$} & 0{\scriptsize$\,\pm0$} & 0{\scriptsize$\,\pm0$} & 0{\scriptsize$\,\pm0$} & 0{\scriptsize$\,\pm0$} & \textbf{20}{\scriptsize$\,\pm8$} \\
\nolinkurl{cube-triple} & 9{\scriptsize$\,\pm5$} & 15{\scriptsize$\,\pm9$} & 0{\scriptsize$\,\pm0$} & 13{\scriptsize$\,\pm4$} & \textbf{20}{\scriptsize$\,\pm0$} & 0{\scriptsize$\,\pm0$} & 0{\scriptsize$\,\pm0$} & 0{\scriptsize$\,\pm0$} & 1{\scriptsize$\,\pm1$} & \textbf{17}{\scriptsize$\,\pm1$} \\
\nolinkurl{cube-quadruple} & 0{\scriptsize$\,\pm0$} & 2{\scriptsize$\,\pm3$} & 0{\scriptsize$\,\pm0$} & 6{\scriptsize$\,\pm1$} & \textbf{14}{\scriptsize$\,\pm3$} & 0{\scriptsize$\,\pm0$} & 0{\scriptsize$\,\pm0$} & 0{\scriptsize$\,\pm0$} & 0{\scriptsize$\,\pm0$} & \textbf{10}{\scriptsize$\,\pm3$} \\
\midrule
\rev{\nolinkurl{scene}} & \rev{0{\scriptsize$\,\pm0$}} & \rev{0{\scriptsize$\,\pm0$}} & \rev{0{\scriptsize$\,\pm0$}} & \rev{5{\scriptsize$\,\pm1$}} & \rev{\textbf{29}{\scriptsize$\,\pm4$}} & \rev{0{\scriptsize$\,\pm0$}} & \rev{0{\scriptsize$\,\pm0$}} & \rev{0{\scriptsize$\,\pm0$}} & \rev{0{\scriptsize$\,\pm0$}} & \rev{\textbf{9}{\scriptsize$\,\pm3$}} \\
\bottomrule
\end{tabular}%
}
\end{table}

We next compare LAPIS with goal-conditioned baselines adapted to learning under limited physical-action supervision. These adaptations use hindsight goal relabeling, with implementation details provided in Appendix~\ref{app:baseline-details}. \rev{Apart from \texttt{antmaze-giant}, where every method stays near zero at both label budgets, LAPIS is the strongest method with $5\%$ labels, and with $1\%$ labels it still succeeds while the competing methods largely collapse (Table~\ref{tab:semisup}).} These results show that LAPIS can effectively exploit action-free trajectories even when physical action supervision is extremely limited.

\subsection{Effect of supported policy improvement (Q3)}\label{sec:exp-diag}
\begin{figure}[t]
\centering
\includegraphics[width=0.92\textwidth]{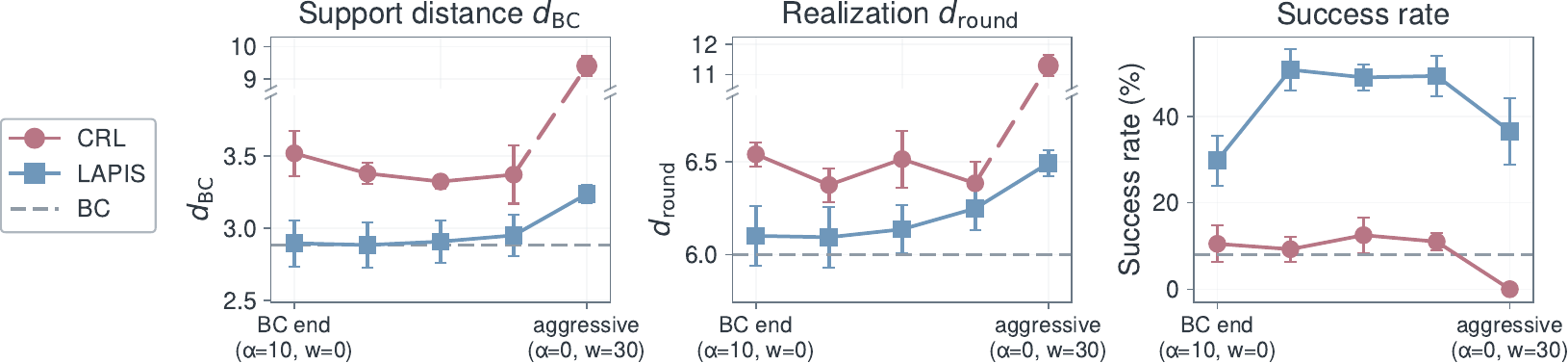}

\caption{Support--success diagnostic (\texttt{antmaze}). We sweep CRL's BC coefficient $\alpha$ and LAPIS guidance $w$, measuring support distance $d_{\mathrm{BC}}$, realization round-trip error $d_{\mathrm{round}}$, and task success.}
\label{fig:diagnostic}

\end{figure}

\vspace{-0.4\baselineskip}
We examine whether successful latent policy improvement remains close to the behavior-supported latent region. Starting from behavior-like policies, we strengthen critic-driven improvement by decreasing CRL's BC coefficient $\alpha$ and increasing LAPIS guidance $w$. Figure~\ref{fig:diagnostic} measures $d_{\mathrm{BC}}$, the distance from a policy latent to the nearest latent sampled from the BC policy at the same state--goal pair, and $d_{\mathrm{round}}$, the round-trip latent error obtained by decoding a selected latent into physical control, executing it in the environment, and measuring the distance between the original latent and the latent re-encoded from the resulting transition; lower values indicate more faithful physical realization. As CRL becomes more aggressive, both distances increase and task success collapses, whereas LAPIS remains closer to the behavior-supported region.

\vspace{-0.4\baselineskip}
\Needspace{4\baselineskip}
\subsection{Which components drive LAPIS? (Q4)}\label{sec:exp-ablation}
\vspace{-0.3\baselineskip}
\begin{table}[t]
\centering
\caption{\rev{Success rate (\%) with $5\%$ action labels (mean $\pm$ std over four seeds); bold marks the best value in each column of each panel. (a)} Improvement $\times$ actor\rev{. (b) Critic $\times$ extraction rule with the same LAMs and decoders.}}
\label{tab:ablation}
\vspace{-0.5\baselineskip}
\footnotesize
\setlength{\tabcolsep}{4pt}
\rev{\textbf{(a)} Improvement $\times$ actor}\par\vspace{2pt}
\begin{tabular}{ll ccc}
\toprule
Improvement & Actor & \rev{\texttt{antmaze-medium}} & \rev{\texttt{cube-single}} & \rev{\texttt{scene}} \\
\midrule
CRL (max-$Q$) & det  & 11{\scriptsize$\,\pm2$} & 4{\scriptsize$\,\pm2$} & 6{\scriptsize$\,\pm4$} \\
CRL (max-$Q$) & flow & 5{\scriptsize$\,\pm2$} & 5{\scriptsize$\,\pm3$} & 7{\scriptsize$\,\pm4$} \\
LAPIS (gate)  & det  & 43{\scriptsize$\,\pm4$} & 36{\scriptsize$\,\pm6$} & 16{\scriptsize$\,\pm4$} \\
LAPIS (gate)  & flow & \textbf{49}{\scriptsize$\,\pm3$} & \textbf{62}{\scriptsize$\,\pm5$} & \textbf{29}{\scriptsize$\,\pm4$} \\
\bottomrule
\end{tabular}

\vspace{5pt}
\begin{revblock}
\textbf{(b)} Critic $\times$ extraction rule\par\vspace{2pt}
\resizebox{\textwidth}{!}{%
\begin{tabular}{llcccc}
\toprule
Critic & Extraction rule & \texttt{antmaze-medium} & \texttt{cube-single} & \texttt{cube-double} & \texttt{scene} \\
\midrule
TD-InfoNCE & Gate + flow (LAPIS) & 49{\scriptsize$\,\pm3$} & \textbf{62}{\scriptsize$\,\pm5$} & \textbf{31}{\scriptsize$\,\pm6$} & \textbf{29}{\scriptsize$\,\pm4$} \\
TD-InfoNCE & AWR-weighted flow & 57{\scriptsize$\,\pm12$} & 54{\scriptsize$\,\pm6$} & 25{\scriptsize$\,\pm4$} & 20{\scriptsize$\,\pm4$} \\
TD-InfoNCE & IDQL-style selection & \textbf{89}{\scriptsize$\,\pm4$} & 33{\scriptsize$\,\pm8$} & 10{\scriptsize$\,\pm2$} & 16{\scriptsize$\,\pm3$} \\
\midrule
IQL & Gate + flow & 32{\scriptsize$\,\pm4$} & 42{\scriptsize$\,\pm14$} & 12{\scriptsize$\,\pm4$} & 20{\scriptsize$\,\pm3$} \\
IQL & AWR-weighted flow & 18{\scriptsize$\,\pm4$} & 55{\scriptsize$\,\pm8$} & 13{\scriptsize$\,\pm4$} & 19{\scriptsize$\,\pm4$} \\
IQL & IDQL-style selection & 66{\scriptsize$\,\pm5$} & 28{\scriptsize$\,\pm8$} & 5{\scriptsize$\,\pm2$} & 5{\scriptsize$\,\pm2$} \\
IQL & DDPG+BC flow & 38{\scriptsize$\,\pm7$} & 6{\scriptsize$\,\pm4$} & 5{\scriptsize$\,\pm1$} & 9{\scriptsize$\,\pm1$} \\
\bottomrule
\end{tabular}}
\end{revblock}
\end{table}
LAPIS combines critic-guided filtering with a multimodal flow actor. To isolate their contributions, we perform a $2\times2$ ablation crossing the policy-improvement mechanism $\{$max-$Q$, gate$\}$ and actor parameterization $\{$deterministic, flow$\}$, while keeping the LAM, decoder, critic, and training budget fixed (Table~\ref{tab:ablation}\rev{(a)}). Critic-guided filtering consistently improves success with either actor parameterization, while the flow actor provides a further gain when combined with the gate.

\rev{\textbf{Extraction rule versus critic.} Table~\ref{tab:ablation}(b) crosses two critics (TD-InfoNCE, IQL) with four extraction rules (gate, AWR~\citep{peng2019awr}, IDQL-style selection~\citep{hansenestruch2023idql}, DDPG+BC~\citep{park2025fql}; Appendix~\ref{app:extraction-grid}) at fixed LAM, decoder, and labels. The critic matters most: under the same rule, switching to the IQL critic lowers success in $10$ of the $12$ comparable cells and leaves the other two similar. With the TD-InfoNCE critic, the gate is best on manipulation, whereas IDQL-style selection is best on \texttt{antmaze-medium} (LAPIS$^\dagger$ in Table~\ref{tab:main}). DDPG+BC, which maximizes the critic directly, is among the lowest on manipulation, consistent with the max-$Q$ rows of Table~\ref{tab:ablation}(a).}
\vspace{-0.4\baselineskip}
\Needspace{13\baselineskip}
\subsection{Effect of decoder conditioning (Q5)}\label{sec:exp-decoder}
\vspace{-0.3\baselineskip}
\begin{wraptable}{r}{0.46\textwidth}
\centering
\vspace{-1.1\baselineskip}

\setlength{\tabcolsep}{3pt}
\caption{Task success (\%) across decoder types and action-label budgets.}
\label{tab:decoder-cond}
\vspace{-4pt}
\resizebox{0.46\textwidth}{!}{%
\begin{tabular}{ll >{\columncolor{oraclegray}}c cc}
\toprule
Env & Decoder & $100\%$ labels & $5\%$ labels & $1\%$ labels \\
\midrule
\multirow{2}{*}{\texttt{cube}} & $\Psi(a\mid z)$ & 54{\scriptsize$\,\pm2$} & \textbf{62}{\scriptsize$\,\pm5$} & \textbf{50}{\scriptsize$\,\pm6$} \\
 & $\Psi(a\mid s,z)$ & \textbf{74}{\scriptsize$\,\pm4$} & 55{\scriptsize$\,\pm6$} & 46{\scriptsize$\,\pm3$} \\
\midrule
\multirow{2}{*}{\texttt{scene}} & $\Psi(a\mid z)$ & 25{\scriptsize$\,\pm4$} & \textbf{29}{\scriptsize$\,\pm4$} & \textbf{24}{\scriptsize$\,\pm5$} \\
 & $\Psi(a\mid s,z)$ & \textbf{38}{\scriptsize$\,\pm6$} & 20{\scriptsize$\,\pm5$} & 16{\scriptsize$\,\pm3$} \\
\midrule
\multirow{2}{*}{\texttt{antmaze}} & $\Psi(a\mid z)$ & 48{\scriptsize$\,\pm6$} & \textbf{49}{\scriptsize$\,\pm3$} & \textbf{44}{\scriptsize$\,\pm5$} \\
 & $\Psi(a\mid s,z)$ & \textbf{52}{\scriptsize$\,\pm4$} & 27{\scriptsize$\,\pm2$} & 38{\scriptsize$\,\pm4$} \\
\bottomrule
\end{tabular}%
}
\vspace{-1.2\baselineskip}
\end{wraptable}

We compare state-marginal $\Psi(a\mid z)$ and state-conditioned $\Psi(a\mid s,z)$ decoders across label budgets, keeping the $5\%$-label LAM fixed and varying only decoder supervision (Table~\ref{tab:decoder-cond}). With $5\%$ and $1\%$ action labels, the marginal decoder achieves higher success rates across all three domains. With full action supervision, the state-conditioned decoder performs better in every domain. \rev{Conditioning on the state closes the transfer gap in theory (Theorem~\ref{thm:eval}), but in practice it pays off only when labels are plentiful. LAPO~\citep[Sec.~6.2]{schmidt2024lapo} attributes the plateau of latent-only decoders to state dependence; with scarce labels, however, a state-conditioned decoder does not help, so LAPIS uses $\Psi(a\mid z)$.}

\subsection{\texorpdfstring{\rev{Does the score offset change the gate? (Q6)}}{Does the score offset change the gate? (Q6)}}\label{sec:exp-offset}
\vspace{-0.3\baselineskip}
\begin{wraptable}{r}{0.42\textwidth}
\centering
\vspace{-1.1\baselineskip}

\setlength{\tabcolsep}{3pt}
\caption{\rev{$f$ versus $\bar f$ in trained critics.}}
\label{tab:score-offset}
\vspace{-4pt}
\begin{revblock}
\small
\begin{tabular}{lccc}
\toprule
 & \texttt{antmaze} & \texttt{cube} & \texttt{scene} \\
\midrule
Spearman & $0.79$ & $0.86$ & $0.87$ \\
Gate agreement & $87\%$ & $91\%$ & $91\%$ \\
\bottomrule
\end{tabular}
\end{revblock}
\vspace{-1.5\baselineskip}
\end{wraptable}

\leavevmode\rev{The LAPIS gate (Equation~\eqref{eq:gate}) uses the raw TD-InfoNCE score $f$, which is determined only up to an offset $c(s,z)$. The offset-free score $\bar f$ removes this offset but requires a log-partition estimate (Appendix~\ref{app:critic-ordering}). To see whether the offset matters in practice, we compare the two scores on candidate latents in trained LAPIS critics (Table~\ref{tab:score-offset}). They rank the candidates similarly, and gates on the two scores make the same keep/drop decision for about $90\%$ of transitions. The raw score is thus a practical surrogate for $\bar f$.}
\vspace{-0.3\baselineskip}
\section{Conclusion}\label{sec:disc}
\vspace{-0.3\baselineskip}
We study offline policy improvement in latent action spaces learned from action-free data. We show that state-conditioned realizability, non-convex realizable sets, and outcome-based abstraction can make standard offline RL unreliable, and characterize when latent improvement transfers to physical execution. In goal-conditioned RL, LAPIS performs support-aware policy improvement through critic-guided filtering over observed state--latent transitions and a multimodal flow policy. \rev{With only $5\%$ action labels, LAPIS recovers much of the performance of its fully supervised counterpart and outperforms baselines on most tasks, especially in manipulation.}

\IfFileExists{references.bib}{%
\IfFileExists{iclr2027_conference.bst}{\bibliographystyle{iclr2027_conference}}{\bibliographystyle{plainnat}}%
\bibliography{references}%

\begin{thebibliography}{43}
\providecommand{\natexlab}[1]{#1}
\providecommand{\url}[1]{\texttt{#1}}
\expandafter\ifx\csname urlstyle\endcsname\relax
  \providecommand{\doi}[1]{doi: #1}\else
  \providecommand{\doi}{doi: \begingroup \urlstyle{rm}\Url}\fi

\bibitem[Ajay et~al.(2021)Ajay, Kumar, Agrawal, Levine, and
  Nachum]{singh2020opal}
Anurag Ajay, Aviral Kumar, Pulkit Agrawal, Sergey Levine, and Ofir Nachum.
\newblock {OPAL}: Offline primitive discovery for accelerating offline
  reinforcement learning.
\newblock In \emph{International Conference on Learning Representations
  (ICLR)}, 2021.

\bibitem[Asadi et~al.(2018)Asadi, Misra, and Littman]{asadi2018lipschitz}
Kavosh Asadi, Dipendra Misra, and Michael~L. Littman.
\newblock Lipschitz continuity in model-based reinforcement learning.
\newblock In \emph{International Conference on Machine Learning (ICML)}, 2018.

\bibitem[Baker et~al.(2022)Baker, Akkaya, Zhokov, Huizinga, Tang, Ecoffet,
  Houghton, Sampedro, and Clune]{baker2022vpt}
Bowen Baker, Ilge Akkaya, Peter Zhokov, Joost Huizinga, Jie Tang, Adrien
  Ecoffet, Brandon Houghton, Raul Sampedro, and Jeff Clune.
\newblock Video {PreTraining} ({VPT}): Learning to act by watching unlabeled
  online videos.
\newblock In \emph{Advances in Neural Information Processing Systems
  (NeurIPS)}, 2022.

\bibitem[Bruce et~al.(2024)Bruce, Dennis, Edwards, Parker-Holder, Shi, Hughes,
  Lai, Mavalankar, Steigerwald, Apps, Aytar, Bechtle, Behbahani, Chan, Heess,
  Gonzalez, Osindero, Ozair, Reed, Zhang, Zolna, Clune, de~Freitas, Singh, and
  Rockt{\"a}schel]{bruce2024genie}
Jake Bruce, Michael~D Dennis, Ashley Edwards, Jack Parker-Holder, Yuge Shi,
  Edward Hughes, Matthew Lai, Aditi Mavalankar, Richie Steigerwald, Chris Apps,
  Yusuf Aytar, Sarah Bechtle, Feryal Behbahani, Stephanie Chan, Nicolas Heess,
  Lucy Gonzalez, Simon Osindero, Sherjil Ozair, Scott Reed, Jingwei Zhang,
  Konrad Zolna, Jeff Clune, Nando de~Freitas, Satinder Singh, and Tim
  Rockt{\"a}schel.
\newblock Genie: Generative interactive environments.
\newblock In \emph{International Conference on Machine Learning (ICML)}, 2024.

\bibitem[Cachet et~al.(2024)Cachet, Dance, and Sigaud]{cachet2024language}
Th\'{e}o Cachet, Christopher~R. Dance, and Olivier Sigaud.
\newblock Bridging environments and language with rendering functions and
  vision-language models.
\newblock In \emph{International Conference on Machine Learning (ICML)}, volume
  235 of \emph{Proceedings of Machine Learning Research}, pp.\  5135--5188,
  2024.

\bibitem[Chang et~al.(2022)Chang, Gupta, and Gupta]{chang2022laq}
Matthew Chang, Arjun Gupta, and Saurabh Gupta.
\newblock Learning value functions from undirected state-only experience.
\newblock In \emph{International Conference on Learning Representations
  (ICLR)}, 2022.

\bibitem[Chen et~al.(2024)Chen, Guo, He, Zhang, Zhang, Yang, Zhao, and
  Bian]{chen2024igor}
Xiaoyu Chen, Junliang Guo, Tianyu He, Chuheng Zhang, Pushi Zhang, Derek~Cathera
  Yang, Li~Zhao, and Jiang Bian.
\newblock {IGOR}: Image-{GO}al representations are the atomic control units for
  foundation models in embodied {AI}.
\newblock \emph{arXiv preprint arXiv:2411.00785}, 2024.

\bibitem[Chen et~al.(2025)Chen, Wei, Zhang, Zhang, Wang, Guo, Yang, Wang, Xiao,
  Zhao, Chen, and Bian]{chen2025villax}
Xiaoyu Chen, Hangxing Wei, Pushi Zhang, Chuheng Zhang, Kaixin Wang, Yanjiang
  Guo, Rushuai Yang, Yucen Wang, Xinquan Xiao, Li~Zhao, Jianyu Chen, and Jiang
  Bian.
\newblock {villa-X}: Enhancing latent action modeling in vision-language-action
  models.
\newblock \emph{arXiv preprint arXiv:2507.23682}, 2025.

\bibitem[Edwards et~al.(2019)Edwards, Sahni, Schroecker, and
  Isbell]{edwards2019ilpo}
Ashley~D Edwards, Himanshu Sahni, Yannick Schroecker, and Charles~L Isbell.
\newblock Imitating latent policies from observation.
\newblock In \emph{International Conference on Machine Learning (ICML)}, 2019.

\bibitem[Eysenbach et~al.(2022)Eysenbach, Zhang, Levine, and
  Salakhutdinov]{eysenbach2022crl}
Benjamin Eysenbach, Tianjun Zhang, Sergey Levine, and Ruslan Salakhutdinov.
\newblock Contrastive learning as goal-conditioned reinforcement learning.
\newblock In \emph{Advances in Neural Information Processing Systems
  (NeurIPS)}, 2022.

\bibitem[Frans et~al.(2025)Frans, Park, Abbeel, and Levine]{frans2025cfgrl}
Kevin Frans, Seohong Park, Pieter Abbeel, and Sergey Levine.
\newblock Diffusion guidance is a controllable policy improvement operator.
\newblock \emph{arXiv preprint arXiv:2505.23458}, 2025.

\bibitem[Fu et~al.(2020)Fu, Kumar, Nachum, Tucker, and Levine]{fu2020d4rl}
Justin Fu, Aviral Kumar, Ofir Nachum, George Tucker, and Sergey Levine.
\newblock {D4RL}: Datasets for deep data-driven reinforcement learning.
\newblock \emph{arXiv preprint arXiv:2004.07219}, 2020.

\bibitem[Fujimoto et~al.(2019)Fujimoto, Meger, and Precup]{fujimoto2019bcq}
Scott Fujimoto, David Meger, and Doina Precup.
\newblock Off-policy deep reinforcement learning without exploration.
\newblock In \emph{International Conference on Machine Learning (ICML)}, 2019.

\bibitem[Gao et~al.(2025)Gao, Zhou, Du, Zhang, and Gan]{gao2025adaworld}
Shenyuan Gao, Siyuan Zhou, Yilun Du, Jun Zhang, and Chuang Gan.
\newblock {AdaWorld}: Learning adaptable world models with latent actions.
\newblock In \emph{International Conference on Machine Learning (ICML)}, 2025.

\bibitem[Gelada et~al.(2019)Gelada, Kumar, Buckman, Nachum, and
  Bellemare]{gelada2019deepmdp}
Carles Gelada, Saurabh Kumar, Jacob Buckman, Ofir Nachum, and Marc~G.
  Bellemare.
\newblock {DeepMDP}: Learning continuous latent space models for representation
  learning.
\newblock In \emph{International Conference on Machine Learning (ICML)}, 2019.

\bibitem[Ghosh et~al.(2023)Ghosh, Bhateja, and Levine]{ghosh2023icvf}
Dibya Ghosh, Chethan~Anand Bhateja, and Sergey Levine.
\newblock Reinforcement learning from passive data via latent intentions.
\newblock In \emph{International Conference on Machine Learning (ICML)}, 2023.

\bibitem[{Google DeepMind}(2024)]{genie2}
{Google DeepMind}.
\newblock Genie 2: A large-scale foundation world model.
\newblock Technical report, Google DeepMind, 2024.

\bibitem[Gupta et~al.(2020)Gupta, Kumar, Lynch, Levine, and
  Hausman]{gupta2020relay}
Abhishek Gupta, Vikash Kumar, Corey Lynch, Sergey Levine, and Karol Hausman.
\newblock Relay policy learning: Solving long-horizon tasks via imitation and
  reinforcement learning.
\newblock In \emph{Proceedings of the Conference on Robot Learning}, volume 100
  of \emph{Proceedings of Machine Learning Research}, pp.\  1025--1037. PMLR,
  2020.

\bibitem[Hansen-Estruch et~al.(2023)Hansen-Estruch, Kostrikov, Janner, Kuba,
  and Levine]{hansenestruch2023idql}
Philippe Hansen-Estruch, Ilya Kostrikov, Michael Janner, Jakub~Grudzien Kuba,
  and Sergey Levine.
\newblock {IDQL}: Implicit {Q}-learning as an actor-critic method with
  diffusion policies.
\newblock \emph{arXiv preprint arXiv:2304.10573}, 2023.

\bibitem[Ho \& Salimans(2022)Ho and Salimans]{ho2022cfg}
Jonathan Ho and Tim Salimans.
\newblock Classifier-free diffusion guidance.
\newblock \emph{arXiv preprint arXiv:2207.12598}, 2022.

\bibitem[Ki et~al.(2025)Ki, Oh, Shim, and Lee]{ki2025prior}
Donghyeon Ki, JunHyeok Oh, Seong-Woong Shim, and Byung-Jun Lee.
\newblock Prior-guided diffusion planning for offline reinforcement learning.
\newblock In \emph{Advances in Neural Information Processing Systems
  (NeurIPS)}, 2025.

\bibitem[Kostrikov et~al.(2022)Kostrikov, Nair, and Levine]{kostrikov2022iql}
Ilya Kostrikov, Ashvin Nair, and Sergey Levine.
\newblock Offline reinforcement learning with implicit {Q}-learning.
\newblock In \emph{International Conference on Learning Representations
  (ICLR)}, 2022.

\bibitem[Levine et~al.(2020)Levine, Kumar, Tucker, and Fu]{levine2020offline}
Sergey Levine, Aviral Kumar, George Tucker, and Justin Fu.
\newblock Offline reinforcement learning: Tutorial, review, and perspectives on
  open problems.
\newblock \emph{arXiv preprint arXiv:2005.01643}, 2020.

\bibitem[Li et~al.(2006)Li, Walsh, and Littman]{li2006unified}
Lihong Li, Thomas~J. Walsh, and Michael~L. Littman.
\newblock Towards a unified theory of state abstraction for {MDP}s.
\newblock In \emph{Proceedings of the International Symposium on Artificial
  Intelligence and Mathematics}, 2006.

\bibitem[Lipman et~al.(2023)Lipman, Chen, Ben-Hamu, Nickel, and
  Le]{lipman2023flow}
Yaron Lipman, Ricky T~Q Chen, Heli Ben-Hamu, Maximilian Nickel, and Matt Le.
\newblock Flow matching for generative modeling.
\newblock In \emph{International Conference on Learning Representations
  (ICLR)}, 2023.

\bibitem[Lu et~al.(2025)Lu, Han, Shen, and Li]{lu2025gooddiffusion}
Haofei Lu, Dongqi Han, Yifei Shen, and Dongsheng Li.
\newblock What makes a good diffusion planner for decision making?
\newblock In \emph{International Conference on Learning Representations
  (ICLR)}, 2025.

\bibitem[Lynch et~al.(2020)Lynch, Khansari, Xiao, Kumar, Tompson, Levine, and
  Sermanet]{lynch2020play}
Corey Lynch, Mohi Khansari, Ted Xiao, Vikash Kumar, Jonathan Tompson, Sergey
  Levine, and Pierre Sermanet.
\newblock Learning latent plans from play.
\newblock In \emph{Conference on Robot Learning (CoRL)}, 2020.

\bibitem[Myers et~al.(2023)Myers, He, Fang, Walke, Hansen-Estruch, Cheng,
  Jalobeanu, Kolobov, Dragan, and Levine]{myers2023goal}
Vivek Myers, Andre He, Kuan Fang, Homer Walke, Philippe Hansen-Estruch,
  Ching-An Cheng, Mihai Jalobeanu, Andrey Kolobov, Anca Dragan, and Sergey
  Levine.
\newblock Goal representations for instruction following: A semi-supervised
  language interface to control.
\newblock In \emph{Conference on Robot Learning (CoRL)}, 2023.

\bibitem[Park et~al.(2023)Park, Ghosh, Eysenbach, and Levine]{park2023hiql}
Seohong Park, Dibya Ghosh, Benjamin Eysenbach, and Sergey Levine.
\newblock {HIQL}: Offline goal-conditioned {RL} with latent states as actions.
\newblock In \emph{Advances in Neural Information Processing Systems
  (NeurIPS)}, 2023.

\bibitem[Park et~al.(2025{\natexlab{a}})Park, Frans, Eysenbach, and
  Levine]{park2025ogbench}
Seohong Park, Kevin Frans, Benjamin Eysenbach, and Sergey Levine.
\newblock {OGB}ench: Benchmarking offline goal-conditioned {RL}.
\newblock In \emph{International Conference on Learning Representations
  (ICLR)}, 2025{\natexlab{a}}.

\bibitem[Park et~al.(2025{\natexlab{b}})Park, Li, and Levine]{park2025fql}
Seohong Park, Qiyang Li, and Sergey Levine.
\newblock Flow {Q}-learning.
\newblock In \emph{International Conference on Machine Learning (ICML)},
  2025{\natexlab{b}}.

\bibitem[Peng et~al.(2019)Peng, Kumar, Zhang, and Levine]{peng2019awr}
Xue~Bin Peng, Aviral Kumar, Grace Zhang, and Sergey Levine.
\newblock Advantage-weighted regression: Simple and scalable off-policy
  reinforcement learning.
\newblock \emph{arXiv preprint arXiv:1910.00177}, 2019.

\bibitem[Pertsch et~al.(2020)Pertsch, Lee, and Lim]{pertsch2020spirl}
Karl Pertsch, Youngwoon Lee, and Joseph~J Lim.
\newblock Accelerating reinforcement learning with learned skill priors.
\newblock In \emph{Conference on Robot Learning (CoRL)}, 2020.

\bibitem[Ravindran \& Barto(2003)Ravindran and Barto]{ravindran2003smdp}
Balaraman Ravindran and Andrew~G. Barto.
\newblock {SMDP} homomorphisms: An algebraic approach to abstraction in
  semi-markov decision processes.
\newblock In \emph{Proceedings of the 18th International Joint Conference on
  Artificial Intelligence}, pp.\  1011--1016, 2003.

\bibitem[Rybkin et~al.(2019)Rybkin, Pertsch, Derpanis, Daniilidis, and
  Jaegle]{rybkin2019learning}
Oleh Rybkin, Karl Pertsch, Konstantinos~G Derpanis, Kostas Daniilidis, and
  Andrew Jaegle.
\newblock Learning what you can do before doing anything.
\newblock In \emph{International Conference on Learning Representations
  (ICLR)}, 2019.

\bibitem[Schmidt \& Jiang(2024)Schmidt and Jiang]{schmidt2024lapo}
Dominik Schmidt and Minqi Jiang.
\newblock Learning to act without actions.
\newblock In \emph{International Conference on Learning Representations
  (ICLR)}, 2024.

\bibitem[Torabi et~al.(2018)Torabi, Warnell, and Stone]{torabi2018bco}
Faraz Torabi, Garrett Warnell, and Peter Stone.
\newblock Behavioral cloning from observation.
\newblock In \emph{International Joint Conference on Artificial Intelligence
  (IJCAI)}, 2018.

\bibitem[Wang et~al.(2023)Wang, Torralba, Isola, and Zhang]{wang2023qrl}
Tongzhou Wang, Antonio Torralba, Phillip Isola, and Amy Zhang.
\newblock Optimal goal-reaching reinforcement learning via quasimetric
  learning.
\newblock In \emph{International Conference on Machine Learning (ICML)}, 2023.

\bibitem[Ye et~al.(2025)Ye, Jang, Jeon, Joo, Yang, Peng, Mandlekar, Tan, Chao,
  Lin, Liden, Lee, Gao, Zettlemoyer, Fox, and Seo]{ye2025lapa}
Seonghyeon Ye, Joel Jang, Byeongguk Jeon, Sejune Joo, Jianwei Yang, Baolin
  Peng, Ajay Mandlekar, Reuben Tan, Yu-Wei Chao, Bill~Yuchen Lin, Lars Liden,
  Kimin Lee, Jianfeng Gao, Luke Zettlemoyer, Dieter Fox, and Minjoon Seo.
\newblock Latent action pretraining from videos.
\newblock In \emph{International Conference on Learning Representations
  (ICLR)}, 2025.

\bibitem[Zhang et~al.(2025)Zhang, Pearce, Zhang, Wang, Chen, Shen, Zhao, and
  Bian]{zhang2025whatlams}
Chuheng Zhang, Tim Pearce, Pushi Zhang, Kaixin Wang, Xiaoyu Chen, Wei Shen,
  Li~Zhao, and Jiang Bian.
\newblock What do latent action models actually learn?
\newblock In \emph{Advances in Neural Information Processing Systems
  (NeurIPS)}, 2025.

\bibitem[Zheng et~al.(2024)Zheng, Salakhutdinov, and Eysenbach]{zheng2024cdpc}
Chongyi Zheng, Ruslan Salakhutdinov, and Benjamin Eysenbach.
\newblock Contrastive difference predictive coding.
\newblock In \emph{International Conference on Learning Representations
  (ICLR)}, 2024.

\bibitem[Zheng et~al.(2023)Zheng, Henaff, Amos, and Grover]{zheng2023semiaf}
Qinqing Zheng, Mikael Henaff, Brandon Amos, and Aditya Grover.
\newblock Semi-supervised offline reinforcement learning with action-free
  trajectories.
\newblock In \emph{International Conference on Machine Learning (ICML)}, 2023.

\bibitem[Zhou et~al.(2020)Zhou, Bajracharya, and Held]{zhou2020plas}
Wenxuan Zhou, Sujay Bajracharya, and David Held.
\newblock {PLAS}: Latent action space for offline reinforcement learning.
\newblock In \emph{Conference on Robot Learning (CoRL)}, 2020.

\end{thebibliography}
}{\input{references_recovered.tex}}

\clearpage
\appendix

\clearpage
\newpage
\section{Extended Related Work}
\label{app:related_work}

\subsection{Offline RL in Learned Latent or Skill Spaces}
Prior work has used learned latent or skill representations to constrain policy optimization to behavior-like controls. PLAS \citep{zhou2020plas} learns a continuous latent action space with a conditional VAE and performs offline policy optimization through a behavior-trained decoder. Skill-based methods such as OPAL \citep{singh2020opal}, SPiRL \citep{pertsch2020spirl}, and latent planning approaches \citep{lynch2020play} instead learn temporally extended action representations and optimize high-level policies over them. These approaches typically rely on action-labeled experience to learn the latent control representation or its low-level execution model. In contrast, we study offline policy improvement in a latent action space learned primarily from action-free trajectories, using scarce action labels only to ground latent decisions in physical control.

\subsection{Action-Free and Semi-Supervised Reinforcement Learning}
Learning control from trajectories with missing actions has been studied through imitation from observation \citep{torabi2018bco}, action inference from video \citep{baker2022vpt}, latent-intention learning \citep{ghosh2023icvf}, hierarchical goal-conditioned learning \citep{park2023hiql}, and semi-supervised offline RL \citep{zheng2023semiaf}. These approaches exploit action-free experience through inferred actions, state or goal abstractions, or value learning without complete action supervision. Our setting differs in that policy improvement is performed directly in a learned latent action space and physical actions are needed only for decoding. This makes the relationship between latent-space optimization and physical execution a central part of the problem.

\subsection{Offline Goal-Conditioned and Contrastive RL}

Offline goal-conditioned RL learns multi-goal policies from fixed datasets without additional environment interaction. Contrastive RL \citep{eysenbach2022crl} formulates goal-conditioned control through density-ratio estimation of future-state occupancy, while subsequent work develops related contrastive objectives for offline control \citep{zheng2024cdpc}. LAPIS adopts this goal-conditioned formulation in the learned latent-action space, using future states as goals and a contrastive critic to estimate which observed state--latent transitions are useful for reaching them.

Recent work has also explored expressive generative policies and guidance-based policy extraction \citep{lipman2023flow,ho2022cfg,frans2025cfgrl}. Most closely related, CFGRL \citep{frans2025cfgrl} derives classifier-free guidance as a controllable policy-improvement operator; in its value-free goal-conditioned formulation, guidance combines unconditional and goal-conditioned generative policy branches without an explicit value function. LAPIS shares guided flow-based extraction, but differs in where policy improvement is introduced. LAPIS first estimates goal-conditioned future-state occupancy with a TD-InfoNCE critic and filters only observed state--latent transitions to construct a high-value distribution whose support remains within the observed conditional latent support. The flow policy then models this support-aware target, and classifier-free guidance is applied only at inference.

This distinction is particularly important in learned latent-action spaces. Unlike physical actions, a latent action is a learned transition label whose usefulness depends on state-conditioned realizability and on whether decoding faithfully produces the represented outcome. Thus, improving a generative policy over latents does not by itself imply improved physical execution. Our transfer analysis characterizes when latent-space improvement survives decoding, while the direct comparison in Appendix~\ref{app:additional-ablations} shows that increasing goal guidance can improve physical-action CFGRL but does not provide consistent improvement when the same construction is applied directly in the learned latent space.

\subsection{Relation to Batch-Constrained RL and MDP Abstraction}
\label{app:prior-theory}

Our theoretical analysis is related to classical results in batch-constrained reinforcement learning and MDP abstraction, but addresses a different cross-representation question. Batch-constrained RL studies policy evaluation and improvement when the policy is restricted to actions represented in a fixed offline dataset, thereby avoiding extrapolation to unsupported physical actions \citep{fujimoto2019bcq}. In our setting, however, the policy acts in a \emph{learned} action space: observed physical transitions are relabeled by an inverse model as latent actions, and each latent decision must subsequently be decoded back to physical control. Consequently, restricting the policy to observed latent actions is not sufficient by itself to establish correspondence with physical execution. The latent support is state-dependent, a supported pair $(s,z)$ must identify the transition outcome represented by $z$, and the decoder must realize that outcome through physical actions.

Theorem~1 characterizes this additional structure. We construct physical- and latent-action batch MDPs from the same observed transitions and introduce supported outcome identifiability together with a batch-exact decoder that links a supported latent decision to its physical realizations. Under deterministic dynamics, this construction yields exact equality between the return of a batch-constrained latent policy and that of its decoded physical policy. Thus, the result extends the batch-constrained perspective from policy optimization within a fixed action space to policy improvement through a learned action representation whose semantics and physical execution must themselves be related.

Our optimality analysis is also connected to MDP homomorphism and abstraction theory, which studies when an abstract decision process preserves the value structure of the underlying MDP \citep{ravindran2003smdp,li2006unified}. The latent actions considered here induce an outcome-based action abstraction learned from observed transitions: distinct physical controls can map to the same latent action whenever they produce the same represented outcome. This abstraction can preserve execution while discarding distinctions that matter for reward. Theorem~2 identifies the corresponding condition for optimality alignment: when reward is determined by the transition outcome, an optimal latent policy decodes to an optimal policy within the decoder-covered, batch-constrained physical policy class. The counterexample in Appendix~C.3 shows that this condition is substantive: latent-to-physical return transfer can hold exactly even when latent and physical optima do not align.

Taken together, these results connect batch-constrained and abstraction-based reasoning to learned latent actions, where the representation is inferred from action-free transitions rather than given a priori. The additional questions are therefore not only whether a policy remains within dataset support or whether an abstraction preserves an MDP, but whether a learned latent label identifies a physically executable outcome and whether the distinctions removed by that representation are compatible with the downstream objective.

\clearpage
\newpage
\section{Empirical realizability diagnostics}\label{app:fig1-protocol}

The bottom-row diagnostics in Figure~\ref{fig:properties_concept} evaluate the three structural properties discussed in Section~\ref{sec:properties} on \nolinkurl{visual-cube-single-play-v0}. All latent-space distances are computed in the standardized, quantized latent coordinates used throughout the experiments.

\paragraph{State-conditioned realizability.}
For Figure~\ref{fig:properties_concept}(a), we select latents with known physical realizations at their source states and evaluate whether the same latents can be realized from other query states. For each state--latent pair, realization error is approximated by finite search over physical action chunks. The candidate set combines known realizing chunks, dataset chunks retrieved from nearby states and re-executed at the query state, and multi-start cross-entropy-method (CEM) refinements within the environment's action bounds. Rows correspond to query states and columns to source latents.

\paragraph{Non-convex realizable set.}
For Figure~\ref{fig:properties_concept}(b), we select pairs of separated latents that are both realizable from the same state and linearly interpolate between them. Realization error is evaluated on a uniform interpolation grid. The action chunks realizing both endpoints are included in the candidate set at every interpolation point, together with the same dataset-based and CEM search used above. Confidence bands are computed across latent pairs.

\paragraph{Diagnostic threshold.}
Panels~(a--b) use a common threshold calibrated from two finite-search error samples. Let $D_{\mathrm{self}}$ contain errors for known-realizable source latents searched at their source states without supplying their original realizing chunks, and let $D_{\mathrm{cross}}$ contain errors for those latents searched from different states. Using $16$ anchor states and CEM population $256$ with $5$ iterations, we set
\[
\epsilon=\frac{Q_{0.9}(D_{\mathrm{self}})+Q_{0.1}(D_{\mathrm{cross}})}{2}\approx3.14,
\]
where $Q_p$ denotes the empirical $p$-quantile. We use the rounded value $3.14$ in both panels. 

\paragraph{Outcome-based abstraction.}
For Figure~\ref{fig:properties_concept}(c), we test whether substantially different physical action chunks can realize the same target outcome. For each target, targeted CEM search minimizes outcome distance, with restarts repelled from previously discovered chunks to encourage distinct realizations. Outcome distance is Euclidean in simulator generalized-position coordinates standardized using a candidate rollout pool at each anchor state, while action distance is Euclidean over flattened chunks. A candidate is retained when its outcome distance is at most $\tau_{\mathrm{out}}$ and its distance from every previously retained chunk is at least $\tau_a$. We use $12$ anchor states with $8$ target outcomes each and $20$ CEM restarts per target, all run without early stopping. At each anchor, $\tau_{\mathrm{out}}$ is set to the $1$st percentile of pairwise outcome distances and $\tau_a$ to the $40$th percentile of pairwise action-chunk distances. Each target's observed source chunk is included as the first realization.

\paragraph{Search-budget robustness.}
Because realization error is estimated through finite action-space search, we additionally repeat the search with a substantially larger CEM budget. The default search uses population $256$ and $5$ iterations, while the enlarged search uses population $1024$ and $12$ iterations. We also evaluate a known-realizable floor by searching for source latents without providing their original realizing action chunks. This comparison tests whether the conclusions of panels~(a--b) persist under substantially stronger search, without interpreting failure to find a realization as proof of global unrealizability. Results are summarized in Table~\ref{tab:search-budget-robustness}.

\begin{table}[htbp]
\centering

\setlength{\tabcolsep}{5pt}
\caption{Realization error under different search budgets (lower is better; $\epsilon=3.14$). Interpolation denotes interior points in Figure~\ref{fig:properties_concept}(b), and state swap denotes off-diagonal pairs in Figure~\ref{fig:properties_concept}(a).}
\label{tab:search-budget-robustness}
\begin{tabular}{lccc}
\toprule
Search & Population $\times$ iterations & Interpolation & State swap \\
\midrule
Default & $256\times5$ & $4.02$ & $4.12$ \\
Enlarged & $1024\times12$ & $3.69$ & $3.51$ \\
\bottomrule
\end{tabular}
\end{table}

\newpage
\section{Proofs and theoretical derivations}\label{app:proofs}

\begin{figure}[H]
\centering
\includegraphics[width=\textwidth]{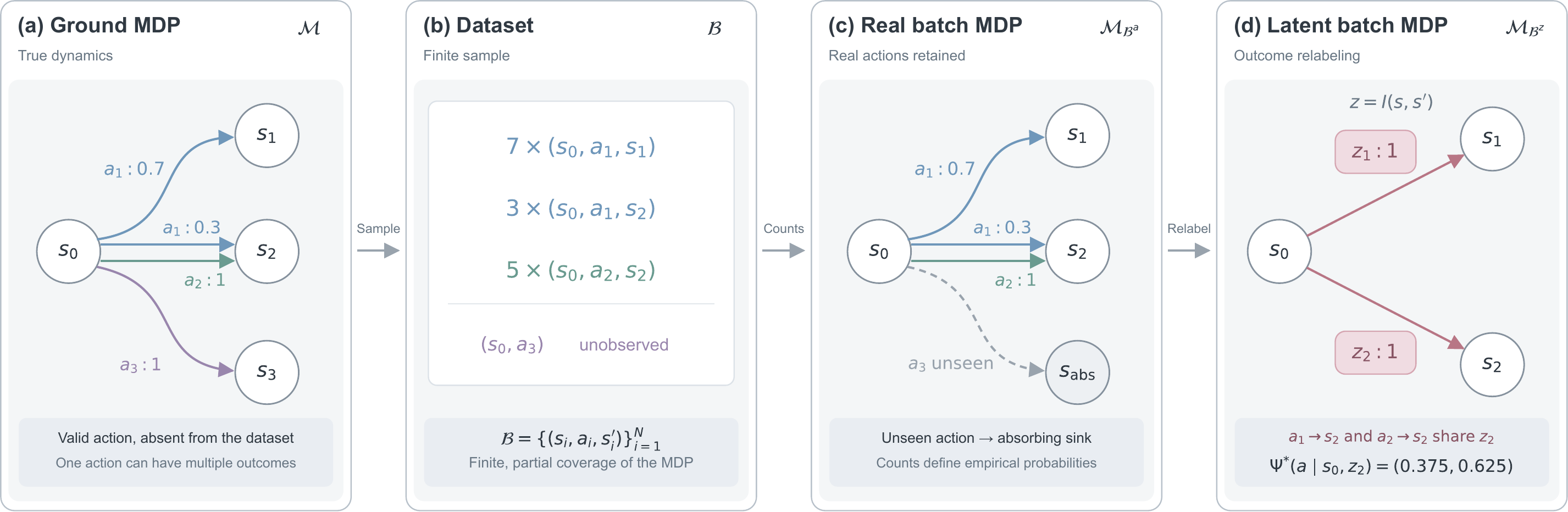}
\caption{\textbf{Paired real-action and latent-action batch MDPs.} \textbf{(a)} A real action in the ground MDP can induce multiple outcomes. \textbf{(b,c)} The finite dataset defines an empirical real-action batch MDP, with unseen actions routed to an absorbing sink. \textbf{(d)} Relabeling transitions by $z=I(s,s')$ separates distinct outcomes while merging different real actions that reach the same successor.}
\label{fig:paired-batch-mdps}
\end{figure}

We first formalize the paired physical and latent batch MDPs, then establish policy improvement transfer, optimal-policy alignment, and the supported reweighting guarantee underlying LAPIS. Appendix~\ref{app:proofs-notation} fixes notation and gives the exact construction of the paired physical and latent batch MDPs together with the batch-exact decoder $\Psi^*$. Appendix~\ref{app:proof-eval-exact} proves the on-support transfer result (Theorem~\ref{thm:eval}). Appendix~\ref{app:proof-anypolicy} defines the decoder-covered policy class and proves optimal-policy alignment (Theorem~\ref{thm:align}). Appendix~\ref{app:proofs-gate} establishes the supported reweighting guarantee underlying LAPIS's policy improvement principle. The correspondence between this analysis and practical LAPIS, together with the associated diagnostics, is discussed in Appendix~\ref{app:gaps}. We discuss the relationship of these results to classical batch-constrained RL and MDP abstraction theory in Appendix~\ref{app:prior-theory}.

\subsection{Batch-MDP definitions and decoder}\label{app:proofs-notation}

Throughout the analysis, we assume bounded rewards, $R_{\max}:=\sup_{s,a,s'}|r(s,a,s')|<\infty$, and a discount factor $0\leq\gamma<1$. For a policy $\pi$ in an MDP $\mathcal N$, we write $J_{\mathcal N}(\pi)=\mathbb E_{s_0\sim\mu_0}[V^\pi_{\mathcal N}(s_0)]$, using the same initial-state distribution $\mu_0$ for all comparisons. For the analysis, $\Ba$ pairs each observed transition with the physical action that generated it, whether or not that action label is available to the learning algorithm. The batch-MDP rewards are defined for the analysis; the learning algorithm does not require reward labels.

\paragraph{Support domain and termination.}
Let $S_{\B}$ be the set of states and successors observed in the batch, and let $N(s,a,s')$ denote the number of occurrences of $(s,a,s')$ in $\Ba$. The observed physical and latent actions at state $s$ are
\[
A_{\B}(s):=\left\{a:\sum_{s'}N(s,a,s')>0\right\},\qquad
Z_{\B}(s):=\left\{I(s,s'):\sum_aN(s,a,s')>0\right\}.
\]
We assume $\supp\mu_0\subseteq S_{\B}$ and consider a nonempty class of policies that remain on observed pairs until true termination. True terminal states have zero-reward absorbing self-loops in both the physical and latent processes. A trajectory cutoff is not treated as termination: a policy that reaches a nonterminal state with no observed continuation is excluded from the comparison class. All exact comparisons concern states reachable from $\mu_0$ under the policies being compared.

Both batch MDPs are defined on $S\cup\{s_{\mathrm{abs}}\}$, with unobserved pairs routed to the zero-reward absorbing state $s_{\mathrm{abs}}$. Starting from $S_{\B}$, their transitions remain within the finite set $S_{\B}\cup\{s_{\mathrm{abs}}\}$, even when the ground state space is continuous. For a fixed policy, the Bellman operator is a $\gamma$-contraction on bounded functions and has a unique fixed point. Sums in the exact analysis are over finite empirical support; in the continuous-kernel extension of Appendix~\ref{app:stochastic-transfer}, they are interpreted as integrals.

\begin{definition}[Real action batch MDP]\label{def:MB}
The physical-action batch MDP $\MB=(S\cup\{s_{\mathrm{abs}}\},A,p_{\Ba},r_{\Ba},\gamma)$ uses the empirical transition probabilities
\[
p_{\Ba}(s'\mid s,a)=\frac{N(s,a,s')}{\sum_{\tilde s}N(s,a,\tilde s)}
\quad\text{for }a\in A_{\B}(s).
\]
On observed transitions, $r_{\Ba}(s,a,s')=r(s,a,s')$. For unobserved pairs, $p_{\Ba}(s_{\mathrm{abs}}\mid s,a)=1$. The sink is absorbing, and all transitions into or within it receive zero reward. Consequently, $\|V^\pi_{\MB}\|_\infty\leq R_{\max}/(1-\gamma)$ for every policy $\pi$.
\end{definition}

\begin{definition}[Supported outcome identifiability]\label{def:supp-id}
At a fixed state $s$, any two observed successors $s'_1,s'_2$ satisfy
\[
I(s,s'_1)=I(s,s'_2)\quad\Longrightarrow\quad s'_1=s'_2.
\]
Thus, each $z\in Z_{\B}(s)$ identifies a unique observed successor, denoted $F_{\B}(s,z)$.
\end{definition}

This condition applies only to observed successors at the same state. It allows distinct physical actions to share a latent when they reach the same successor, but excludes different observed successors from sharing a latent. The map $F_{\B}$ is an exact lookup on the supported transition graph, not the learned forward predictor $F$ in Equation~\eqref{eq:lam}. Its definition does not require exact reconstruction by $F$, and the analysis evaluates $F_{\B}$ only at supported pairs.

\begin{definition}[Latent action batch MDP]\label{def:MBz}
Relabel each observed transition with $z_i=I(s_i,s_i')$ to obtain $\Bz=\{(s_i,z_i,s_i')\}_{i=1}^N$, and let $N^z(s,z,s')$ denote the relabeled transition counts. Under Definition~\ref{def:supp-id}, the latent-action batch MDP $\MBz=(S\cup\{s_{\mathrm{abs}}\},Z,p_{\Bz},r_{\Bz},\gamma)$ has, for $z\in Z_{\B}(s)$,
\begin{align*}
p_{\Bz}(\cdot\mid s,z)&=\delta_{F_{\B}(s,z)},\\
r_{\Bz}(s,z)&=\frac{\sum_aN(s,a,F_{\B}(s,z))r(s,a,F_{\B}(s,z))}
{\sum_aN(s,a,F_{\B}(s,z))}.
\end{align*}
The transition kernel is equivalently obtained by normalizing the relabeled counts. The reward averages the physical rewards associated with the represented successor, using their empirical frequencies. For unsupported pairs, $p_{\Bz}(s_{\mathrm{abs}}\mid s,z)=1$ and $r_{\Bz}(s,z)=0$, with the same absorbing-sink convention as in $\MB$. For every latent policy $\pi_z$, $\|V^{\pi_z}_{\MBz}\|_\infty\leq R_{\max}/(1-\gamma)$.
\end{definition}

\paragraph{Batch-exact decoder.}
For a supported pair $(s,z)$, write $s_z'=F_{\B}(s,z)$. The batch-exact decoder selects physical actions according to their observed frequencies among transitions from $s$ to $s_z'$:
\begin{equation}\label{eq:batch-decoder}
\Psi^*(a\mid s,z)=\frac{N(s,a,s_z')}{\sum_bN(s,b,s_z')}
=\frac{\pi_{\beta,\B}(a\mid s)p_{\Ba}(s_z'\mid s,a)}
{\sum_b\pi_{\beta,\B}(b\mid s)p_{\Ba}(s_z'\mid s,b)},
\end{equation}
Here, $\pi_{\beta,\B}(a\mid s)=\sum_{s'}N(s,a,s')/\sum_{b,s'}N(s,b,s')$ is the empirical action distribution at $s$. The denominators in Equation~\eqref{eq:batch-decoder} are positive on the supported domain.

Equation~\eqref{eq:batch-decoder} is an empirical Bayes decoder: it conditions the batch action distribution on the represented successor using $p_{\Ba}$, rather than the ground transition kernel $p$. Under deterministic ground dynamics, these kernels agree on observed state--action pairs, so the decoded actions reproduce the represented successor. Under stochastic dynamics, an action observed to reach $s_z'$ may reach a different successor when executed again. This analytical decoder $\Psi^*(a\mid s,z)$ is distinct from the learned state-marginal decoder $\Psi(a\mid z)$ used by LAPIS; their relationship is discussed in Appendix~\ref{app:decoder-gap}.

\paragraph{Illustrative example.} Figure~\ref{fig:paired-batch-mdps} illustrates how the same observed transitions define the physical- and latent-action batch MDPs. The physical-action batch MDP retains the empirical successor distribution of each observed action and routes the unobserved action $a_3$ to the absorbing sink, even though it is valid in the ground MDP. Relabeling by $z=I(s,s')$ separates the two observed successors of $a_1$ into distinct latents $z_1$ and $z_2$, while assigning the transitions of both $a_1$ and $a_2$ to $s_2$ the same latent $z_2$. The batch-exact decoder selects physical actions according to their frequencies for that successor: the three $a_1$-transitions and five $a_2$-transitions to $s_2$ give decoding probabilities $3/8$ and $5/8$, respectively. Because $a_1$ is stochastic, decoding $z_2$ need not reproduce $s_2$, despite its unique successor in the latent batch MDP. This distinction motivates the deterministic-dynamics assumption used for exact transfer in the following subsection.

\subsection{Proof of Theorem~\ref{thm:eval}}\label{app:proof-eval-exact}

We first establish that batch constraints are preserved by decoding: selecting only observed latent actions at a state also restricts physical execution to actions observed at that state.

\begin{remark}[Batch constraints under decoding]\label{rem:bc-transfer}
At any state $s$, if $\pi_z(\cdot\mid s)$ is supported on $Z_{\B}(s)$, then the decoded physical policy $(\Psi^*\circ\pi_z)(\cdot\mid s)$ is supported on $A_{\B}(s)$.
\end{remark}
\begin{proof}
By Equation~\eqref{eq:batch-decoder}, $\Psi^*(a\mid s,z)>0$ only if $N(s,a,F_{\B}(s,z))>0$, which implies $a\in A_{\B}(s)$. Averaging these decoder distributions over $z\sim\pi_z(\cdot\mid s)$ therefore preserves the physical-action support restriction.
\end{proof}

Under deterministic dynamics, these decoded actions also reproduce the successor represented by the selected latent. Together with the batch reward construction, this correspondence yields equality between latent and decoded physical returns.

\restated{Theorem}{thm:eval}{On-support policy improvement transfer}\emph{Under Assumption~\ref{ass:detT}, every batch-constrained latent policy $\pi_z$ satisfies $J_{\MBz}(\pi_z)=J_\M(\Psi^*\circ\pi_z)$. Consequently, return improvement between two such latent policies transfers to their batch-exact decoded physical policies.}

\begin{proof}

Fix a supported pair $(s,z)$ and write $s_z'=F_{\B}(s,z)$. Every action selected by $\Psi^*(\cdot\mid s,z)$ was observed to produce $s_z'$ from $s$. Under Assumption~\ref{ass:detT}, each such action satisfies $T(s,a)=s_z'$, giving
\[
\sum_a\Psi^*(a\mid s,z)p(\cdot\mid s,a)=\delta_{s_z'}=p_{\Bz}(\cdot\mid s,z).
\]
The decoder weights also match those used to define the latent reward:
\[
\sum_a\Psi^*(a\mid s,z)r(s,a,s_z')
=\frac{\sum_aN(s,a,s_z')r(s,a,s_z')}{\sum_aN(s,a,s_z')}
=r_{\Bz}(s,z).
\]
Thus, the one-step transition and expected reward agree for every supported pair. Individual decoded actions may receive different rewards, but their expectation matches the latent reward; Assumption~\ref{ass:reward} is not required.

Starting from the common initial distribution $\mu_0$, the latent policy selects only supported latents, and Remark~\ref{rem:bc-transfer} ensures that decoding selects observed physical actions. The transition identity above then implies, by induction, that the two processes have the same state distribution at every step. The continuation and terminal conventions in Appendix~\ref{app:proofs-notation} ensure that both processes remain well-defined on this common reachable support.

Averaging over $\pi_z$, the two processes therefore share the state-level transition kernel and expected reward
\[
P_{\pi_z}(\cdot\mid s)=\sum_z\pi_z(z\mid s)\delta_{F_{\B}(s,z)},\qquad
\bar r_{\pi_z}(s)=\sum_z\pi_z(z\mid s)r_{\Bz}(s,z).
\]
Both $V^{\pi_z}_{\MBz}$ and $V^{\Psi^*\circ\pi_z}_{\M}$ satisfy
\[
V=\bar r_{\pi_z}+\gamma P_{\pi_z}V
\]
on the common reachable support. Since $\gamma<1$, this Bellman equation has a unique bounded solution, so the two value functions agree there. Averaging over $\mu_0$ establishes return equality. Applying this equality separately to $\pi_z$ and $\pi_z'$ proves the improvement implication.
\end{proof}

\subsection{Proof of Theorem~\ref{thm:align} and the decoder-covered policy class}\label{app:proof-anypolicy}

Theorem~\ref{thm:eval} establishes return equality between a latent policy and its own decoded physical policy. This does not yet imply that an optimal latent policy attains the best return available to batch-constrained physical policies.

To see why, consider a deterministic one-step problem with two equally frequent observed actions satisfying $T(s,a_1)=T(s,a_2)=s_{\mathrm{term}}$, but receiving rewards $1$ and $0$, respectively. Both actions share one latent, and the batch-exact decoder selects each with probability $1/2$. The latent and decoded returns are therefore both $1/2$, consistent with Theorem~\ref{thm:eval}. A physical policy can instead select $a_1$ and obtain return $1$. Thus, outcome-based abstraction can create an optimality gap even without off-support actions or decoder error.

To establish alignment, we construct a value-equivalent latent counterpart for each physical policy in the comparison class. We first define the required coverage in terms of represented successors.

\begin{definition}[Decoder-covered policy]\label{def:covered}
Under deterministic dynamics, a physical policy $\pi$ is \emph{decoder-covered} from $\mu_0$ if, at every nonterminal state reachable under $\pi$ from $\mu_0$, each action in its support satisfies
\[
T(s,a)=F_{\B}(s,z)\quad\text{for some }z\in Z_{\B}(s).
\]

\end{definition}

This condition requires the successor itself to be represented in the batch, rather than merely requiring its code $I(s,T(s,a))$ to have appeared. Under Assumption~\ref{ass:detT}, every batch-constrained physical policy is decoder-covered because each observed action reaches its recorded successor. Coverage can also include unobserved actions that reach the same successors as observed actions, although Theorem~\ref{thm:align} compares only the batch-constrained subclass.

Under Assumption~\ref{ass:reward}, actions reaching the same successor from a given state also receive the same reward. We can therefore group a decoder-covered policy's action probabilities by successor without changing its return.

\begin{lemma}[Decoder-covered lift]\label{lem:cover}
Under Assumptions~\ref{ass:detT}--\ref{ass:reward}, a decoder-covered physical policy $\pi$ has the latent lift
\[
\pi_z(z\mid s)=\sum_a\pi(a\mid s)\mathbf1[T(s,a)=F_{\B}(s,z)],\qquad z\in Z_{\B}(s),
\]
on its reachable support. This lift is batch-constrained and satisfies
$J_\M(\pi)=J_\M(\Psi^*\circ\pi_z)=J_{\MBz}(\pi_z)$.
The lift may be extended arbitrarily outside the reachable domain. For a continuous physical-action policy, the sum over $a$ is interpreted as an integral; the supported latent set remains finite.
\end{lemma}
\begin{proof}

Decoder coverage ensures that every selected action reaches a represented successor. Since the encoder is deterministic and supported outcome identifiability holds, each action is assigned to exactly one supported latent. The displayed probabilities therefore sum to one and define a policy supported on $Z_{\B}(s)$.

By construction, at every reachable nonterminal state,
\[
\sum_z\pi_z(z\mid s)\delta_{F_{\B}(s,z)}
=\sum_a\pi(a\mid s)\delta_{T(s,a)}.
\]
The left-hand side is also the successor distribution induced by $\Psi^*\circ\pi_z$, as established in the proof of Theorem~\ref{thm:eval}. Thus, $\pi$ and its lifted-and-decoded policy have the same successor distribution. Under Assumption~\ref{ass:reward}, they also have the same expected reward.

Starting from $\mu_0$, these identities imply that the two policies induce the same state distributions at every step. Their value functions satisfy the same Bellman equation on the common reachable support and hence agree there, giving
\[
J_\M(\pi)=J_\M(\Psi^*\circ\pi_z).
\]

The lift selects only observed latents along this reachable support and satisfies the continuation convention of Appendix~\ref{app:proofs-notation}, so it is batch-constrained. Theorem~\ref{thm:eval} then gives the remaining equality with $J_{\MBz}(\pi_z)$.
\end{proof}

Every batch-constrained physical policy therefore has a latent counterpart with the same return. Conversely, decoding a batch-constrained latent policy preserves its return and produces a batch-constrained physical policy. These two correspondences establish optimal-policy alignment.

\restated{Theorem}{thm:align}{Optimal-policy alignment}
Under Assumptions~\ref{ass:detT}--\ref{ass:reward}, for the same initial distribution $\mu_0$,
\[
\pi_z^*\in\arg\max_{\pi_z\in\Pi_{\Bz}}J_{\MBz}(\pi_z)
\quad\Longrightarrow\quad
(\Psi^*\circ\pi_z^*)\in\arg\max_{\pi\in\Pi_{\Ba}}J_\M(\pi).
\]
\begin{proof}
Let $\pi_z^*$ maximize $J_{\MBz}$ over $\Pi_{\Bz}$. For any $\pi\in\Pi_{\Ba}$, decoder coverage follows from Assumption~\ref{ass:detT}, and Lemma~\ref{lem:cover} provides a lift $\widetilde\pi_z\in\Pi_{\Bz}$ with the same return. Consequently,
\[
J_\M(\pi)=J_{\MBz}(\widetilde\pi_z)
\le J_{\MBz}(\pi_z^*)
=J_\M(\Psi^*\circ\pi_z^*),
\]
where the inequality follows from latent optimality and the last equality follows from Theorem~\ref{thm:eval}.

Moreover, $(\Psi^*\circ\pi_z^*)\in\Pi_{\Ba}$ by Remark~\ref{rem:bc-transfer} and the reachable-support argument in Theorem~\ref{thm:eval}. The decoded policy therefore belongs to the physical comparison class and attains at least as much return as every policy in that class, proving the claimed optimality. In particular,
\[
J_\M(\Psi^*\circ\pi_z^*)=\max_{\pi\in\Pi_{\Ba}}J_\M(\pi)=J_{\MBz}(\pi_z^*).
\]
The alignment concerns the stated batch-constrained policy classes, not the unrestricted physical optimum.
\end{proof}

\subsection{Supported reweighting guarantee}\label{app:proofs-gate}

We now formalize the policy-improvement principle underlying critic-guided filtering. The update reweights a fixed reference policy toward higher-value latent actions without introducing actions outside its support. Throughout this subsection, we fix a goal $g$ and use the common closed supported comparison domain of Appendix~\ref{app:proofs-notation}.

We first show that a reweighting that favors higher-value actions cannot decrease the expected action value under the reference policy.

\begin{lemma}[Co-monotone reweighting]\label{lem:chebyshev}
Fix $(s,g)$ and a reference latent policy $\mu(\cdot\mid s,g)$ on the finite supported latent set. Let $\omega(z)$ be bounded, nonnegative, and nondecreasing in $Q^\mu(s,z,g)$, with $C_\omega:=\E_\mu[\omega]>0$. Then the reweighted policy
\[
\mu^\omega(z\mid s,g)=\frac{\mu(z\mid s,g)\omega(z)}{C_\omega}
\]
satisfies
\[
\E_{\mu^\omega}[Q^\mu(s,z,g)]\ge\E_\mu[Q^\mu(s,z,g)]=V^\mu(s,g).
\]
\end{lemma}
\begin{proof}
Let $z,\tilde z$ be independent samples from $\mu(\cdot\mid s,g)$, and abbreviate $Q_z=Q^\mu(s,z,g)$. Since the weights are nondecreasing in value,
\[
0\le\tfrac12\E[(Q_z-Q_{\tilde z})(\omega(z)-\omega(\tilde z))]
=\E_\mu[Q\omega]-\E_\mu[Q]\E_\mu[\omega].
\]
Dividing by $C_\omega>0$ gives
\[
\E_{\mu^\omega}[Q]=\frac{\E_\mu[Q\omega]}{C_\omega}\ge\E_\mu[Q],
\]
which proves the claim.
\end{proof}

The weights need not depend directly on the numerical value of $Q^\mu$. Advantage-based weights and thresholded scores also satisfy the required ordering.

\begin{corollary}\label{cor:gate}
A bounded, nonnegative, nondecreasing function $h$ of the advantage
\[
A^\mu(s,z,g)=Q^\mu(s,z,g)-V^\mu(s,g)
\]
defines weights $\omega(z)=h(A^\mu(s,z,g))$ satisfying Lemma~\ref{lem:chebyshev} whenever the normalizing mass is positive. More generally, the thresholded weights
\[
\omega(z)=\mathbf1[f(s,z,g)>b(s,g)]
\]
satisfy Lemma~\ref{lem:chebyshev} whenever $f(s,\cdot,g)$ is nondecreasing in $Q^\mu(s,\cdot,g)$, $b$ is independent of $z$, and the normalizing mass is positive.
\end{corollary}
\begin{proof}
Subtracting a baseline independent of $z$ preserves the value ordering. Applying a nondecreasing function or thresholding a nondecreasing score at a $z$-independent cutoff likewise preserves this ordering. The resulting weights therefore satisfy Lemma~\ref{lem:chebyshev} under the stated positivity condition.
\end{proof}

This observation connects value-based reweighting to the contrastive score used to motivate LAPIS.

\begin{remark}[Density-ratio scores and value ordering]\label{rem:score-order}
For the binary density-ratio objective introduced in Section~\ref{sec:prelim}, the optimal score is
\[
f^\star(s,z,g)=\log Q^\mu(s,z,g)-\log p_{\B}(g)
\]
wherever the ratio is defined. At fixed $(s,g)$, the term $-\log p_{\B}(g)$ is independent of $z$, so $f^\star$ preserves the ordering of $Q^\mu$. A $z$-independent score threshold therefore produces nonnegative weights that are nondecreasing in value; the threshold need not equal $V^\mu(s,g)$. Appendix~\ref{app:critic-ordering} relates this score-space filtering principle to the learned critic.
\end{remark}

Applying the statewise inequality at every state yields a return guarantee for the reweighted policy. When all weights vanish, retaining the reference policy makes the update well-defined.

\begin{proposition}[Goal-conditioned supported improvement]\label{prop:supported-improve}
Let $\pi_\beta^z$ be a batch-constrained reference policy, with advantage $A_\beta^z$ in $\MBz$. For $\eta>0$, define
\[
\pi_\eta^z(z\mid s,g)\propto\pi_\beta^z(z\mid s,g)h(A_\beta^z(s,z,g))^\eta,
\]
where $h$ is bounded, nonnegative, and nondecreasing. At states with zero normalizing mass, set $\pi_\eta^z=\pi_\beta^z$, and define $\pi_0^z=\pi_\beta^z$. Then
\[
\supp\pi_\eta^z(\cdot\mid s,g)\subseteq\supp\pi_\beta^z(\cdot\mid s,g),\qquad
J_{\MBz}(\pi_\eta^z;g)\ge J_{\MBz}(\pi_\beta^z;g).
\]
Under the conditions of Theorem~\ref{thm:eval}, the same return ordering holds for their batch-exact decoded physical policies.
\end{proposition}
\begin{proof}
Reweighting cannot assign positive probability to a latent action outside the reference policy's support, and the fallback leaves the reference policy unchanged. The updated policy therefore remains batch-constrained on the common supported domain.

For $\eta>0$, the weights $h(A_\beta^z)^\eta$ remain nonnegative and nondecreasing in the reference action value. At each state with positive normalizing mass, Lemma~\ref{lem:chebyshev} gives
\[
\E_{z\sim\pi_\eta^z(\cdot\mid s,g)}\!\left[Q^{\pi_\beta^z}_{\MBz}(s,z,g)\right]\ge V^{\pi_\beta^z}_{\MBz}(s,g).
\]
At fallback states, or when $\eta=0$, equality holds. Writing $V_\beta=V^{\pi_\beta^z}_{\MBz}(\cdot,g)$ and $\mathcal T^{\pi_\eta^z}$ for the updated policy's Bellman operator, this inequality is
\[
\mathcal T^{\pi_\eta^z}V_\beta\ge V_\beta.
\]
By monotonicity, repeated application of $\mathcal T^{\pi_\eta^z}$ produces a nondecreasing sequence starting from $V_\beta$. Since the operator is a $\gamma$-contraction, this sequence converges to $V^{\pi_\eta^z}_{\MBz}(\cdot,g)$, establishing pointwise value improvement. Averaging over $\mu_0$ proves the return inequality.

Both policies are batch-constrained, so Theorem~\ref{thm:eval} transfers this ordering to their decoded physical returns under its assumptions. The argument holds for each fixed goal and therefore also after averaging over a common goal distribution.
\end{proof}

The exponent $\eta$ controls distributional reweighting and is distinct from the flow-guidance scale $w$ in Section~\ref{sec:lapis}. In particular, for the hard gate $h=\mathbf1[\cdot>0]$, we have $h^\eta=h$ for every $\eta>0$, so the reweighted policy is independent of $\eta$ over this range. The proposition concerns this exact distributional update; guided flow samples need not follow the same distribution. Their relationship is discussed in Appendix~\ref{app:extraction-gap}.

\begin{revblock}
\subsection{$H$-step endpoint latents: proof of Theorem~\ref{thm:eval-h}}\label{app:proof-hstep}

\begin{theoremprime}[On-support transfer for $H$-step latents]\label{thm:eval-h}
Let $\MBzH$ be the latent batch MDP over observed endpoint latents with the empirical one-step transition $\pH$ and reward $\rH$ given $(s_t,z_t)$, and $\Psi^*_H(a\mid s,z)$ the empirical first-action decoder (Definition~\ref{def:MBzH}). Under Assumption~\ref{ass:detT}, every latent policy $\pi_z$ batch-constrained in $\MBzH$ satisfies $J_{\MBzH}(\pi_z)=J_\M(\Psi^*_H\circ\pi_z)$, so return improvement transfers to the decoded policies. Supported outcome identifiability is not required, and $\pH(\cdot\mid s,z)$ may be stochastic. For $H=1$ under supported outcome identifiability, $\mathcal M_{\mathcal B^z_1}=\MBz$ and $\Psi^*_1=\Psi^*$, recovering Theorem~\ref{thm:eval}.
\end{theoremprime}

The deployed controller labels each recorded step with an $H$-step endpoint latent and executes only the first decoded action before replanning (Appendix~\ref{app:macro-align}). This subsection defines the corresponding latent batch MDP and proves Theorem~\ref{thm:eval-h}. We keep the conventions of Appendix~\ref{app:proofs-notation} for rewards, the absorbing sink, initial states, and termination.

\paragraph{Setting.} For every recorded step $(s_t,a_t,s_{t+1})$ whose endpoint $s_{t+H}$ is also recorded, let $z_t=I(s_t,s_{t+H})$, and let $N(s,a,z,s')$ count the occurrences of $(s_t,a_t,z_t,s_{t+1})=(s,a,z,s')$. Marginal counts omit arguments, for example $N(s,z)=\sum_{a,s'}N(s,a,z,s')$ and $N(s,a,z)=\sum_{s'}N(s,a,z,s')$.

\begin{definition}[$H$-step endpoint latent batch MDP]\label{def:MBzH}
The latent batch MDP $\MBzH=(S\cup\{s_{\mathrm{abs}}\},Z,\pH,\rH,\gamma)$ has observed latents $Z_{\B,H}(s)=\{z:N(s,z)>0\}$ and, for $z\in Z_{\B,H}(s)$,
\[
\pH(s'\mid s,z)=\frac{N(s,z,s')}{N(s,z)},\qquad
\rH(s,z)=\frac{\sum_{a,s'}N(s,a,z,s')\,r(s,a,s')}{N(s,z)}.
\]
Unsupported pairs transition to $s_{\mathrm{abs}}$ with zero reward, as in Definition~\ref{def:MBz}. The first-action batch decoder is
\[
\Psi^*_H(a\mid s,z)=\frac{N(s,a,z)}{N(s,z)}.
\]
\end{definition}

The kernel $\pH$ is the one-step successor distribution conditioned on the endpoint latent introduced in Appendix~\ref{app:macro-align}. No identifiability condition is imposed: the same $z$ at $s$ may be followed by different first actions and different successors, so $\pH(\cdot\mid s,z)$ can be stochastic even under deterministic ground dynamics. Because $\rH$ averages bounded rewards, $\|V^{\pi_z}_{\MBzH}\|_\infty\le R_{\max}/(1-\gamma)$ for every latent policy. A latent policy is batch-constrained in $\MBzH$ if it selects only $z\in Z_{\B,H}(s)$ along its reachable states, with the continuation convention of Appendix~\ref{app:proofs-notation}. Decoded execution samples $z\sim\pi_z(\cdot\mid s)$ and $a\sim\Psi^*_H(\cdot\mid s,z)$, executes $a$, and samples again at the next state.

\paragraph{The case $H=1$.} Then $z_t=I(s_t,s_{t+1})$ and $N(s,a,z,s')=N(s,a,s')\mathbf 1[z=I(s,s')]$. Under supported outcome identifiability (Definition~\ref{def:supp-id}), $N(s,z,s')>0$ for $z\in Z_{\B}(s)$ only if $s'=F_{\B}(s,z)$. Hence $\mathcal{M}_{\mathcal{B}^{z}_{1}}=\MBz$ and $\Psi^*_1(a\mid s,z)=N(s,a,F_{\B}(s,z))/\sum_bN(s,b,F_{\B}(s,z))=\Psi^*(a\mid s,z)$, and Theorem~\ref{thm:eval} is the case $H=1$ of Theorem~\ref{thm:eval-h}.

\begin{proof}[Proof of Theorem~\ref{thm:eval-h}]
\emph{Batch constraints under decoding.} $\Psi^*_H(a\mid s,z)>0$ implies $N(s,a,z)>0$, hence $a\in A_{\B}(s)$, as in Remark~\ref{rem:bc-transfer}.

\emph{Transition and reward identities.} Under Assumption~\ref{ass:detT}, every recorded step from $(s,a)$ reaches $T(s,a)$, so $N(s,a,z,s')=N(s,a,z)\mathbf 1[s'=T(s,a)]$. For a supported pair $(s,z)$,
\begin{align*}
\sum_a\Psi^*_H(a\mid s,z)\,p(s'\mid s,a)
&=\sum_a\frac{N(s,a,z)\mathbf 1[T(s,a)=s']}{N(s,z)}\\
&=\sum_a\frac{N(s,a,z,s')}{N(s,z)}
=\pH(s'\mid s,z),
\end{align*}
\[
\sum_a\Psi^*_H(a\mid s,z)\,r(s,a,T(s,a))
=\frac{\sum_{a,s'}N(s,a,z,s')\,r(s,a,s')}{N(s,z)}
=\rH(s,z).
\]
Thus the decoded one-step transition and expected reward agree with those of $\MBzH$ at every supported pair.

\emph{Return equality.} The remainder is the proof of Theorem~\ref{thm:eval} with $\delta_{F_{\B}(s,z)}$ and $r_{\Bz}$ replaced by $\pH(\cdot\mid s,z)$ and $\rH$. Averaging over $\pi_z$, both processes have the state-level kernel $P_{\pi_z}(\cdot\mid s)=\sum_z\pi_z(z\mid s)\pH(\cdot\mid s,z)$ and expected reward $\bar r_{\pi_z}(s)=\sum_z\pi_z(z\mid s)\rH(s,z)$ on the common reachable support, so $V^{\pi_z}_{\MBzH}$ and $V^{\Psi^*_H\circ\pi_z}_\M$ solve the same Bellman equation $V=\bar r_{\pi_z}+\gamma P_{\pi_z}V$, whose bounded solution is unique for $\gamma<1$. Averaging over $\mu_0$ gives return equality, and applying it to two policies gives the improvement implication.
\end{proof}

\paragraph{Correspondence with practical LAPIS.} LAPIS selects $z_t$ at $s_t$, executes only the first decoded action ($E=1$), and selects again at $s_{t+1}$, which is decoded execution in $\MBzH$. Its critic propagates value from $(s_t,z_t)$ to $(s_{t+1},z_{t+1})$ with the recorded continuation $z_{t+1}=I(s_{t+1},s_{t+1+H})$ (Appendix~\ref{app:macro-align}), so it evaluates the behavior latent policy in the same MDP. Theorem~\ref{thm:eval-h} covers the batch-exact first-action decoder; Proposition~\ref{prop:stoch-transfer} with reference kernel $\pH$ covers learned decoders (Appendix~\ref{app:stochastic-transfer}).
\end{revblock}

\newpage
\section{Theory--Practice Correspondence}\label{app:gaps}

The analysis in Section~\ref{sec:cfg-crl} isolates the conditions under which latent policy improvement remains valid after physical execution. Practical LAPIS realizes these conditions through learned latent representations, decoding, value estimation, and policy extraction. This appendix makes each correspondence explicit and extends the exact analysis to the learned components used in practice. Exact transfer arises as the zero-mismatch case of the transition and reward correspondence below; supported reweighting defines the target distribution approximated by the learned policy.

\subsection{Learned representations and supported outcome identifiability}\label{app:vq-identifiability}

Supported outcome identifiability (Definition~\ref{def:supp-id}) requires each supported state--latent pair $(s,z)$ to identify a unique observed successor $F_{\B}(s,z)$. In practice, this structure is learned through the inverse model $I$ and forward model $F$. Reconstruction connects the learned representation to outcome identification. If two observed successors are assigned the same latent, $I(s,s'_1)=I(s,s'_2)=z$, then
\[
\|s'_1-s'_2\|_2
\le \|s'_1-F(s,z)\|_2+\|s'_2-F(s,z)\|_2.
\]
Thus, when both transitions are reconstructed accurately, successors sharing a latent are close in the reconstruction metric. The theorem states supported outcome identifiability as an exact structural condition, while the reconstruction objective encourages this structure on observed transitions.

The experimental policy operates on quantized latent codes. Identifiability concerns whether the code supplied to the policy and decoder distinguishes the observed outcomes relevant for control. The learned predictor $F$ and the exact supported lookup $F_{\B}$ play complementary roles: $F$ is trained to reconstruct outcomes, whereas $F_{\B}$ states the exact correspondence on the observed transition graph. Accurate reconstruction encourages outcome consistency; exact identifiability additionally requires that the supported code distinguish these outcomes.

\subsection{State representations and Markov control}\label{app:visual-markov}

The analysis in Section~\ref{sec:cfg-crl} uses Markov states to isolate the relationship between latent decisions and physical execution from perceptual representation learning. The experiments use single-frame RGB observations processed by learned visual encoders. Practical LAPIS applies the latent policy improvement principle to this learned observation representation.

Different underlying states can produce indistinguishable observations, so the observation-level process need not be exactly Markov. This concerns the state representation, whereas latent-action realizability concerns whether a selected latent transition can be physically executed from the current state. The exact results characterize the control-level correspondence under Markov state information. Temporal or recurrent representations provide possible ways to retain additional state information; the analysis applies when the resulting representation satisfies the stated Markov and transition conditions.

\subsection{Approximate transfer under execution mismatch}\label{app:stochastic-transfer}

Theorem~\ref{thm:eval} gives exact return equality when supported latent transitions are executed through the batch-exact decoder under deterministic dynamics. More generally, the same correspondence admits a quantitative extension whenever decoded execution induces transitions and rewards close to those used for latent evaluation. We state the bound for a decoder $\widehat\Psi(a\mid s,z)$, which includes both $\Psi^*$ and the state-marginal choice $\widehat\Psi(a\mid s,z)=\Psi(a\mid z)$.

\rev{\paragraph{Reference latent process.} The bound compares decoded execution with a reference latent batch MDP. For one-step latents this is $\MBz$ with $(Z_{\B},p_{\Bz},r_{\Bz})$. For the $H$-step endpoint latents of the deployed controller it is $\MBzH$ with $(Z_{\B,H},\pH,\rH)$ (Definition~\ref{def:MBzH}). We write the statements below for $\MBz$. They hold verbatim for $\MBzH$ after replacing $p_{\Bz}$ and $r_{\Bz}$ by $\pH$ and $\rH$, since the argument uses only the two Bellman equations and the reward bound, not the form of the reference kernel. Because the controller executes only the first decoded action before replanning (Appendix~\ref{app:macro-align}), $\widehat\Psi(a\mid s,z)$ denotes the distribution of this first action; for a chunk decoder, it is the first-action marginal of the decoded chunk.}

Fix a latent policy $\pi_z$ that is batch-constrained in $\MBz$. We compare the two processes on a common Markov state space and decision interval, with the same $\gamma$, reward bound $R_{\max}$, and initial distribution $\mu_0$. Use the ambient-space batch kernels of Appendix~\ref{app:proofs-notation}, extend the ground process with the same absorbing sink, and define $\pi_z$ and $\widehat\Psi$ as probability kernels throughout this space. Where an empirical policy or decoder is undefined, fix an arbitrary probability-kernel extension. The occupancy below includes any off-support visits under this specified execution. Sums over continuous variables denote integrals.

The decoder induces
\begin{align*}
\tilde p_{\widehat\Psi}(s'\mid s,z)
&:=\sum_a\widehat\Psi(a\mid s,z)\,p(s'\mid s,a),\\
\tilde r_{\widehat\Psi}(s,z)
&:=\sum_a\widehat\Psi(a\mid s,z)\sum_{s'}p(s'\mid s,a)\,r(s,a,s').
\end{align*}
For this fixed decoder, define the mismatches
\begin{align*}
\epsilon_p(s,z)&:=\TV\!\bigl(p_{\Bz}(\cdot\mid s,z),\tilde p_{\widehat\Psi}(\cdot\mid s,z)\bigr),\\
\epsilon_r(s,z)&:=\bigl|r_{\Bz}(s,z)-\tilde r_{\widehat\Psi}(s,z)\bigr|,
\end{align*}
where $\TV(P,Q)=\sup_A|P(A)-Q(A)|$. Let $d^{\widehat\Psi\circ\pi_z}$ be the normalized $\gamma$-discounted occupancy of $(s,z)$ under decoded execution in $\M$ from $\mu_0$, with $z\sim\pi_z(\cdot\mid s)$. Write $\bar\epsilon_p=\E_{d^{\widehat\Psi\circ\pi_z}}[\epsilon_p]$ and $\bar\epsilon_r=\E_{d^{\widehat\Psi\circ\pi_z}}[\epsilon_r]$.

\begin{proposition}[Approximate transfer under execution mismatch]\label{prop:stoch-transfer}
For a batch-constrained latent policy $\pi_z$ and the fixed, globally defined decoder $\widehat\Psi$ above,
\begin{equation}\label{eq:learned-decoder-transfer}
\bigl|J_\M(\widehat\Psi\circ\pi_z)-J_{\MBz}(\pi_z)\bigr|
\le\frac{\bar\epsilon_r}{1-\gamma}
+\frac{2\gamma R_{\max}}{(1-\gamma)^2}\,\bar\epsilon_p.
\end{equation}
Zero transition and reward mismatch yields exact return equality. In particular, $\widehat\Psi=\Psi^*$ under Assumption~\ref{ass:detT} recovers Theorem~\ref{thm:eval}. \rev{With reference $\MBzH$, $\widehat\Psi=\Psi^*_H$ under Assumption~\ref{ass:detT} makes both mismatches zero by the identities in the proof of Theorem~\ref{thm:eval-h}, which recovers Theorem~\ref{thm:eval-h}.}
\end{proposition}

\begin{proof}
\emph{Step 1: Bellman residual.} Write $V_1=V^{\widehat\Psi\circ\pi_z}_\M$ and $V_2=V^{\pi_z}_{\MBz}$. Their Bellman equations are
\begin{align*}
V_1(s)&=\sum_z\pi_z(z\mid s)\Bigl[\tilde r_{\widehat\Psi}(s,z)+\gamma\sum_{s'}\tilde p_{\widehat\Psi}(s'\mid s,z)V_1(s')\Bigr],\\
V_2(s)&=\sum_z\pi_z(z\mid s)\Bigl[r_{\Bz}(s,z)+\gamma\sum_{s'}p_{\Bz}(s'\mid s,z)V_2(s')\Bigr].
\end{align*}
Let $(\bar P_{\widehat\Psi}h)(s)=\sum_{z,s'}\pi_z(z\mid s)\tilde p_{\widehat\Psi}(s'\mid s,z)h(s')$. Subtracting the equations gives $\Delta=\delta+\gamma\bar P_{\widehat\Psi}\Delta$, where $\Delta=V_1-V_2$ and
\[
\delta(s)=\sum_z\pi_z(z\mid s)\Bigl[
\tilde r_{\widehat\Psi}(s,z)-r_{\Bz}(s,z)
+\gamma\sum_{s'}(\tilde p_{\widehat\Psi}-p_{\Bz})(s'\mid s,z)V_2(s')\Bigr].
\]
\emph{Step 2: bounding the residual.} Definition~\ref{def:MBz} gives $\|V_2\|_\infty\le R_{\max}/(1-\gamma)$. The total-variation inequality $|\E_P h-\E_Q h|\le2\|h\|_\infty\TV(P,Q)$ therefore yields
\[
|\delta(s)|\le\sum_z\pi_z(z\mid s)\Bigl[\epsilon_r(s,z)+\frac{2\gamma R_{\max}}{1-\gamma}\epsilon_p(s,z)\Bigr].
\]
\emph{Step 3: occupancy-weighted bound.} Since $\bar P_{\widehat\Psi}$ is a stochastic operator and $\gamma<1$, unrolling the residual along decoded execution gives
\begin{align*}
\bigl|J_\M(\widehat\Psi\circ\pi_z)-J_{\MBz}(\pi_z)\bigr|
&\le\sum_{t\ge0}\gamma^t\E[|\delta(s_t)|]\\
&\le\frac{1}{1-\gamma}\E_{d^{\widehat\Psi\circ\pi_z}}
\left[\epsilon_r+\frac{2\gamma R_{\max}}{1-\gamma}\epsilon_p\right],
\end{align*}
which is the stated bound. If $\widehat\Psi=\Psi^*$ and Assumption~\ref{ass:detT} holds, the supported transition and expected-reward identities in the proof of Theorem~\ref{thm:eval} make both mismatches zero along the rollout. This recovers exact transfer without requiring Assumption~\ref{ass:reward}.
\end{proof}

For the fixed decoder, let $\varepsilon_{\widehat\Psi}(\pi_z)$ denote the right-hand side of Equation~\eqref{eq:learned-decoder-transfer}, using the occupancy of $\widehat\Psi\circ\pi_z$.

\begin{corollary}[Approximate improvement transfer]\label{cor:stoch-improve}
For batch-constrained latent policies $\pi_z,\pi_z'$ and the same decoder $\widehat\Psi$, define $\Delta_z=J_{\MBz}(\pi_z')-J_{\MBz}(\pi_z)$. Then
\[
J_\M(\widehat\Psi\circ\pi_z')-J_\M(\widehat\Psi\circ\pi_z)
\ge\Delta_z-\varepsilon_{\widehat\Psi}(\pi_z')-\varepsilon_{\widehat\Psi}(\pi_z).
\]
Thus a latent improvement exceeding the two execution-gap bounds transfers to physical improvement.
\end{corollary}
\rev{Proposition~\ref{prop:stoch-transfer} and Corollary~\ref{cor:stoch-improve} hold unchanged with $\MBzH$ as the reference.}
\begin{proof}
Add and subtract the two latent returns and apply Proposition~\ref{prop:stoch-transfer} separately to $\pi_z'$ and $\pi_z$.
\end{proof}

\paragraph{Interpretation.}
The bound accounts jointly for stochastic dynamics and decoder approximation through their induced transition and reward mismatch. Its quantitative usefulness depends on this mismatch being small in the stated metric: a point mass and an atomless transition law have total-variation distance one even if they are geometrically close. Appendix~\ref{app:macro-align} describes the experimental controller.

\subsection{Learned decoding under scarce action supervision}\label{app:decoder-gap}\label{app:marginal}\label{app:decoder-conditioning}\label{app:decoder-cond}

The batch-exact decoder establishes an exact correspondence, while successful execution fundamentally requires realizing the represented successor. Recovering a particular recorded action distribution is unnecessary when alternative controls induce the same transition and reward. Under Assumptions~\ref{ass:detT}--\ref{ass:reward}, a decoder satisfying
\[
\widehat\Psi\bigl(\{a:T(s,a)=F_{\B}(s,z)\}\mid s,z\bigr)=1
\]
on visited supported pairs induces the same successor and reward as the latent transition. Hence $\epsilon_p=\epsilon_r=0$ in Proposition~\ref{prop:stoch-transfer}, even if its action distribution differs from the empirical weights of $\Psi^*$. In this outcome-compatible setting, consistent execution of the latent outcome, rather than recovery of a particular recorded action, is sufficient for transfer.

Explicit state conditioning is therefore not a requirement of the transfer argument when the decoder realizes the represented outcome on every visited supported pair. LAPIS learns a state-marginal decoder $\Psi(a\mid z)$ from scarce action labels. Table~\ref{tab:decoder-cond} compares it with a state-conditioned decoder using the same frozen $5\%$-label LAM across decoder-label budgets; within each budget, the remaining policy and training setup is shared between decoder variants. Marginal decoding achieves higher success across all three domains at $1\%$ and $5\%$ labels, while state conditioning performs better with full supervision. These results support marginal decoding for the scarce-label setting studied here.

\subsection{Goal-conditioned objectives and outcome alignment}\label{app:goal-outcome}

Theorem~\ref{thm:align} requires physical controls that produce the same transition outcome to receive the same reward. The goal-conditioned objective used by LAPIS satisfies this condition under the deterministic dynamics of Assumption~\ref{ass:detT}. Contrastive RL uses the goal-reaching reward
\[
r_g(s,a)=(1-\gamma)p(s'=g\mid s,a).
\]
Under Assumption~\ref{ass:detT}, $s'=T(s,a)$, so
\[
r_g(s,a)=(1-\gamma)\mathbf 1[T(s,a)=g].
\]
Consequently, if two physical actions produce the same successor,
\[
T(s,a_1)=T(s,a_2)=s',
\]
then
\[
r_g(s,a_1)=r_g(s,a_2)=(1-\gamma)\mathbf 1[s'=g].
\]
Thus, in the deterministic setting analyzed in Section~\ref{sec:cfg-crl}, the goal-conditioned reward depends on a physical control only through its transition outcome. Outcome-based abstraction therefore preserves the reward distinctions relevant to the goal-conditioned objective, providing the objective-alignment condition used in Theorem~\ref{thm:align}.

\subsection{Representation horizon and receding-horizon control}\label{app:macro-align}

The analysis in Section~\ref{sec:cfg-crl} is stated at the policy decision interval, where a latent action represents the successor of a single transition. In visual latent-action learning, however, the temporal gap used to construct the latent representation also controls how much transition information is visible to the encoder. We distinguish three quantities: the LAM temporal gap $H$ in $z_t=I(s_t,s_{t+H})$; the physical-action chunk size $C$, the number of actions predicted by a real-action policy or a latent-action decoder; and the execution horizon $E$, the number of actions executed before replanning. The symbol $H$ refers only to the LAM temporal gap and does not apply to real-action policies without a LAM.

Our default LAM uses a representation horizon of $H=3$,
\[
z_t=I(s_t,s_{t+3}),
\]
but LAPIS remains a one-step receding-horizon controller: the decoder predicts a chunk of $C=3$ actions, only the first action is executed ($E=1$), and the policy replans from $s_{t+1}$. The critic likewise propagates value at this one-step interval using $s_{t+1}$ and the recorded continuation latent $z_{t+1}=I(s_{t+1},s_{t+4})$. Thus, the longer temporal gap changes the representation used for control, rather than requiring longer open-loop execution.

\paragraph{Representation and execution horizons.}
We empirically separate these two roles by varying the representation horizon and the number of decoded actions executed before replanning. Both $H=1$ and our default $H=3$ controller execute a single action before replanning. With this execution scheme, reducing the representation horizon from $H=3$ to $H=1$ sharply reduces task success across all three representative domains. Conversely, keeping the $H=3$ representation but executing the full three-action chunk does not account for the benefit of the longer horizon: one-step replanning achieves higher mean performance in all three domains, with a particularly large difference on AntMaze.

\begin{table}[htbp]
\centering

\caption{LAM temporal gap $H$ and execution horizon $E$. Success rate (\%) on representative tasks. The decoder chunk size is $C=H$ in these ablations; $E$ is the number of decoded actions executed before replanning.}
\label{tab:execution-horizon}
\begin{tabular}{lccccc}
\toprule
Setting & $H$ & $E$ & AntMaze-medium & Cube-single & Scene \\
\midrule
One-step representation & 1 & 1 & $4${\scriptsize$\,\pm1$} & $2${\scriptsize$\,\pm1$} & $0${\scriptsize$\,\pm0$} \\
LAPIS & 3 & 1 & $\mathbf{49}${\scriptsize$\,\mathbf{\pm3}$} & $\mathbf{62}${\scriptsize$\,\mathbf{\pm5}$} & $\mathbf{29}${\scriptsize$\,\mathbf{\pm4}$} \\
Full-chunk execution & 3 & 3 & $9${\scriptsize$\,\pm5$} & $56${\scriptsize$\,\pm3$} & $26${\scriptsize$\,\pm7$} \\
\bottomrule
\end{tabular}
\end{table}

The control results are consistent with the representation diagnostics. With matched frozen LAMs, increasing the temporal gap from $H=1$ to $H=3$ raises held-out action-decodability $R^2$ from $0.501$ to $0.597$ on AntMaze, $0.586$ to $0.778$ on Cube, and $0.628$ to $0.811$ on Scene (Table~\ref{tab:horizon-lam-policy}). Adjacent visual observations can contain only small action-dependent changes, whereas a longer gap exposes more visible transition structure to the latent bottleneck. Together, the representation probes and control results indicate that $H=3$ is useful primarily as a \emph{representation horizon}, while one-step replanning remains effective for closed-loop execution.

\paragraph{Connection to the one-step analysis.}
When $H=1$, the encoded endpoint coincides with the successor at the control interval, giving the direct latent--successor correspondence analyzed in Section~\ref{sec:cfg-crl}. For $H>1$, an endpoint latent can in principle summarize trajectories with different intermediate successors even though execution and value propagation remain one-step. The appropriate practical object is therefore the one-step successor distribution conditioned on the endpoint latent,
\[
p^{(1)}_{\B,H}(s^+\mid s,z)
:=\Pr_{\B}\!\left(s_{t+1}=s^+\mid s_t=s,\,I(s_t,s_{t+H})=z\right).
\]
\rev{The corresponding physical process is the one-step transition induced by the first decoded action. Theorem~\ref{thm:eval-h} gives exact transfer for this object: in the $H$-step endpoint latent batch MDP $\MBzH$, whose transition kernel is $p^{(1)}_{\B,H}$, decoding with the first-action batch decoder $\Psi^*_H$ preserves return under deterministic dynamics (Appendix~\ref{app:proof-hstep}). For learned decoders, Proposition~\ref{prop:stoch-transfer} with reference kernel $p^{(1)}_{\B,H}$ bounds the gap (Appendix~\ref{app:stochastic-transfer}). The critic bootstraps from $(s_t,z_t)$ to $(s_{t+1},z_{t+1})$ and therefore evaluates latent policies in this same MDP.}

\rev{This distinction allows the analytical and practical roles of the temporal horizon to be separated cleanly: the batch-MDP analysis (Theorems~\ref{thm:eval} and~\ref{thm:eval-h}) isolates the conditions required for exact latent-to-physical correspondence, while the longer observation gap provides a substantially more informative representation for the visual domains studied here.}

\paragraph{First-step execution fidelity.}
Although the $H=3$ latent represents an endpoint three steps ahead, the deployed controller executes only the first decoded action before replanning. We therefore directly evaluate whether this longer-horizon representation preserves accurate one-step execution. On 100 held-out transitions per environment, we encode the corresponding latent, decode the action chunk, execute only its first action, and measure the resulting next-state cube-position error. The deployed $H=3$ configuration achieves lower error than the $H=1$ configuration on both Cube (1.51 vs.\ 3.26 mm) and Scene (0.80 vs.\ 1.22 mm). The same trend holds on transitions where the cube actually moves, with 3.59 vs.\ 7.69 mm on Cube and 4.20 vs.\ 6.42 mm on Scene (Table~\ref{tab:first-step-fidelity}). These results provide direct evidence that the practical $H=3$ representation can preserve accurate first-step execution of the measured object transition despite encoding a longer-horizon endpoint, supporting its use with one-step receding-horizon control. This diagnostic compares existing configurations using held-out dataset latents and a geometric next-state metric; the configurations differ in temporal gap, decoder chunking, and training augmentation.

\begin{table}[htbp]
\centering

\setlength{\tabcolsep}{5pt}
\renewcommand{\arraystretch}{1.12}
\caption{First-step cube-position error (mm, $\downarrow$) on matched held-out transitions using seed-0 models ($C=H$, $E=1$). Moving subsets have recorded cube displacement $>1$ mm.}
\label{tab:first-step-fidelity}
\begin{tabular}{llrcc}
\toprule
Environment & Subset & $n$ & $H=1$ & $H=3$ \\
\midrule
\multirow{2}{*}{\nolinkurl{v-cube-single-play}} & All & 100 & 3.26 & \textbf{1.51} \\
 & Moving & 42 & 7.69 & \textbf{3.59} \\
\addlinespace[4pt]
\multirow{2}{*}{\nolinkurl{v-scene-play}} & All & 100 & 1.22 & \textbf{0.80} \\
 & Moving & 19 & 6.42 & \textbf{4.20} \\
\bottomrule
\end{tabular}
\end{table}

\subsection{Learned value ordering and critic-guided reweighting}\label{app:critic-ordering}

Proposition~\ref{prop:supported-improve} formalizes policy improvement as reweighting a supported reference distribution toward higher-value latent actions. LAPIS implements this principle through the learned critic and the gate in Equation~\eqref{eq:gate}. For a fixed critic and threshold snapshot, let $q_{\B}(z\mid s,g)$ denote the conditional distribution of observed latent transitions presented to the actor, including its goal-relabeling procedure. Define the selected mass
\[
Z_G(s,g)=\E_{z\sim q_{\B}(\cdot\mid s,g)}\left[G(s,z,g)\right],
\]
and, whenever $Z_G(s,g)>0$, the gated distribution
\begin{equation}\label{eq:gated-target-law}
q_G(z\mid s,g)=\frac{q_{\B}(z\mid s,g)G(s,z,g)}{Z_G(s,g)}.
\end{equation}
Because the gate only removes examples,
\[
\supp q_G(\cdot\mid s,g)\subseteq\supp q_{\B}(\cdot\mid s,g)\subseteq Z_{\B}(s)
\]
at sampled state--goal pairs. Thus, at the target-distribution level, critic-guided filtering is exactly a supported reweighting of observed state--latent transitions.

The improvement result does not require the critic score to be a calibrated numerical estimate of $Q$; it requires the score to preserve the ordering of the reference action values over $z$ at fixed $(s,g)$. For the ideal binary density-ratio objective in Remark~\ref{rem:score-order},
\[
f^\star(s,z,g)=\log Q^\mu(s,z,g)-\log p_{\B}(g),
\]
and the second term is independent of $z$. Hence $f^\star$ preserves the ordering of $Q^\mu$, and thresholding it at any $z$-independent state--goal threshold produces weights that are nondecreasing in value.

In the supported comparison of Appendix~\ref{app:proofs-gate}, take the reference policy to be
\[
\mu(\cdot\mid s,g)=q_{\B}(\cdot\mid s,g),
\]
with $q_G=\mu$ when $Z_G(s,g)=0$. Whenever the score preserves the ordering of
\[
Q^\mu_{\MBz}(s,\cdot,g),
\]
Corollary~\ref{cor:gate} and Proposition~\ref{prop:supported-improve} give
\[
J_{\MBz}(q_G;g)\ge J_{\MBz}(q_{\B};g).
\]
Under the conditions of Theorem~\ref{thm:eval}, this ordering transfers to batch-exact decoded execution, while Corollary~\ref{cor:stoch-improve} provides the corresponding approximate-transfer statement for learned decoding.

\rev{Practical LAPIS uses the TD-InfoNCE critic and the learned score-space threshold described in Appendix~\ref{app:tdinfonce}. The gate is applied only to observed state--latent transitions, so for every fixed critic and threshold snapshot it defines an exact supported reweighting $q_G$ of the actor-training distribution $q_{\B}$; this support property holds for any score. During joint training, the critic and hence the reweighted target evolve, while the actor is trained toward the target defined by the current gate. Whether this reweighting also improves the reference depends on whether the learned score preserves the reference-value ordering over $z$, which the TD-InfoNCE objective does not determine, as follows.}

\rev{\paragraph{Score offset in TD-InfoNCE.} Both cross-entropy terms in Equation~\eqref{eq:td-loss} are softmaxes over candidate goals within the same row $(s_i,z_i)$, and the bootstrap weights $W_{ij}$ are softmaxes over goals as well. Adding any function $c(s,z)$ to the score therefore leaves the loss unchanged, so the optimum is determined only up to such an offset, as noted for TD-InfoNCE by \citet[Sec.~3.1]{zheng2024cdpc}. An offset that depends on $z$ can change the ordering of latents at fixed $(s,g)$ and hence the raw-score gate in Equation~\eqref{eq:gate}. The L$_2$-normalized bilinear parameterization of Appendix~\ref{app:tdinfonce} bounds the score, $|f|\le1/\tau_{\mathrm{rep}}$, but does not remove the offset: a component of the normalized goal features shared across goals contributes a term that depends only on $(s,z)$. Neither the ensemble minimum nor the threshold $\hat V(s,g)$, which does not depend on $z$, removes it. Preservation of value ordering by the learned score is therefore an empirical property of the trained critic, not a consequence of the objective.}

\rev{\paragraph{Offset-free score.} Let $p_c$ denote the critic's candidate-goal distribution (Appendix~\ref{app:tdinfonce}) and define
\[
\bar f(s,z,g)=f(s,z,g)-\log\E_{g'\sim p_c}\bigl[\exp f(s,z,g')\bigr].
\]}
\begin{revblock}
\begin{remark}[Offset-free scores and value ordering]\label{rem:score-order-bar}
Suppose the optimal TD-InfoNCE critic satisfies $\exp f^\star(s,z,g)=e^{c(s,z)}Q^\mu(s,z,g)/p_c(g)$ for an arbitrary $c$, where $Q^\mu(s,z,\cdot)$ is the discounted future-state occupancy density~\citep[Sec.~3.1]{zheng2024cdpc} and $p_c>0$ on its support. Since $\E_{g'\sim p_c}[Q^\mu(s,z,g')/p_c(g')]=1$, we have $\log\E_{p_c}[\exp f^\star(s,z,g')]=c(s,z)$, so
\[
\bar f^\star(s,z,g)=\log Q^\mu(s,z,g)-\log p_c(g)
\]
for every offset $c$. At fixed $(s,g)$ the last term does not depend on $z$, so $\bar f^\star$ preserves the ordering of $Q^\mu$. Remark~\ref{rem:score-order}, Corollary~\ref{cor:gate}, and Proposition~\ref{prop:supported-improve} therefore apply with $\bar f^\star$ in place of the binary density-ratio score $f^\star$.
\end{remark}
\end{revblock}

\rev{The log-partition in $\bar f$ can be estimated by a log-sum-exp over the candidate goals in a minibatch, the same normalization TD-InfoNCE uses for its importance weights, but this finite-sample estimate is biased and noisy. We therefore measure how closely the raw score $f$, which the gate uses, follows $\bar f$ in trained LAPIS critics (Table~\ref{tab:score-offset} in Section~\ref{sec:exp-offset}). The protocol for that table is as follows. We use \texttt{antmaze-medium}, \texttt{cube-single}, and \texttt{scene} with four seeds each, average the $300$k, $400$k, and $500$k checkpoints within each seed, and report the mean over seeds. For each training state, we score $16$ candidate latents, the dataset latent and $15$ samples from the behavior branch, and estimate the log-partition of $\bar f$ with $16{,}384$ goals from $p_c$, averaged over four independent draws. The threshold rule, a stand-in for $\hat V$, keeps the dataset latent if its score exceeds the $0.9$-expectile of the scores of the $15$ sampled latents. Standard deviations across seeds are below $0.01$ for the correlation and below $1\%$ for the agreement. Table~\ref{tab:ablation}(b) compares the TD-InfoNCE critic with an IQL critic, whose regression targets make values directly comparable across $z$, under fixed extraction rules.}

Consistent with this supported-update view, Table~\ref{tab:ablation}\rev{(a)} shows that replacing direct max-$Q$ actor optimization with critic-guided filtering improves performance for both deterministic and flow actors.

\subsection{Flow-based extraction and guided inference}\label{app:extraction-gap}

The remaining correspondence is between the supported target distribution defined by critic-guided filtering and the learned flow policy. For a fixed gate snapshot, define
\[
\ell_{\theta}^{(o)}(s,z,g)=\mathbb{E}_{x_0,t}\left[\left\|v_\theta(x_t,t;s,g,o)-(z-x_0)\right\|_2^2\right],
\qquad x_t=(1-t)x_0+tz,
\]
where noise and time are sampled as in Equation~\eqref{eq:actor}. Conditioning the actor objective on $(s,g)$ gives
\begin{equation}\label{eq:gated-flow-identity}
\begin{aligned}
\mathcal{L}_{\mathrm{actor}}(s,g)
&=\mathbb{E}_{q_{\B}}\left[G\,\ell_{\theta}^{(1)}\right]+\lambda\mathbb{E}_{q_{\B}}\left[\ell_{\theta}^{(0)}\right]\\
&=Z_G(s,g)\mathbb{E}_{q_G}\left[\ell_{\theta}^{(1)}\right]+\lambda\mathbb{E}_{q_{\B}}\left[\ell_{\theta}^{(0)}\right].
\end{aligned}
\end{equation}
Thus, the high-value branch is trained toward $q_G$, while the behavior branch is trained toward $q_{\B}$. The full loss averages over sampled $(s,g)$, with $Z_G(s,g)$ retaining the original weighting across state--goal contexts. When $Z_G(s,g)=0$, the gated loss term vanishes; the reference-policy fallback in Appendix~\ref{app:critic-ordering} is used only to define the ideal policy comparison. Equation~\eqref{eq:gated-flow-identity} therefore gives the direct correspondence between the practical flow-matching objective and the supported reweighting analyzed in Appendix~\ref{app:proofs-gate}.

The support inclusion established in Appendix~\ref{app:critic-ordering} is exact for the target distributions $q_{\B}$ and $q_G$. A finite-capacity neural flow approximates these conditional distributions while sharing statistical strength across visual states. At inference, generated continuous latents are projected to the learned product codebook before decoding. This projection restricts samples to the learned code space, but does not by itself guarantee state-conditioned support; that depends on how faithfully the learned policy represents its conditional target.

Classifier-free guidance combines the two learned velocity fields,
\[
v_w^{\mathrm{CFG}}=(1-w)v_\theta(\cdot;0)+wv_\theta(\cdot;1).
\]
At $w=0$, inference uses the branch trained toward $q_{\B}$, while $w=1$ uses the branch trained toward $q_G$. Values $w>1$ extrapolate the velocity field toward the high-value branch and therefore need not preserve the exact support guarantee of $q_G$. This extrapolation differs from direct max-$Q$ actor optimization, however. Direct max-$Q$ updates the selected latent by following critic values in the continuous latent space, including regions where the critic is not grounded by observed transitions. In LAPIS, the critic is used first to select high-value examples only among observed state--latent transitions, producing $q_G\subseteq q_{\B}$; guidance then extrapolates between policy fields trained on $q_{\B}$ and this supported high-value reweighting. The guidance direction is therefore anchored to distributions constructed entirely from observed state--latent transitions, although sufficiently strong extrapolation can still move generated samples away from their exact support.

The supported-reweighting result thus identifies $q_G$ at $w=1$ as the theoretical anchor, while $w>1$ provides an inference-time extension that trades stricter adherence to this target for stronger emphasis on the high-value branch. The guidance scale $w$ is distinct from the distributional exponent $\eta$ in Proposition~\ref{prop:supported-improve}, and the supported-improvement guarantee does not by itself extend to arbitrary $w>1$.

The experiments provide complementary evidence for these stages. Table~\ref{tab:ablation}\rev{(a)} shows that the flow actor improves policy extraction when paired with critic-guided filtering. Table~\ref{tab:lapis-guidance-scale} further separates the supported high-value branch from guidance extrapolation. Across the visual manipulation tasks, $w=1$, which uses the branch trained toward $q_G$, already captures most of the improvement from the behavior branch $w=0$ to the default $w=3$ policy. Stronger guidance therefore provides an additional empirical refinement rather than being solely responsible for the gains of LAPIS. Section~\ref{sec:exp-diag} further compares the effect of stronger direct max-$Q$ optimization and stronger guidance through behavior-support distance $d_{\mathrm{BC}}$, realization round-trip error $d_{\mathrm{round}}$, and task success. As CRL becomes more aggressive, its support distance and realization error increase while success collapses, whereas LAPIS remains closer to the behavior-supported region over the guidance sweep. These results show that guided inference stays closer to the behavior-supported region than unconstrained critic maximization in this comparison.

The direct CFGRL comparison in Appendix~\ref{app:additional-ablations} further illustrates why this
distinction matters in a learned latent action space. Goal-conditioned CFGRL
benefits from increasing guidance when applied to physical actions, most clearly
on AntMaze, but provides no consistent improvement when the same construction
is applied directly to learned latent actions. In contrast, LAPIS obtains large gains already at $w=1$, before extrapolative guidance, with its high-value branch trained toward the critic-filtered supported target $q_G$ rather than the unfiltered goal-conditioned behavior distribution. Together with the component ablation in Table~\ref{tab:ablation}\rev{(a)}, this provides empirical evidence that support-aware target construction, rather than guidance alone, is consequential for latent policy improvement.

\begin{revblock}
\subsection{Realization error and return}\label{app:realization-return}
Proposition~\ref{prop:stoch-transfer} measures execution mismatch in total variation, which equals one between two distinct point masses even when the two successors are geometrically close. The analysis of Section~\ref{sec:cfg-crl} restricts policies to observed state--latent pairs, where $\delta(s,z)=0$ (Definition~\ref{def:realizable}). Here we bound how far latent and decoded returns can separate when the executed latents are not realized exactly, under Lipschitz conditions that are standard in the analysis of latent and model-based RL~\citep{asadi2018lipschitz,gelada2019deepmdp}.

Let $\M^F$ denote the latent outcome process that, from state $s$ and latent $z$, moves to the predicted successor $F(s,z)$ with reward $r(s,F(s,z))$. For a decoder $\widehat\Psi(a\mid s,z)$, state-conditioned or not, define its expected one-step round-trip error and its excess over the realization error,
\[
\eta_{\widehat\Psi}(s,z)=\E_{a\sim\widehat\Psi(\cdot\mid s,z)}\bigl\|I\bigl(s,T(s,a)\bigr)-z\bigr\|_2,
\qquad
\epsilon_{\widehat\Psi}(s,z)=\eta_{\widehat\Psi}(s,z)-\delta(s,z)\ \ge 0,
\]
where the inequality holds because $T(s,a)\in\mathcal R(s)$.

\begin{proposition}[Return gap under realization error]\label{prop:realization-gap}
Suppose Assumptions~\ref{ass:detT} and~\ref{ass:reward} hold, and
(i) the forward model is accurate on reachable outcomes up to $\epsilon_F\ge 0$, $\|F(s,I(s,s'))-s'\|_2\le\epsilon_F$ for every $s'\in\mathcal R(s)$;
(ii) $F(s,\cdot)$ is $L_F$-Lipschitz, $\|F(s,z)-F(s,z')\|_2\le L_F\|z-z'\|_2$;
(iii) $r(s,\cdot)$, extended to all states with $|r|\le R_{\max}$, is $L_r$-Lipschitz, and the latent value $V^{\pi_z}_{\M^F}$ is $L_V$-Lipschitz in the state.
Then, for any latent policy $\pi_z$ and decoder $\widehat\Psi$,
\[
\bigl|J_{\M^F}(\pi_z)-J_\M(\widehat\Psi\circ\pi_z)\bigr|
\le (L_r+\gamma L_V)\Bigl[L_F\sum_{t\ge 0}\gamma^t\,\E\bigl[\delta(s_t,z_t)+\epsilon_{\widehat\Psi}(s_t,z_t)\bigr]+\frac{\epsilon_F}{1-\gamma}\Bigr],
\]
where the expectation is over decoded execution of $\widehat\Psi\circ\pi_z$ in $\M$ from $\mu_0$.
\end{proposition}

\begin{proof}
Write $V_F=V^{\pi_z}_{\M^F}$ and $V_E=V^{\widehat\Psi\circ\pi_z}_{\M}$. For a state $s$, with $z\sim\pi_z(\cdot\mid s)$, $a\sim\widehat\Psi(\cdot\mid s,z)$ and $s'=T(s,a)$, the Bellman equations give
\[
V_F(s)-V_E(s)=\E\Bigl[r\bigl(s,F(s,z)\bigr)-r(s,s')+\gamma\bigl(V_F(F(s,z))-V_F(s')\bigr)+\gamma\bigl(V_F(s')-V_E(s')\bigr)\Bigr].
\]
Since $s'\in\mathcal R(s)$, (i) and (ii) give
\[
\|F(s,z)-s'\|_2\le\|F(s,z)-F(s,I(s,s'))\|_2+\|F(s,I(s,s'))-s'\|_2\le L_F\|z-I(s,s')\|_2+\epsilon_F.
\]
By (iii), the first two terms are bounded in absolute value by $(L_r+\gamma L_V)\bigl(L_F\|z-I(s,s')\|_2+\epsilon_F\bigr)$, and the conditional expectation of $\|z-I(s,s')\|_2$ given $(s,z)$ is $\eta_{\widehat\Psi}(s,z)=\delta(s,z)+\epsilon_{\widehat\Psi}(s,z)$. Hence
\[
|V_F(s)-V_E(s)|\le (L_r+\gamma L_V)\Bigl(L_F\,\E\bigl[\delta(s,z)+\epsilon_{\widehat\Psi}(s,z)\bigr]+\epsilon_F\Bigr)+\gamma\,\E\bigl|V_F(s')-V_E(s')\bigr|.
\]
Unrolling this inequality along decoded execution, using that both values are bounded by $R_{\max}/(1-\gamma)$ so the remainder vanishes, and averaging over $s_0\sim\mu_0$ gives the claim.
\end{proof}

\paragraph{Interpretation.} The bound has three terms. The realization error $\delta$ depends only on which latents the policy selects; it is zero for latents of transitions observed at the current state. The excess $\epsilon_{\widehat\Psi}$ depends only on the decoder; with the batch-exact decoder $\Psi^*$, execution of a batch-constrained policy stays on observed transitions, and both terms vanish. The last term is the forward model's own error and does not depend on the policy. With a finite codebook, distinct reachable outcomes can share a latent, so $\epsilon_F>0$ in general; $\epsilon_F=0$ would require a global form of supported outcome identifiability (Definition~\ref{def:supp-id}). Theorem~\ref{thm:eval} states the batch-constrained case exactly for $\MBz$, without a forward model. For a fixed forward model, decoder and Lipschitz constants, the bound therefore grows linearly with the realization error of the latents the policy selects. The $H$-step analogue of $\eta_{\widehat\Psi}$ is the round-trip error $d_{\mathrm{round}}$ of Figure~\ref{fig:diagnostic}. As for Lipschitz bounds in model-based RL, condition (iii) is an idealization; in particular, it excludes 0/1 success rewards, so for our tasks the statement is qualitative. Proposition~\ref{prop:realization-gap} is stated for one-step latents, and the same argument extends it to $H$-step latents, with the $H$-step transition and the discounted $H$-step reward in place of $T$ and $r$, and $\gamma^H$ in place of $\gamma$.
\end{revblock}

\newpage
\section{Experimental details and additional analyses}\label{app:expdetails}

This appendix provides the experimental setup and supplementary analyses. We describe the benchmarks and evaluation protocols (Appendix~\ref{app:data-evaluation}), latent-action representation learning and decoding (\ref{app:lam-decoder}), LAPIS critic and policy learning (\ref{app:tdinfonce}), and baseline implementations (\ref{app:baseline-details}). We then report additional sampling, control, and guidance analyses (\ref{app:additional-ablations}), analyze action decodability and temporal-gap effects in the learned latent representation (\ref{app:zar2}), and present qualitative rollouts (\ref{sec:exp-qualitative}).

\subsection{Datasets and evaluation}\label{app:data-evaluation}

\paragraph{Benchmarks and label regimes.}
We use the OGBench~\citep{park2025ogbench} visual \texttt{cube}, \texttt{scene}, and \texttt{antmaze-navigate} domains with $64\times64$ RGB observations. Manipulation actions have five dimensions (end-effector position change, yaw change, and gripper control), while antmaze actions are eight-dimensional joint torques. For the ablations in Tables~\ref{tab:ablation}--\rev{\ref{tab:score-offset}}, we use one representative task from each family: \nolinkurl{visual-cube-single-play}, \nolinkurl{visual-scene-play}, and \nolinkurl{visual-antmaze-medium-navigate}\rev{; Table~\ref{tab:ablation}(b) also includes \nolinkurl{visual-cube-double-play}}. The diagnostics in Figure~\ref{fig:diagnostic} focus on \nolinkurl{visual-antmaze-medium-navigate}. Latent-action methods use all observed transitions for LAM and latent-policy learning, while the data used by the action-scarce baselines are specified in Appendix~\ref{app:baseline-details}.

\begin{figure}[htbp]
\centering
\includegraphics[width=0.62\textwidth,height=0.30\textheight,keepaspectratio]{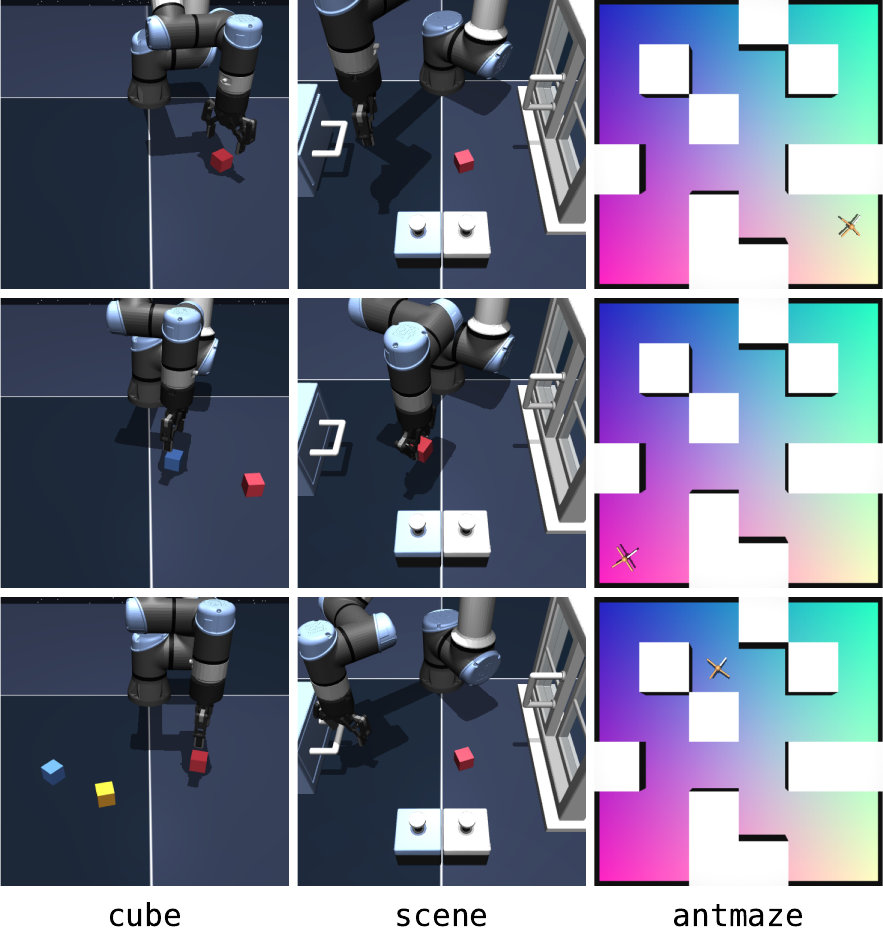}
\caption{Example observations, rendered from our own environment instances, spanning the \texttt{cube} and \texttt{scene} manipulation families and the \texttt{antmaze} navigation family.}
\label{fig:envs}
\end{figure}

\paragraph{OGBench evaluation.}
Each of four seeds is evaluated on $20$ episodes per task at the $300$k, $400$k, and $500$k checkpoints. We average these checkpoint scores within each seed, then report the mean and standard deviation across seeds. \rev{All entries of Table~\ref{tab:main} follow this protocol.}

\paragraph{Visual Franka Kitchen.}\label{app:kitchen}
Kitchen uses $64\times64$ rendered RGB observations, nine-dimensional bounded actions, and the original D4RL simulator with a $280$-step episode limit. The \texttt{partial} variant targets microwave, kettle, light switch, and slide cabinet; \texttt{mixed} targets microwave, kettle, bottom burner, and light switch. Goal images are rendered with the evaluation camera after setting the required objects to their target configurations and the robot to its initial posture. Completed subtasks are removed from the remaining-task set, and an episode terminates when all required subtasks are completed or the time limit is reached.

Table~\ref{tab:kitchen} reports the percentage of required subtasks completed, averaged over evaluation episodes:
\[
\bar C=\frac{1}{N_{\mathrm{eval}}}\sum_{j=1}^{N_{\mathrm{eval}}}100\frac{n_{c,j}}{N_j},
\]
where $n_{c,j}$ of the $N_j=4$ required subtasks are completed in episode $j$. All latent policies use the canonical \texttt{partial} training split and are evaluated against both target-task sets; \texttt{mixed} does not denote a separately trained latent policy. Each seed uses its own frozen LAM and decoder trained for $500$k updates, with the $H=3$, $C=3$, $E=1$ receding-horizon setup described in Appendix~\ref{app:macro-align}. For seeds $0$--$3$, we average the $300$k, $400$k, and $500$k policy-checkpoint scores ($20$ episodes per target-task set per checkpoint), then report the mean and population standard deviation across the four seed averages. Table~\ref{tab:kitchen} reports LAPIS at guidance $w=3$; the latent LAPIS run uses a TD-InfoNCE critic and a log-sum-exp value baseline. The real-action oracle results are reported separately from the latent policies.

\begin{figure}[htbp]
\centering
\includegraphics[width=0.95\textwidth]{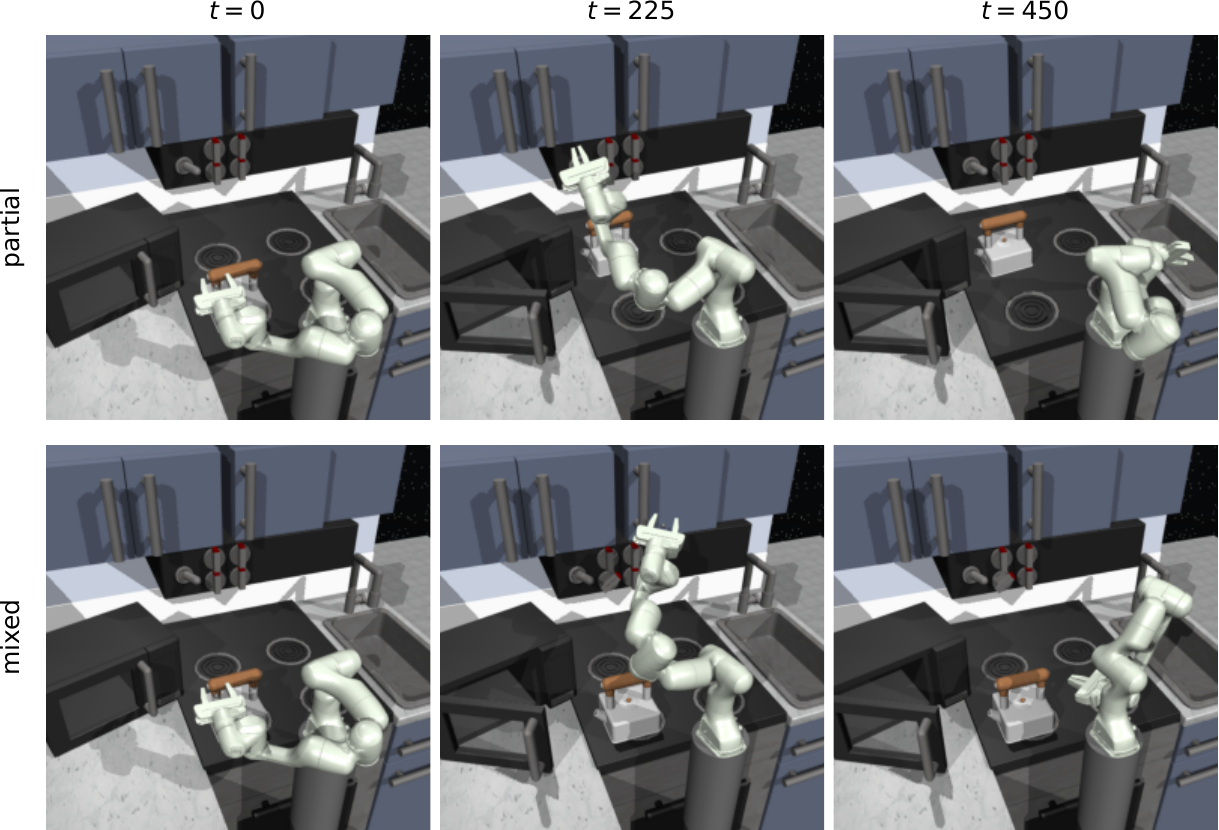}
\caption{Example observations from visual Franka Kitchen. Rows show \texttt{partial} (top) and \texttt{mixed} (bottom); columns show steps $t=0,225,450$ of illustrative LAPIS rollouts with guidance $w=3$ and reset seed $0$. The displayed simulator states are rendered at $512\times512$ using the evaluation camera, while policy observations remain $64\times64$. These illustrative $450$-step rollouts are separate from the $280$-step D4RL evaluations reported in Table~\ref{tab:kitchen}.}
\label{fig:kitchen-envs}
\end{figure}
\FloatBarrier

\subsection{Latent action model and decoder}\label{app:lam-decoder}

\paragraph{Representation learning.}
The IDM/FDM pair is a product-quantized transition autoencoder trained by pixel reconstruction with independent cross-crop inputs. It uses single-frame observations, temporal gap $H=3$, and IDM encoder scale $4$. The codebook configuration is $M{:}H_z{:}D{:}K=2{:}4{:}8{:}128$: two codebooks, four slots per codebook, eight embedding coordinates, and $128$ entries per codebook, giving $64$ flattened latent coordinates. We denote the continuous encoder output by $z_e$ and its quantization by $z_q$.

\paragraph{Latent coordinates and codebook projection.}\label{app:latent-coordinates}
The policy, critic, and decoder use quantized latents standardized per codebook and embedding dimension:
\begin{equation}\label{eq:latent-standardization}
[\mathcal N(z_q)]_{mhd}=\frac{(z_q)_{mhd}-\mu_{md}}{\sigma_{md}},
\end{equation}
where the frozen means $\mu_{md}$ and stabilized standard deviations $\sigma_{md}$ are shared across slots $h$. The same transform is applied to the codebook entries. Generated continuous latents are projected to the nearest product-codebook entries in these coordinates before decoding.

\paragraph{Action decoding.}
The experimental state-marginal flow decoder $\Psi(\mathbf a\mid z)$ is trained on labeled examples to predict three-action chunks from standardized quantized latents and is frozen for deployment. Predicted actions are clipped to the environment bounds and executed with the receding-horizon procedure of Appendix~\ref{app:macro-align}. The state-conditioned variant in Table~\ref{tab:decoder-cond} additionally takes $s$ as input. For Table~\ref{tab:decoder-cond}, we keep the same frozen $5\%$-label LAM across all decoder-label budgets and vary only decoder supervision. 

\subsection{Critic and policy learning}\label{app:tdinfonce}

\paragraph{Shared policy-learning settings.}
Table~\ref{tab:domain-overrides} consolidates the optimization, architecture, critic, flow-policy, and evaluation hyperparameters used by LAPIS on OGBench. In Table~\ref{tab:main}, the physical-action chunk sizes $C$ for F-GCBC, GCIQL, CRL, and LAPIS are $(3,1,1,3)$, respectively, while their latent-action counterparts use $C=3$. All latent-action methods use the $H=3$ LAM of Appendix~\ref{app:lam-decoder}, and all settings execute only the first physical action before replanning ($E=1$; Appendix~\ref{app:macro-align}).

\begin{table}[htbp]
\centering

\setlength{\tabcolsep}{4pt}
\caption{LAPIS hyperparameters for OGBench policy learning.}
\label{tab:domain-overrides}
\begin{tabular}{@{}lcc@{}}
\toprule
Setting & Visual manipulation & Visual antmaze \\
\midrule
Adam learning rate                    & $3\times10^{-4}$ & $3\times10^{-4}$ \\
Batch size                            & $256$   & $256$ \\
Gradient steps                        & $500$k  & $500$k \\
Visual encoder                        & small IMPALA & small IMPALA \\
Policy MLP trunk                      & $(512,512,512)$ & $(512,512,512)$ \\
Layer normalization                   & yes     & yes \\
$p_{\mathrm{aug}}$ (random crop)      & $0.5$   & $0.0$ \\
Discount $\gamma$                     & $0.99$  & $0.99$ ($0.995$ giant) \\
LAPIS critic                          & TD-InfoNCE & TD-InfoNCE \\
Critic ensemble size                  & $2$     & $2$ \\
Critic feature dimension              & $512$   & $512$ \\
Critic temperature $\tau_{\mathrm{rep}}$ & $0.1$ & $0.1$ \\
Gate expectile $\tau$                 & $0.9$   & $0.9$ \\
Flow time-embedding dimension         & $64$    & $64$ \\
Flow training-time distribution       & $\mathrm{Beta}(1.5,1.0)$ & $\mathrm{Beta}(1.5,1.0)$ \\
Maximum training time $t_{\max}$      & $0.999$ & $0.999$ \\
Flow sampling steps                   & $10$ Euler & $10$ Euler \\
Behavior-branch weight $\lambda$      & $0.1$   & $0.1$ \\
Evaluation guidance $w$               & $3$     & $3$ \\
Evaluation episodes                   & $20$    & $20$ \\
\bottomrule
\end{tabular}
\end{table}

\paragraph{Categorical TD objective.}
LAPIS uses categorical TD-InfoNCE on both visual manipulation and antmaze. For a minibatch of $n$ recorded transitions, let $s_i^+=s_{t_i+1}$ and let $z_i^+$ be the LAM label inferred from its recorded continuation. Critic candidate goals are sampled from a mixture of current states, future states in recorded trajectories, and random dataset states. With ensemble index $e\in\{1,2\}$, define
\begin{align*}
L^{\mathrm{next}}_{ije}&=f_{\phi,e}(s_i,z_i,s_j^+),&
L^{\mathrm{goal}}_{ije}&=f_{\phi,e}(s_i,z_i,g_j),\\
W_{ij}&=\operatorname{sg}\!\left[\operatorname{softmax}_{j}\!\left(\min_e f_{\bar\phi,e}(s_i^+,z_i^+,g_j)\right)\right],
\end{align*}
where $\operatorname{sg}$ denotes stop-gradient and $\bar\phi$ is the target critic. For $\operatorname{CE}(\ell,y)=-\sum_j y_j\log\operatorname{softmax}_j(\ell)$ and $I_{ij}=\mathbf1[i=j]$, the loss is
\begin{equation}\label{eq:td-loss}
\mathcal L_{\mathrm{TD}}=\frac1{2n}\sum_{i,e}\left[(1-\gamma)\operatorname{CE}(L^{\mathrm{next}}_{i,:,e},I_{i,:})+\gamma\operatorname{CE}(L^{\mathrm{goal}}_{i,:,e},W_{i,:})\right].
\end{equation}
The first term predicts immediate successors; the second propagates future-state occupancy through behavior continuations using a detached categorical bootstrap~\citep{zheng2024cdpc}. Appendix~\ref{app:goal-sampling-ablation} reports the goal-candidate sampling ablation.

\paragraph{Score-space gate.}
The score $f_\phi$ in Equation~\eqref{eq:gate} is the ensemble minimum $f_{\min}(s,z,g)=\min_e f_{\phi,e}(s,z,g)$. For actor-training goals $g$, we fit the state--goal threshold by expectile regression,
\begin{equation}\label{eq:score-value}
\mathcal L_{\hat V}=\E\left[\rho_\tau\!\left(\operatorname{sg}[f_{\min}(s,z,g)]-\hat V(s,g)\right)\right],\qquad
\rho_\tau(u)=|\tau-\mathbf1[u<0]|u^2,
\end{equation}
with the expectile $\tau$ specified in Table~\ref{tab:domain-overrides}. The gate is $\operatorname{sg}[\mathbf1[f_{\min}>\hat V]]$. The critic, score threshold, and actor are updated jointly as in Algorithm~\ref{alg:cfgcrl}.

\paragraph{Critic features and flow policy.}
The bilinear critic uses L$_2$-normalized features with temperature $\tau_{\mathrm{rep}}$:
\[
f_{\phi,e}(s,z,g)=\frac{\langle\phi_e(s,z)/\|\phi_e(s,z)\|_2,\;\psi_e(g)/\|\psi_e(g)\|_2\rangle}{\tau_{\mathrm{rep}}}.
\]
The critic dimensions and all flow-training and guided-inference hyperparameters are summarized in Table~\ref{tab:domain-overrides}.
\FloatBarrier

\subsection{Baseline implementations}\label{app:baseline-details}

\paragraph{Table~\ref{tab:main} baselines.}
\rev{All entries of Table~\ref{tab:main} are from our runs. The fully supervised GCBC, GCIVL, GCIQL, QRL, and HIQL use the OGBench implementations and pixel-based hyperparameters~\citep{park2025ogbench}. With $5\%$ labels, GCIVL, QRL, and HIQL use the same implementations and hyperparameters in the latent action space, with the same LAM and decoder as LAPIS. LAPIS$^\dagger$ is described in Appendix~\ref{app:extraction-grid}.} F-GCBC replaces the Gaussian GCBC head with our flow-matching actor and is not an OGBench baseline. GCIQL in Table~\ref{tab:main} follows standard goal-conditioned IQL, learning an action-conditioned critic $Q(s,a,g)$ and value function $V(s,g)$ and extracting the actor from their advantage. CRL follows the upstream OGBench implementation with a DDPG+BC actor and InfoMax/InfoNCE critic, with hyperparameters tuned per domain. The domain-specific BC coefficient is reported in Table~\ref{tab:crl-overrides}. F-GCBC, CRL, and LAPIS use the shared visual encoder, batch size, and augmentation settings within a domain. The method-specific action-chunk sizes used in Table~\ref{tab:main} are summarized in Appendix~\ref{app:tdinfonce}, and their sensitivity is examined in Table~\ref{tab:real-action-chunk}.

\begin{table}[htbp]
\centering

\caption{CRL-specific OGBench hyperparameter.}
\label{tab:crl-overrides}
\begin{tabular}{@{}lcc@{}}
\toprule
Setting & Visual manipulation & Visual antmaze \\
\midrule
BC coefficient $\alpha$ & $3.0$ & $0.1$ \\
\bottomrule
\end{tabular}
\end{table}

\paragraph{Action-scarce baselines.}
Table~\ref{tab:semisup} uses goal-conditioned physical-action adaptations designed for limited action supervision. In particular, its \rev{GCIVL entry, a value-only variant that the original submission labeled GCIQL,} is distinct from the standard action-conditioned GCIQL used in Table~\ref{tab:main}; Appendix~\ref{app:baseline-protocol-discussion} discusses this protocol difference. Let $\Dlab$ contain genuinely labeled transitions and $\B$ the action-free transitions, with goal-reaching reward $r(s,g)$. We use the following value-learning, labeled-action, and inverse-prediction objectives:
\begin{align*}
\mathcal L_V&=\E_{s,s',g}\!\left[L_2^\tau\bigl(r(s,g)+\gamma\bar V(s',g)-V(s,g)\bigr)\right],\\
L_2^\tau(u)&=|\tau-\mathbf1[u<0]|u^2,\\
\mathcal L^{\mathrm{AWR}}_\pi&=\E_{(s,a,g)\sim\Dlab}\!\left[-\exp(\alpha A)\log\pi(a\mid s,g)\right],\\
\mathcal L_{\mathrm{IDM}}&=\E_{(s,a,s')\sim\Dlab}\|I_a(s,s')-a\|_2^2,\qquad \hat a=I_a(s,s')\text{ on }\B,\\
\mathcal L_{\mathrm{BC}}&=\E[-\log\pi(a\mid s,g)],
\end{align*}
where $\bar V$ is the target value and $\tau$ is the upper expectile. The methods combine these objectives as follows:
\begin{itemize}
\item \textbf{\rev{GCIVL}}~\rev{\citep{park2025ogbench,kostrikov2022iql}}: $\mathcal L_V$ plus a flat AWR actor with the action-free advantage $A=V(s',g)-V(s,g)$. \rev{It learns no action-conditioned $Q$ and corresponds to GCIVL in OGBench.}
\item \textbf{HIQL}~\citep{park2023hiql}: $\mathcal L_V$ plus two AWR actors---a high-level actor over subgoals ($A^h=V(s_{t+k},g)-V(s,g)$) and a low-level actor over actions toward the subgoal ($A^\ell=V(s',s_{t+k})-V(s,s_{t+k})$, on $\Dlab$). \rev{We use the OGBench HIQL implementation with its learned subgoal representation, which the high-level actor predicts and the low-level actor conditions on. All HIQL runs in Table~\ref{tab:semisup} use this representation.}
\item \textbf{VPT-RL (VPT-style action inference)}~\citep{baker2022vpt}: train $I_a$ via $\mathcal L_{\mathrm{IDM}}$, pseudo-label $\B$, then extract a value-guided (classifier-free-guided flow) RL policy from the (pseudo-)labeled real actions---the LAPIS policy objective of Equation~\eqref{eq:actor} applied to actions instead of latents.
\item \textbf{SS-ORL}~\citep{zheng2023semiaf}: offline RL ($\mathcal L_V$ + AWR actor) on the union $\Dlab\cup\{(s,\hat a,s'):(s,s')\in\B\}$ of true and IDM-pseudo-labeled actions.
\end{itemize}
Action-free value objectives use all transitions. \rev{GCIVL}'s physical-action actor and HIQL's low-level actor use genuine labels; HIQL's subgoal learning also uses action-free trajectories. The inverse-action predictor is fitted on genuine labels, and pseudo-label-based policy objectives use the union of genuine and inferred actions. Architecture and optimization settings follow the shared recipe where applicable.

\subsection{Additional control and sampling ablations}\label{app:additional-ablations}

\paragraph{Goal-candidate sampling.}\label{app:goal-sampling-ablation}
Because the LAM represents $H=3$ endpoint transitions, we test whether future-state candidates for critic training should also be restricted to the same temporal grid. We compare the standard sampling scheme, which allows arbitrary future offsets, with a grid-restricted variant that rounds each trajectory-future offset $k$ to $H\max(1,\mathrm{round}(k/H))$, with $H=3$. Current-state and random-state candidates, actor-goal sampling, the next-state positive, and one-step bootstrapping are unchanged. The ablation therefore changes only the temporal sampling of future candidates in Equation~\eqref{eq:td-loss}. Table~\ref{tab:macro-align} uses the standard LAPIS reference from Table~\ref{tab:main}. Standard sampling achieves higher success than grid-restricted sampling on cube ($62$ vs. $53$), scene ($29$ vs. $23$), and antmaze ($49$ vs. $34$), supporting future-goal candidates without a temporal-grid restriction.

\begin{table}[htbp]
\centering

\caption{Goal-candidate sampling ablation for LAPIS (success rate \%). Only the trajectory-future candidate component is grid-restricted; the standard reference is from Table~\ref{tab:main}.}
\label{tab:macro-align}
\begin{tabular}{lcc}
\toprule
Task & Standard candidates & Grid-restricted candidates \\
\midrule
\nolinkurl{v-cube-single-play} & $62${\scriptsize$\,\pm5$} & $53${\scriptsize$\,\pm4$} \\
\nolinkurl{v-scene-play} & $29${\scriptsize$\,\pm4$} & $23${\scriptsize$\,\pm3$} \\
\nolinkurl{v-antmaze-medium-navigate} & $49${\scriptsize$\,\pm3$} & $34${\scriptsize$\,\pm7$} \\
\bottomrule
\end{tabular}
\end{table}

\paragraph{Action chunk size in real-action policies.}
Table~\ref{tab:real-action-chunk} compares action chunk sizes $C=1$ and $C=3$ on three representative tasks, with one action executed before replanning in both settings. This ablation helps separate the policy-improvement mechanism from the method-specific action chunking used in the fully supervised block of Table~\ref{tab:main}.

\begin{table}[htbp]
\centering

\caption{Success rate (\%) of real-action policies with $100\%$ action labels (mean $\pm$ standard deviation over four seeds). Results for $C=1$ and LAPIS are reproduced from Table~\ref{tab:main}. New $C=3$ results average the $300$k, $400$k, and $500$k checkpoints within each seed.}
\label{tab:real-action-chunk}
\begin{tabular}{lcccc}
\toprule
Method & $C$ & AntMaze-medium & Cube-single & Scene \\
\midrule
GCIQL & 1 & \rev{$12${\scriptsize$\,\pm2$}} & \rev{$30${\scriptsize$\,\pm7$}} & $12${\scriptsize$\,\pm2$} \\
GCIQL & 3 & $19${\scriptsize$\,\pm2$} & $24${\scriptsize$\,\pm4$} & $3${\scriptsize$\,\pm1$} \\
\midrule
CRL & 1 & $95${\scriptsize$\,\pm2$} & $18${\scriptsize$\,\pm10$} & $13${\scriptsize$\,\pm3$} \\
CRL & 3 & $95${\scriptsize$\,\pm1$} & $23${\scriptsize$\,\pm10$} & $13${\scriptsize$\,\pm1$} \\
\midrule
LAPIS & 3 & $48${\scriptsize$\,\pm3$} & $45${\scriptsize$\,\pm4$} & $45${\scriptsize$\,\pm1$} \\
\bottomrule
\end{tabular}
\end{table}
\FloatBarrier

\paragraph{Guidance scale.}
Table~\ref{tab:lapis-guidance-scale} evaluates LAPIS at $w\in\{0,1,3\}$. Here $w=0$ uses the behavior branch, $w=1$ uses the high-value branch trained toward the supported target $q_G$, and $w=3$ further extrapolates in the high-value direction. On all visual manipulation tasks, most of the improvement from $w=0$ to the default $w=3$ is already obtained at $w=1$. For example, success increases from $13$ to $53$ to $62$ on cube-single, from $13$ to $29$ to $31$ on cube-double, and from $18$ to $25$ to $29$ on scene. Thus, stronger guidance provides additional gains in some tasks, but the supported high-value branch itself accounts for the dominant improvement in these manipulation domains.

\begin{table}[t]
\centering

\caption{Success rate (\%) of latent-action LAPIS with $5\%$ action labels at different guidance scales.}
\label{tab:lapis-guidance-scale}
\begin{tabular}{lccc}
\toprule
Task & $w=0$ & $w=1$ & $w=3$ \\
\midrule
\nolinkurl{v-antmaze-medium-navigate} & $30${\scriptsize$\,\pm6$} & $51${\scriptsize$\,\pm5$} & $49${\scriptsize$\,\pm3$} \\
\nolinkurl{v-antmaze-large-navigate} & $7${\scriptsize$\,\pm2$} & $8${\scriptsize$\,\pm2$} & $10${\scriptsize$\,\pm2$} \\
\nolinkurl{v-antmaze-giant-navigate} & $0${\scriptsize$\,\pm0$} & $0${\scriptsize$\,\pm0$} & $0${\scriptsize$\,\pm0$} \\
\nolinkurl{v-antmaze-teleport-navigate} & $14${\scriptsize$\,\pm4$} & $16${\scriptsize$\,\pm4$} & $21${\scriptsize$\,\pm3$} \\
\midrule
\nolinkurl{v-cube-single-play} & $13${\scriptsize$\,\pm1$} & $53${\scriptsize$\,\pm3$} & $62${\scriptsize$\,\pm5$} \\
\nolinkurl{v-cube-double-play} & $13${\scriptsize$\,\pm1$} & $29${\scriptsize$\,\pm4$} & $31${\scriptsize$\,\pm6$} \\
\nolinkurl{v-cube-triple-play} & $18${\scriptsize$\,\pm1$} & $20${\scriptsize$\,\pm1$} & $20${\scriptsize$\,\pm0$} \\
\nolinkurl{v-cube-quadruple-play} & $10${\scriptsize$\,\pm4$} & $13${\scriptsize$\,\pm3$} & $14${\scriptsize$\,\pm3$} \\
\midrule
\nolinkurl{v-scene-play} & $18${\scriptsize$\,\pm3$} & $25${\scriptsize$\,\pm4$} & $29${\scriptsize$\,\pm4$} \\
\bottomrule
\end{tabular}
\end{table}

\paragraph{CFGRL in physical and learned latent action spaces.}
Because LAPIS adopts classifier-free guided policy extraction, we additionally
compare with the value-free goal-conditioned formulation of CFGRL
\citep{frans2025cfgrl}. In this setting, standard flow-based GCBC corresponds
to $w=1$, while $w>1$ strengthens goal guidance at inference without learning an
explicit value function. We apply the same CFGRL construction both to physical
actions and directly to learned latent actions, and sweep
$w\in\{0,1,3\}$. In the latent-action setting, CFGRL uses the same latent action model and decoder as LAPIS. Table~\ref{tab:cfgrl-action-space} reports checkpoint-averaged
success rates.

\begin{table}[t]
\centering
\caption{\textbf{Goal-conditioned CFGRL in physical and learned latent action
spaces.} Success rate (\%). The comparison focuses on the effect of increasing
guidance within each action representation.}
\label{tab:cfgrl-action-space}

\begin{tabular}{llccc}
\toprule
Task & Action space & $w=0$ & $w=1$ & $w=3$ \\
\midrule
\nolinkurl{v-antmaze-medium-navigate} & Real   & $7.2${\scriptsize$\,\pm1.9$} & $15.0${\scriptsize$\,\pm1.0$} & $24.1${\scriptsize$\,\pm4.0$} \\
               & Latent & $6.7${\scriptsize$\,\pm1.5$} & $5.0${\scriptsize$\,\pm0.7$} & $6.1${\scriptsize$\,\pm0.9$}  \\
\midrule
\nolinkurl{v-cube-single-play} & Real   & $0.7${\scriptsize$\,\pm0.2$} & $1.5${\scriptsize$\,\pm1.7$} & $4.0${\scriptsize$\,\pm5.8$} \\
               & Latent & $0.5${\scriptsize$\,\pm0.5$} & $0.6${\scriptsize$\,\pm0.3$} & $0.2${\scriptsize$\,\pm0.2$} \\
\midrule
\nolinkurl{v-scene-play} & Real   & $2.8${\scriptsize$\,\pm0.5$} & $4.4${\scriptsize$\,\pm1.0$} & $3.9${\scriptsize$\,\pm1.0$} \\
               & Latent & $1.8${\scriptsize$\,\pm0.5$} & $1.6${\scriptsize$\,\pm0.8$} & $1.7${\scriptsize$\,\pm0.9$} \\
\bottomrule
\end{tabular}
\end{table}

Increasing guidance improves physical-action CFGRL most clearly on AntMaze,
where success rises from $7.2$ at $w=0$ to $24.1$ at $w=3$. In contrast,
directly applying the same goal-conditioned guidance mechanism in the learned
latent space produces no consistent improvement across the three domains:
performance remains nearly unchanged, or decreases, as $w$ increases. Thus,
the gains associated with classifier-free guidance in an executable physical
action space do not automatically transfer when the guided variable is a learned
latent action.

This difference is consistent with the cross-representation issue studied in
Section~\ref{sec:setup}. CFGRL's goal-conditioned construction treats the generated variable
as the action whose future-goal occupancy defines the improvement signal. For a
physical action, this action is executed directly. A learned latent action,
however, is an outcome label inferred from an observed transition and must
subsequently be decoded into physical control. Improving a distribution over
latent actions therefore does not by itself guarantee improvement after
decoding: the selected latent must be appropriate for the current state and its
represented transition must be faithfully realized by the decoder. These are
precisely the additional correspondences captured by supported outcome
identifiability and latent-to-physical execution in Theorem~\ref{thm:eval}.

LAPIS addresses this issue before guided inference. Rather than using guidance
itself to construct an improved latent policy, LAPIS first scores only observed
state--latent transitions and forms the critic-filtered distribution
$q_G(z\mid s,g)$, whose support is contained in the observed conditional
distribution $q_{\B}(z\mid s,g)$. The high-value flow branch is then trained toward
$q_G$, and classifier-free guidance is applied only after this support-aware
target has been constructed. This distinction is already visible without
extrapolative guidance: at $w=1$, latent CFGRL remains weak on AntMaze-medium,
Cube-single, and Scene, whereas the LAPIS high-value branch reaches
$51\%$, $53\%$, and $25\%$, respectively (Table~\ref{tab:lapis-guidance-scale}).
Thus, the main difference is not the use of classifier-free guidance itself,
but the construction of the latent policy-improvement target on observed
state--latent transitions.

\subsection{Action decodability of the latent representation}\label{app:zar2}

\paragraph{Probe protocol.}
We fit a two-hidden-layer MLP (256 units per layer, ReLU activations) from the frozen, unquantized latent $z_e$ to the current action $a_t$. For each OGBench task, we use independently trained $H=3$ LAMs at $500$k training steps. The MLP is trained on the labeled $5\%$ of training trajectories. For OGBench, probe validation and test trajectories are disjoint subsets of the official validation dataset. For D4RL Kitchen, we use four independently trained partial-data LAMs (seeds $0$--$3$) at $500$k steps with $H=3$, and split held-out rendered episodes into disjoint probe validation and test sets. Latent standardization uses probe training data only; we select the MLP by validation MSE over at most $200$ epochs and report test $R^2$ in Tables~\ref{tab:zar2-perdim} and~\ref{tab:kitchen-zar2}.

For action coordinate $d$, let $\mathrm{SSE}_d=\sum_i(a_{id}-\hat a_{id})^2$ and $\mathrm{SST}_d=\sum_i(a_{id}-\bar a_d)^2$ on the test set. We report $R_d^2=1-\mathrm{SSE}_d/\mathrm{SST}_d$ and the variance-weighted aggregate
\[
R^2_{\mathrm{Overall}}=1-\frac{\sum_d\mathrm{SSE}_d}{\sum_d\mathrm{SST}_d}.
\]
For Tables~\ref{tab:zar2-perdim}--\ref{tab:horizon-lam-policy}, we first average the three probe runs within each LAM seed. Per-coordinate entries are then averaged across LAM seeds, while Overall reports the mean $\pm$ sample standard deviation across LAM seeds.

\begin{table}[htbp]
\centering

\setlength{\tabcolsep}{4pt}
\caption{MLP action-decodability $R^2$ on held-out OGBench trajectories ($z_e\!\to\!a_t$, $H=3$). Overall is the variance-weighted $R^2$. Cube/scene actions have five coordinates; antmaze actions have eight.}
\label{tab:zar2-perdim}
\resizebox{\linewidth}{!}{%
\begin{tabular}{lccccccccc}
\toprule
Task & $a_0$ & $a_1$ & $a_2$ & $a_3$ & $a_4$ & $a_5$ & $a_6$ & $a_7$ & Overall $R^2$ \\
\midrule
\nolinkurl{v-antmaze-medium-navigate} & $0.68$ & $0.47$ & $0.61$ & $0.62$ & $0.70$ & $0.49$ & $0.62$ & $0.49$ & $0.597${\scriptsize$\,\pm0.001$} \\
\nolinkurl{v-antmaze-large-navigate} & $0.69$ & $0.46$ & $0.60$ & $0.61$ & $0.70$ & $0.42$ & $0.64$ & $0.53$ & $0.596${\scriptsize$\,\pm0.001$} \\
\nolinkurl{v-antmaze-giant-navigate} & $0.71$ & $0.50$ & $0.63$ & $0.59$ & $0.68$ & $0.45$ & $0.58$ & $0.47$ & $0.588${\scriptsize$\,\pm0.003$} \\
\nolinkurl{v-antmaze-teleport-navigate} & $0.65$ & $0.50$ & $0.59$ & $0.61$ & $0.63$ & $0.41$ & $0.63$ & $0.48$ & $0.575${\scriptsize$\,\pm0.002$} \\
\midrule
\nolinkurl{v-cube-single-play} & $0.83$ & $0.92$ & $0.96$ & $0.03$ & $0.83$ & -- & -- & -- & $0.778${\scriptsize$\,\pm0.000$} \\
\nolinkurl{v-cube-double-play} & $0.80$ & $0.91$ & $0.95$ & $-0.10$ & $0.81$ & -- & -- & -- & $0.740${\scriptsize$\,\pm0.001$} \\
\nolinkurl{v-cube-triple-play} & $0.84$ & $0.94$ & $0.97$ & $-0.01$ & $0.83$ & -- & -- & -- & $0.773${\scriptsize$\,\pm0.000$} \\
\nolinkurl{v-cube-quadruple-play} & $0.85$ & $0.94$ & $0.97$ & $0.19$ & $0.84$ & -- & -- & -- & $0.806${\scriptsize$\,\pm0.001$} \\
\midrule
\nolinkurl{v-scene-play} & $0.90$ & $0.92$ & $0.94$ & $0.57$ & $0.79$ & -- & -- & -- & $0.811${\scriptsize$\,\pm0.000$} \\
\bottomrule
\end{tabular}%
}
\end{table}

\begin{table}[htbp]
\centering

\setlength{\tabcolsep}{4pt}
\caption{MLP action-decodability $R^2$ on held-out D4RL Kitchen trajectories ($z_e\!\to\!a_t$, $H=3$). Overall is the variance-weighted $R^2$.}
\label{tab:kitchen-zar2}
\begin{tabular}{lcccccccccc}
\toprule
Task & $a_0$ & $a_1$ & $a_2$ & $a_3$ & $a_4$ & $a_5$ & $a_6$ & $a_7$ & $a_8$ & Overall $R^2$ \\
\midrule
\texttt{D4RL Kitchen} & $0.91$ & $0.46$ & $0.43$ & $0.81$ & $0.50$ & $0.45$ & $0.53$ & $0.38$ & $0.33$ & $0.660${\scriptsize$\,\pm0.008$} \\
\bottomrule
\end{tabular}
\end{table}
\FloatBarrier

\paragraph{Temporal gap and action decodability.}\label{app:horizon-decodability}
We compare $H=1$ and $H=3$ representations using matched samples and frozen $500$k LAMs. Both probes predict the same current action $a_t$ from the unquantized latent $z_e$ using the MLP protocol above. Table~\ref{tab:horizon-lam-policy} reports variance-weighted test $R^2$. The $H=3$ representation gives higher action decodability in all three domains. A plausible explanation is that a longer observation gap exposes more visible action-dependent changes, making the transition signal easier to retain under visual augmentation and quantization. These probes quantify the action information recoverable from the learned representations.

\begin{table}[htbp]
\centering

\setlength{\tabcolsep}{6pt}
\caption{Temporal gap and action decodability. Variance-weighted test $R^2$ for an MLP predicting $a_t$ from $z_e$, using matched samples and frozen $500$k LAMs.}
\label{tab:horizon-lam-policy}
\begin{tabular}{lcc}
\toprule
Environment & $H=1$ & $H=3$ \\
\midrule
\nolinkurl{v-antmaze-medium-navigate} & $0.501${\scriptsize$\,\pm0.002$} & $0.597${\scriptsize$\,\pm0.001$} \\
\nolinkurl{v-cube-single-play} & $0.586${\scriptsize$\,\pm0.001$} & $0.778${\scriptsize$\,\pm0.000$} \\
\nolinkurl{v-scene-play} & $0.628${\scriptsize$\,\pm0.001$} & $0.811${\scriptsize$\,\pm0.000$} \\
\bottomrule
\end{tabular}
\end{table}
\FloatBarrier

\subsection{Qualitative rollouts}\label{sec:exp-qualitative}

\begin{figure}[H]
\centering
\includegraphics[width=\textwidth]{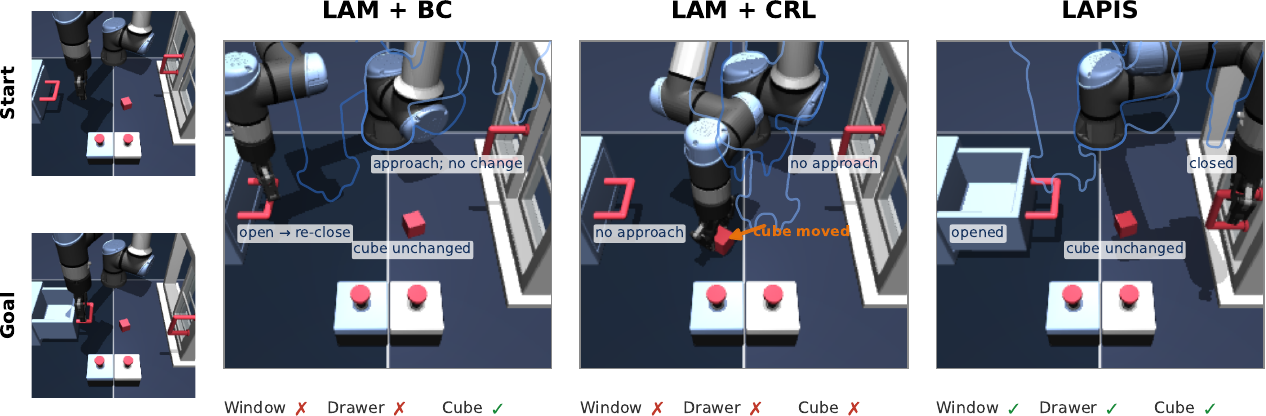}
\caption{Qualitative rollouts on \texttt{scene} ($5\%$ labels), shared start and goal.}
\label{fig:qualitative_policy}
\end{figure}

Figure~\ref{fig:qualitative_policy} compares rollouts from a shared start state with the goal of closing the window, opening the drawer, and keeping the cube unchanged. Each panel overlays the gripper trajectory and two interaction frames on the final state, with per-subtask outcomes below. Recorded simulator positions confirm that LAM${+}$BC opens then re-closes the drawer and leaves the window unchanged; LAM${+}$CRL neither opens the drawer nor closes the window and moves the cube off-target; LAPIS completes all three subtasks.

\begin{revblock}
\subsection{Extraction rule and critic}\label{app:extraction-grid}

\rev{Table~\ref{tab:ablation}(b) separates the contribution of the extraction rule from that of the critic in the same latent space. All entries use the same frozen LAMs and decoders, $5\%$ action labels, flow-actor architecture, training budget, and evaluation protocol as LAPIS (four seeds; the $300$k, $400$k, and $500$k checkpoints are averaged within each seed). The TD-InfoNCE critic is the LAPIS critic. The IQL critic is goal-conditioned IQL~\citep{kostrikov2022iql} over latent actions with expectile $0.9$. The extraction rules are as follows.}
\begin{itemize}
\item \rev{\textbf{Gate + flow}: the LAPIS actor of Equation~\eqref{eq:actor}, evaluated at $w=3$. With the TD-InfoNCE critic this is LAPIS (Table~\ref{tab:main}).}
\item \rev{\textbf{AWR-weighted flow}: the high-value branch is trained with unnormalized weights $\exp(\beta A)$, clipped at $100$, instead of the gate~\citep{peng2019awr}, with a per-environment temperature $\beta$; the behavior branch is unchanged. Evaluated at $w=3$.}
\item \rev{\textbf{IDQL-style selection}: $32$ latents are sampled from the behavior branch ($w=0$) of the trained flow actor, and the one with the highest critic value is executed~\citep{hansenestruch2023idql}. With the TD-InfoNCE critic this is LAPIS$^\dagger$ (Table~\ref{tab:main}).}
\item \rev{\textbf{DDPG+BC flow}: a one-step policy is trained to maximize the critic while being distilled from a flow behavior policy, as in FQL~\citep{park2025fql}.}
\end{itemize}

\rev{Results are in Table~\ref{tab:ablation}(b) (Section~\ref{sec:exp-ablation}); this grid was Table~18 in the previous revision.}

\subsection{Execution failure of generated latents}\label{app:x04}

\rev{\paragraph{Protocol.} We measure whether latents generated by a policy are executed faithfully, on \texttt{cube-single} and \texttt{scene} with four seeds, at every decision step of the evaluation episodes (five tasks and $20$ episodes per task). At each decision step we save the full simulator state, decode the selected latent into its three-action chunk, execute the chunk open loop, re-encode the start observation and the observation three steps later, and record the round-trip error $\rho=\|I(o_t,\tilde o_{t+3})-z\|_2$ in the standardized quantized latent coordinates used for $d_{\mathrm{round}}$ in Figure~\ref{fig:diagnostic}. We then restore the saved state and continue the episode; for every run we checked that restoring reproduces the simulator state exactly and that probing does not change the episode.}

\rev{\paragraph{Reference and failure rate.} For each task and LAM, we set the simulator to $2{,}000$ dataset states and execute the decoded chunk of the recorded dataset latent $z=I(s_t,s_{t+3})$, which gives a reference distribution of $\rho$. A decision step fails when $\rho$ exceeds the $95$th percentile of this reference distribution, so dataset latents fail $5\%$ of the time by construction. The threshold is computed for the LAM and decoder that each policy uses.}

\rev{\paragraph{Settings.} LAPIS at $w=3$ (the reported setting); LAPIS at $w=3$ without codebook projection, where the generated continuous latent is decoded directly; AWR-weighted flow (Appendix~\ref{app:extraction-grid}) at $w=3$; and max-$Q$, CRL-style critic maximization with a flow actor~\citep{eysenbach2022crl}, as in Table~\ref{tab:ablation}\rev{(a)}.}

\begin{table}[htbp]
\centering
\caption{\rev{Execution failure rate (\%, $\downarrow$) of generated latents: the fraction of decision steps whose round-trip error exceeds the $95$th percentile of dataset latents. Mean $\pm$ std over four seeds. Bold marks the lowest rate among policies in each row.}}
\label{tab:x04-fail}
\begin{revblock}
\setlength{\tabcolsep}{4pt}
\begin{tabular}{lcccc}
\toprule
Task & Dataset latents & LAPIS ($w=3$) & AWR-flow & max-$Q$ \\
\midrule
\texttt{cube-single} & $5.0$ & $\mathbf{5.8}${\scriptsize$\,\pm2.3$} & $7.1${\scriptsize$\,\pm0.8$} & $6.7${\scriptsize$\,\pm1.6$} \\
\texttt{scene} & $5.0$ & $\mathbf{12.6}${\scriptsize$\,\pm3.3$} & $15.4${\scriptsize$\,\pm4.7$} & $24.0${\scriptsize$\,\pm4.7$} \\
\bottomrule
\end{tabular}
\end{revblock}
\end{table}

\begin{table}[htbp]
\centering
\caption{\rev{Effect of codebook projection on LAPIS at $w=3$, evaluated on the same checkpoints with and without projection. The success difference is paired by seed (mean $\pm$ std over four seeds).}}
\label{tab:x04-snap}
\begin{revblock}
\begin{tabular}{lcc}
\toprule
 & \texttt{cube-single} & \texttt{scene} \\
\midrule
Failure rate (\%), with / without projection & $5.8$ / $5.8$ & $12.6$ / $12.4$ \\
Success difference, without $-$ with (points) & $-2.3${\scriptsize$\,\pm4.0$} & $+1.5${\scriptsize$\,\pm6.5$} \\
\bottomrule
\end{tabular}
\end{revblock}
\end{table}

\rev{\paragraph{Results.} LAPIS latents fail at nearly the dataset rate on \texttt{cube-single} ($5.8\%$ vs.\ $5\%$) and at the lowest rate among the compared methods on \texttt{scene} ($12.6\%$), where max-$Q$ fails about twice as often ($24.0\%$). Removing codebook projection leaves the failure rate unchanged, and the success difference lies within seed variation (Table~\ref{tab:x04-snap}). Executability and performance therefore come from the latents generated by the policy itself, not from the projection. For guidance strength, the $d_{\mathrm{round}}$ panel of Figure~\ref{fig:diagnostic} reports the mean round-trip error of LAPIS over $w$ on \texttt{antmaze-medium}.}
\end{revblock}

\Needspace{16\baselineskip}
\begin{revblock}
\subsection{\texorpdfstring{$5\%$}{5\%} labels without LAM}\label{app:wolam}

Table~\ref{tab:wolam} reports the $5\%$-label setting without LAM pretraining, where policies are trained on the labeled trajectories only.

\begin{table}[htbp]
\centering
\caption{\rev{Success rate (\%) on the OGBench visual suite with $5\%$ action labels and without LAM pretraining.}}
\label{tab:wolam}
\begin{revblock}
\setlength{\tabcolsep}{4pt}
\begin{tabular}{lccccccc}
\toprule
\texttt{Task} & \texttt{F-GCBC} & \texttt{GCIVL} & \texttt{GCIQL} & \texttt{QRL} & \texttt{HIQL} & \texttt{CRL} & \texttt{LAPIS} \\
\midrule
\nolinkurl{antmaze-medium} & 3{\scriptsize$\,\pm0$} & 3{\scriptsize$\,\pm1$} & 4{\scriptsize$\,\pm1$} & 0{\scriptsize$\,\pm0$} & \textbf{8}{\scriptsize$\,\pm5$} & \textbf{8}{\scriptsize$\,\pm1$} & 2{\scriptsize$\,\pm1$} \\
\nolinkurl{antmaze-large} & 1{\scriptsize$\,\pm0$} & 1{\scriptsize$\,\pm0$} & 0{\scriptsize$\,\pm0$} & 0{\scriptsize$\,\pm1$} & \textbf{2}{\scriptsize$\,\pm2$} & 0{\scriptsize$\,\pm1$} & 0{\scriptsize$\,\pm0$} \\
\nolinkurl{antmaze-giant} & 0{\scriptsize$\,\pm0$} & 0{\scriptsize$\,\pm0$} & 0{\scriptsize$\,\pm0$} & 0{\scriptsize$\,\pm0$} & 0{\scriptsize$\,\pm0$} & 0{\scriptsize$\,\pm0$} & 0{\scriptsize$\,\pm0$} \\
\nolinkurl{antmaze-teleport} & 2{\scriptsize$\,\pm1$} & 1{\scriptsize$\,\pm1$} & 1{\scriptsize$\,\pm1$} & 1{\scriptsize$\,\pm1$} & \textbf{3}{\scriptsize$\,\pm2$} & 1{\scriptsize$\,\pm1$} & 2{\scriptsize$\,\pm1$} \\
\midrule
\nolinkurl{cube-single} & 2{\scriptsize$\,\pm1$} & 5{\scriptsize$\,\pm1$} & 7{\scriptsize$\,\pm2$} & \textbf{8}{\scriptsize$\,\pm3$} & 5{\scriptsize$\,\pm1$} & 1{\scriptsize$\,\pm1$} & 4{\scriptsize$\,\pm1$} \\
\nolinkurl{cube-double} & 0{\scriptsize$\,\pm0$} & 0{\scriptsize$\,\pm0$} & 0{\scriptsize$\,\pm0$} & 0{\scriptsize$\,\pm0$} & \textbf{1}{\scriptsize$\,\pm1$} & 0{\scriptsize$\,\pm0$} & 0{\scriptsize$\,\pm0$} \\
\nolinkurl{cube-triple} & 0{\scriptsize$\,\pm0$} & 2{\scriptsize$\,\pm1$} & 0{\scriptsize$\,\pm0$} & 0{\scriptsize$\,\pm0$} & \textbf{4}{\scriptsize$\,\pm0$} & 0{\scriptsize$\,\pm0$} & 1{\scriptsize$\,\pm1$} \\
\nolinkurl{cube-quadruple} & 0{\scriptsize$\,\pm0$} & 0{\scriptsize$\,\pm0$} & 0{\scriptsize$\,\pm0$} & 0{\scriptsize$\,\pm0$} & 0{\scriptsize$\,\pm0$} & 0{\scriptsize$\,\pm0$} & 0{\scriptsize$\,\pm0$} \\
\midrule
\nolinkurl{scene} & 2{\scriptsize$\,\pm1$} & 4{\scriptsize$\,\pm2$} & 1{\scriptsize$\,\pm1$} & 2{\scriptsize$\,\pm2$} & \textbf{8}{\scriptsize$\,\pm2$} & 3{\scriptsize$\,\pm1$} & 4{\scriptsize$\,\pm0$} \\
\bottomrule
\end{tabular}
\end{revblock}
\end{table}
\end{revblock}

\clearpage
\newpage\section{Additional Discussion of Experimental Settings}\label{app:additional-discussion}

This appendix provides additional context for several experimental comparisons whose interpretation involves differences in policy-improvement mechanisms, supervision protocols, decoder training, or environment dynamics. These details are complementary to the implementation specifications in Appendix~\ref{app:expdetails} and the theory--practice correspondence in Appendix~\ref{app:gaps}.

\subsection{\texorpdfstring{\rev{AntMaze performance and conservative extraction}}{AntMaze performance and conservative extraction}}\label{app:fully-supervised-antmaze}

Table~\ref{tab:main} shows a substantial gap between fully supervised real-action CRL and real-action LAPIS on AntMaze, particularly on the larger mazes. These results highlight a tradeoff of the support-aware update used by LAPIS under full physical-action supervision.

LAPIS is motivated by a failure specific to learned latent-action spaces: direct value maximization can select latent actions whose represented transitions are not physically realizable after decoding. The latent realizability failure underlying this mismatch is absent when the policy directly outputs physical actions, although ordinary offline-RL distribution shift remains. In this setting, restricting policy improvement toward behavior-supported transitions can be unnecessarily conservative, while more aggressive critic-driven optimization may be beneficial. Conversely, in the learned latent-action setting, direct maximization can move the policy toward unrealizable latents, motivating the supported update used by LAPIS.

The larger gap on more difficult AntMaze tasks is therefore consistent with a cost of conservative policy extraction in long-horizon navigation. Accordingly, the fully supervised results provide physical-action references for the limited-supervision comparisons. Our main question is whether effective policy improvement can be retained after replacing physical actions with learned latent actions under limited action supervision.

\subsection{Protocol differences across baseline tables}\label{app:baseline-protocol-discussion}

The GCIQL entries in Table~\ref{tab:main} and the \rev{GCIVL entries in Table~\ref{tab:semisup} (labeled GCIQL in the original submission)} differ in their value-learning and policy-extraction objectives. Table~\ref{tab:main} uses an action-conditioned critic and a state--goal value function, with policy extraction based on the $Q-V$ advantage. This formulation is applied to physical actions at the stated label budgets and to learned latent actions in the LAM setting.

Table~\ref{tab:semisup} uses an action-scarce adaptation that learns a state--goal value function from all observed transitions, without requiring action labels for value learning. Its physical-action policy is trained only on labeled transitions, weighted by the value difference between successive states. The two tables therefore \rev{use different methods} rather than repeated evaluations of the same procedure. \rev{Because the Table~\ref{tab:semisup} variant learns no action-conditioned $Q$, it corresponds to GCIVL in OGBench~\citep{park2025ogbench}, and we name it accordingly.} \rev{GCIVL in Tables~\ref{tab:main} and~\ref{tab:wolam} is the standard OGBench GCIVL, trained only on transitions with (physical or latent) actions; the Table~\ref{tab:semisup} variant also learns its value from action-free transitions.}

\subsection{Interpreting decoder label-budget trends}\label{app:decoder-budget-discussion}

Table~\ref{tab:decoder-cond} isolates decoder supervision from latent representation learning: all label budgets use the same frozen LAM trained with $5\%$ action labels, while only the amount of action supervision available to the decoder is varied. Thus, the non-monotonic performance of the state-marginal decoder across $1\%$, $5\%$, and $100\%$ decoder labels is not caused by changes in the learned latent representation.

The state-marginal decoder $\Psi(a\mid z)$ aggregates physical realizations across states without observing the current state. If the same latent code admits different appropriate controls at different states, broader coverage can make the empirical conditional $p(a\mid z)$ more heterogeneous, partially offsetting the benefit of additional labeled examples. By contrast, the state-conditioned decoder $\Psi(a\mid s,z)$ can use the current state to resolve such variation when sufficient supervision is available.

This provides a plausible interpretation of the pattern in Table~\ref{tab:decoder-cond}: the marginal decoder performs better under $1\%$ and $5\%$ supervision, where sharing action information across states can be beneficial, while the state-conditioned decoder performs better with full supervision. The fact that the marginal cube score at $5\%$ exceeds its $100\%$-label score should therefore not be interpreted as evidence that fewer labels are intrinsically preferable; rather, performance reflects the interaction between decoder conditioning, labeled-data coverage, and finite-sample optimization. \rev{LAPO~\citep[Sec.~6.2]{schmidt2024lapo} reports the related observation that a decoder with access only to the latent action plateaus as labeled data grows, which it attributes to latents that are not state-invariant; it does not compare against a state-conditioned decoder. Table~\ref{tab:decoder-cond} thus measures the gap between the analyzed decoder ${\Psi^*}$ and the implemented ${\Psi(a\mid z)}$, and shows that it depends on the label budget.}

\subsection{Stochastic teleport dynamics}\label{app:teleport-discussion}

The visual-antmaze-teleport benchmark provides a concrete case in which the deterministic-dynamics assumption used for the exact transfer result does not hold globally. In the OGBench teleport environment, entering an inbound teleport region sends the agent to a randomly selected outbound portal. Thus, the same physical state and control can induce different successor states at a teleport transition. This violates Assumption~\ref{ass:detT} and means that the exact equality in Theorem~\ref{thm:eval} is not expected to apply at such transitions. The stochastic teleport implementation can be verified directly in \href{https://github.com/seohongpark/ogbench/blob/master/ogbench/locomaze/maze.py}{OGBench}.

This case is instead described by the approximate-transfer analysis in Appendix~\ref{app:stochastic-transfer}. For a decoded latent policy, Proposition~\ref{prop:stoch-transfer} bounds the return discrepancy by
\[
\left|J_{\mathcal M}(\widehat\Psi\circ\pi_z)-J_{\MBz}(\pi_z)\right|
\leq \frac{\bar\epsilon_r}{1-\gamma}+\frac{2\gamma R_{\max}}{(1-\gamma)^2}\bar\epsilon_p,
\]
where $\bar\epsilon_p$ measures the transition mismatch between latent evaluation and decoded physical execution and $\bar\epsilon_r$ measures the corresponding reward mismatch. Random teleportation can increase $\epsilon_p$ because a latent transition inferred from one observed teleport outcome need not reproduce that same outcome when the decoded action is executed again. The exact deterministic result is recovered when these mismatches vanish.

The teleport task evaluates performance under stochastic transitions, for which the transition- and reward-mismatch framework in Appendix~\ref{app:stochastic-transfer} provides the relevant theoretical perspective.

\subsection{Standalone F-GCBC and the LAPIS behavior branch}\label{app:fgcbc-behavior-branch}

Table~\ref{tab:main} reports F-GCBC as a standalone flow-based goal-conditioned behavioral-cloning baseline, whereas $w=0$ in Table~\ref{tab:lapis-guidance-scale} evaluates the behavior-conditioned branch of a jointly trained LAPIS actor. Although both are trained from observed latent transitions using flow matching, they are not the same trained policy. LAPIS uses a single conditional velocity model $v_\theta(\cdot;o)$ whose $o=0$ branch is trained toward the observed distribution $q_{\B}$ and whose $o=1$ branch is simultaneously trained toward the critic-filtered distribution $q_G$ through Equation~\eqref{eq:actor}. The two losses therefore update shared model parameters.

Consequently, setting $w=0$ at inference selects the behavior-conditioned velocity field of the jointly trained two-branch model. Table~\ref{tab:lapis-guidance-scale} is an inference-time guidance ablation within a fixed LAPIS actor, whereas Table~\ref{tab:main} evaluates a separately trained F-GCBC policy. Their different training objectives and parameter sharing explain the difference between their scores.

\clearpage
\newpage
\section{Future Work and Limitations}\label{app:future-limitations}

\subsection{Future Work}

A promising direction is to extend LAPIS from goal-conditioned to language-conditioned control. Prior work has shown that language instructions can be grounded into visual goal representations and executed through goal-conditioned policies~\citep{myers2023goal,cachet2024language}. This suggests that LAPIS could retain its latent-space policy-improvement framework while replacing explicit visual goals with language-derived goal representations, enabling policy improvement for instruction-following and vision-language-action settings. Extending the framework and its transfer analysis to partially observed settings using temporal or recurrent state representations is another important direction.

\subsection{Limitations}

Our experiments are limited to simulated visual manipulation and navigation benchmarks, and we focus on goal-conditioned control with limited physical-action supervision. Consequently, we do not empirically study objectives with action-dependent rewards, where outcome-based abstraction can create a mismatch between latent and physical policy objectives. Real-robot deployment and language-conditioned control also remain outside the scope of the current evaluation.

Our exact transfer and alignment results rely on deterministic dynamics, supported outcome identifiability, batch-constrained execution through the batch-exact decoder, and Markov state information. Practical LAPIS uses learned visual state representations, a learned state-marginal decoder, an $H=3$ latent representation with one-step replanning, and classifier-free guidance, so these conditions are not instantiated exactly. We explicitly study these departures in Appendix~\ref{app:gaps} and through targeted empirical analyses. The approximate-transfer analysis accounts for transition and decoder mismatch, while decoder-conditioning, temporal-horizon, and first-step execution analyses examine learned decoding and the $H>1$ receding-horizon controller. For guided inference, the $w=1$ high-value branch provides the supported target underlying our analysis, and the guidance sweep shows that it already captures most of the improvement on the manipulation tasks; $w>1$ provides an additional inference-time extrapolation beyond this exact support guarantee. The direct CFGRL comparison further shows that guidance alone does not provide consistent improvement when applied directly in the learned latent space, reinforcing the role of support-aware target construction in practical LAPIS.

\end{document}